\documentclass[sigconf]{acmart}
\renewcommand\footnotetextcopyrightpermission[1]{} 
\setcopyright{none}

\usepackage{tikz}
\usetikzlibrary{shapes,arrows}
\usetikzlibrary{backgrounds, positioning, fit}
\usepackage{dblfloatfix}
\usepackage[inline]{enumitem}
\usepackage{graphics}
\usepackage[T1]{fontenc}
\usepackage{soul}
\usepackage{xspace}
\usepackage{multirow}
\usepackage[table]{colortbl}
\usepackage{versions}
\usepackage{pifont}
\usepackage{subcaption}
\usepackage[normalem]{ulem}

\usepackage{amsmath,amsthm}
\usepackage{booktabs}
\newtheorem{theorem}{Theorem}

\newcommand{\dm}{$\epsilon_\theta$\xspace}
\newcommand{\adv}{\protect{$\mathcal{A}dv$}\xspace}
\newcommand{\tti}{T2I\xspace}
\newcommand{\crt}{CRT\xspace}
\newcommand{\te}{$\phi$\xspace}
\newcommand{\prompt}{\texttt{p}\xspace}
\newcommand{\promptacc}{\texttt{p}^{a}\xspace}
\newcommand{\promptuacc}{\texttt{p}^{u}\xspace}
\newcommand{\advprompt}{\texttt{p}^{adv}\xspace}
\newcommand{\filter}{\texttt{F}\xspace}
\newcommand{\target}{\emph{f}\xspace}

\newcommand{\imgclean}{\texttt{$x$}\xspace}
\newcommand{\imgadv}{\texttt{$x$}^{adv}\xspace}
\newcommand{\concept}{c\xspace}
\newcommand{\cacc}{c^{a}\xspace}
\newcommand{\caccc}{\hat{c}^{a}\xspace}
\newcommand{\cunacc}{c^{u}\xspace}
\newcommand{\cunaccc}{\hat{c}^{u}\xspace}
\newcommand{\imgunacc}{\texttt{$x^{u}$}\xspace}
\newcommand{\imgacc}{\texttt{$x^{a}$}\xspace}
\newcommand{\dtrain}{\mathcal{D}_{tr}\xspace}
\newcommand{\dtestadv}{\mathcal{D}_{adv}^{u}\xspace}
\newcommand{\dtestacc}{\mathcal{D}_{te}^{a}\xspace}
\newcommand{\dtestunacc}{\mathcal{D}_{te}^{u}\xspace}
\newcommand{\dvalacc}
{\mathcal{D}_{val}^{a}\xspace}
\newcommand{\dvalunacc}
{\mathcal{D}_{val}^{u}\xspace}

\newcommand{\method}{\textsc{Espresso}\xspace}

\newcommand{\esd}{\texttt{ESD}\xspace}
\newcommand{\sdd}{\texttt{SDD}\xspace}
\newcommand{\unifiedCE}{\texttt{UCE}\xspace}
\newcommand{\unsafediff}{\texttt{UD}\xspace}
\newcommand{\sa}{\texttt{SA}\xspace}
\newcommand{\forgetNot}{\texttt{FMN}\xspace}

\newcommand{\ca}{\texttt{CA}\xspace}
\newcommand{\moderator}{\texttt{Mod}\xspace}

\newcommand{\cce}{\texttt{CCE}\xspace}
\newcommand{\pez}{\texttt{PEZ}\xspace}
\newcommand{\ringbell}{\texttt{RingBell}\xspace}
\newcommand{\typo}{\texttt{Typo}\xspace}
\newcommand{\sneaky}{\texttt{SneakyPrompt}\xspace}

\usepackage{xcolor}
\definecolor{mynicegreen}{RGB}{171,195,47}
\definecolor{mynicered}{RGB}{255,154,154}
\definecolor{myniceblue}{RGB}{153,153,255}

\def\Snospace~{\S{}}

\includeversion{arxiv}
\excludeversion{submission}

\begin{document}

\title{\method: Robust Concept Filtering in Text-to-Image Models}

\author{Anudeep Das}
\affiliation{%
  \institution{University of Waterloo}
  \country{}
}
\email{anudeep.das@uwaterloo.ca}

\author{Vasisht Duddu}
\affiliation{%
  \institution{University of Waterloo}
  \country{}
}
\email{vasisht.duddu@uwaterloo.ca}

\author{Rui Zhang}
\authornote{Work done while visiting the Secure Systems Group, University of Waterloo.}
\affiliation{%
  \institution{Zhejiang University}
  \country{}
}
\email{zhangrui98@zju.edu.cn}

\author{N. Asokan}
\affiliation{%
  \institution{University of Waterloo}
   \country{}
}
\email{asokan@acm.org}

\begin{abstract}
Diffusion based text-to-image models are trained on large datasets scraped from the Internet, potentially containing \textit{unacceptable concepts} (e.g., copyright-infringing or unsafe). 
We need concept removal techniques (\crt{s}) which are
\begin{enumerate*}[label=\roman*),itemjoin={,\xspace}]
\item \textit{effective} in preventing the generation of images with unacceptable concepts
\item \textit{utility-preserving} on acceptable concepts, and 
\item \textit{robust} against evasion with adversarial prompts.
\end{enumerate*}
No prior \crt satisfies all these requirements simultaneously. 
We introduce \method, the first \emph{\textbf{robust \underline{co}ncept \underline{fi}lter}} based on Contrastive Language-Image Pre-Training (CLIP). 
We identify unacceptable concepts by using the distance between the embedding of a generated image to the text embeddings of \emph{both} unacceptable and acceptable concepts.
This lets us fine-tune for robustness by separating the text embeddings of unacceptable and acceptable concepts while preserving utility.
We present a pipeline to evaluate various \crt{s} to show that \method is \textbf{more effective and robust} than prior \crt{s}, while \textbf{retaining utility}.
\begin{arxiv}\footnote{ACM Conference on Data and Application Security and Privacy (CODASPY), 2025.}\end{arxiv}
\begin{submission}

\end{submission}



\end{abstract}

\maketitle
\pagestyle{plain}

\section{Introduction}\label{sec:introduction}

Diffusion-based text-to-image (\tti) models have demonstrated a remarkable ability to generate high quality images from textual prompts~\cite{stable-diffusion,Rombach_2022_CVPR,ramesh2021dalle}.
They are trained on large datasets of unfiltered content from the Internet~\cite{radford2021learning,schuhmann2022laion}. 
Due to their large capacity, \tti models memorize specific \emph{concepts}, as seen in the generated images~\cite{Korn_2023,somepalli2022diffusion,carlini2023extracting}. 
Some of these concepts, may be \emph{unacceptable} for various reasons, such as copyright infringement (e.g., a movie character or celebrity), or inappropriateness (e.g., ``nudity'' or ``violence'')~\cite{gandikota2023erasing,schramowski2022safe,heng2023selective}. 
We need \emph{concept removal techniques} (\crt{s}) to minimize unacceptable concepts in  generated images.

Ideally, \crt{s} should be \emph{effective} in reducing generation of unacceptable concepts while preserving \emph{utility} on all others, and \emph{robust} to evasion with adversarial prompts.  
No existing \crt \emph{simultaneously} satisfies these requirements (\autoref{sec:evaluation}):
\begin{enumerate*}[label=\roman*),itemjoin={,\xspace}]
\item \emph{fine-tuning \crt{s}} which modify \tti models, trade-off effectiveness for utility~\cite{gandikota2023erasing,kumari2023ablating,heng2023selective,schramowski2022safe}, and  lack robustness~\cite{Tsai2023RingABellHR, wen2023hard, pham2023circumventing, yang2023sneakyprompt}
\item \emph{filtering \crt{s}}, which detect unacceptable concepts, lack robustness~(\cite{rando2022redteaming} and \autoref{sec:filterEval}).
\end{enumerate*}
Our goal is to design a \crt that can simultaneously meet all the requirements.

We opt to use a filter as it will not alter the \tti model, thus minimizing impact on utility. We construct our filter using a Contrastive Language-Image Pre-Training (CLIP) model~\cite{radford2021learning}, an essential component of \tti models. 
CLIP is pre-trained on a vast dataset encompassing a broad spectrum of concepts~\cite{radford2021learning}, making it a versatile choice for a filter, unlike specialized classifiers (e.g.,~\cite{unsafeDiff}).
However, a naive CLIP-based filter is susceptible to evasion~\cite{rando2022redteaming}. 
\begin{arxiv}
Subsequently, prior works have identified the design of a robust filtering-based \crt as an open problem~\cite{rando2022redteaming,li2024safegen}.
\end{arxiv}


We present a \textbf{robust \underline{co}ntent \underline{fi}lter}, \method, by configuring CLIP to identify unacceptable concepts in generated images using the distance of their embeddings to the text embeddings of \emph{both} unacceptable and acceptable concepts. 
This allows fine-tuning to improve robustness by increasing the separation between the text embeddings of unacceptable and acceptable concepts, while preserving utility by maintaining the closeness between embeddings of prompts and their images. 
Our contributions are presenting 
\begin{enumerate}[leftmargin=*]
\item \method\footnote{Code: \url{https://github.com/ssg-research/concept-filtering}.}
\begin{submission}
\footnote{Full paper: \url{https://arxiv.org/abs/2404.19227}}
\end{submission}, the \textbf{first robust \underline{co}ntent \underline{fi}lter} (\autoref{sec:approach}), which identifies unacceptable concepts in generated images by measuring the distance of their embeddings to the text embeddings of \emph{both} unacceptable and acceptable concepts
\item a \textbf{complete pipeline} for comparative evaluation of \crt{s} and attacks, 
(\autoref{sec:setup}), and
\item a comprehensive evaluation of \method against seven fine-tuning \crt{s}, and one filtering \crt, showing that it is \textbf{more effective and robust} while \textbf{retaining utility} (\autoref{sec:evaluation}).
\end{enumerate}
In \autoref{sec:discussions} and \begin{arxiv}Appendix~\ref{sec:bound}\end{arxiv}\begin{submission}Appendix in our paper's full version~\cite{das2024espresso}\end{submission}, we present the first approach for certifiable robustness of \crt{s}, exploring theoretical robustness bounds of \method against a hypothetically strong adversary.
We analyze these bounds to show that \method is more robust in practice.
\section{Background}\label{sec:background}


\subsection{Diffusion based \tti Models}\label{back-diffusion}

A diffusion based \tti model is a function $\target: \prompt \rightarrow \imgclean$ which generates an image $\imgclean$ for a given a textual prompt $\prompt$. It comprises two key components: an encoder (\te) which is used to incorporate the textual prompt in the image generation process, and a diffusion model (\dm) which is responsible for the generation of the image.

A popular encoder is CLIP, trained on a large dataset of image-text pairs, to map the embeddings of images and their corresponding text closer together in a joint text-image embedding space~\cite{radford2021learning}.
Given $N$ images $\{\imgclean_j\}_{j=1}^N$ and their corresponding text prompts $\{\prompt_j\}_{j=1}^{N}$, the training data is $\mathcal{D} = \{(\imgclean_j, \prompt_j)\}_{j=1}^N$. 

CLIP is trained to maximize the cosine similarity between the embeddings of a prompt $\prompt_j$ and its corresponding image $\imgclean_j$ while minimizing the similarity between $\prompt_j$ and any other $\imgclean_k$ for a $k \ne j$.
We denote the cosine similarity as \text{cos}($\phi_p(\prompt_j), \phi_x(\imgclean_j))$, where $\phi_x(\imgclean_j)$ is the CLIP image embedding of the image $\imgclean_j$, and $\phi_p(\prompt_j)$ is the CLIP text embedding of the prompt $\prompt_j$.
To achieve this, CLIP is trained using a contrastive loss function~\cite{yang2023robust,oord2019representation}: \[\mathcal{L}_{\text{Con}}(\mathcal{D}) = - \frac{1}{N} \sum_{j=1}^{N} \log \frac{\exp(\text{cos}(\phi_x(x_j), \phi_p(\prompt_j))/ \tau)}{\sum_{k=1}^{N} \exp(\text{cos}(\phi_x(x_j), \phi_p(\prompt_k))/ \tau)}\] where $\tau$ is the temperature parameter to scale predictions~\cite{radford2021learning, ModalityGap}.

Given access to a pre-trained encoder \te, the actual images in \tti models are generated by a diffusion model, \dm, parameterized by $\theta$.
During training of \dm, Gaussian noise is added to an initial image $x_0$ for $T$ time steps to produce $x_T$, in a process known as the \emph{forward diffusion process}. The noise is then iteratively removed to approximate the initial image $\tilde{x_0}$ in the \emph{reverse diffusion process}.
During inference, the reverse diffusion process generates an image from noise.
Further, \dm can be conditioned with a textual prompt \prompt to guide the generation of $\tilde{x_0}$ to match the description in \prompt. After generating $\phi_p(\prompt)$, 
\dm is trained by minimizing the following loss function: $\mathcal{L} = \mathbb{E}_{\epsilon, \phi_p(\prompt), t}[||\epsilon - \epsilon_\theta(\imgclean_t,\phi_p(\prompt),t)||_2^2]$ for each time step $t$, and random Gaussian noise $\epsilon \sim \mathcal{N}(0,1)$.

Several prominent \tti models (e.g., Stable Diffusion v1.4 (SDv1.4)) improve the efficiency of the diffusion process by computing in the embedding space of a variational autoencoder (VAE), rather than on images~\cite{stable-diffusion}. For a VAE decoder $\mathit{D}$, VAE encoder $\mathcal{E}$, and $z_t \in \mathcal{E}(\imgclean)$ as the latent representation of $\imgclean$ in the VAE's latent space, \dm is trained by minimizing the following objective: \\
$\mathcal{L} = \mathbb{E}_{\epsilon, z_t, \phi_p(\prompt), t}[||\epsilon - \epsilon_\theta(z_t,\phi_p(\prompt),t)||_2^2]$ where $\epsilon \sim \mathcal{N}(0,1)$. \\
The final image is generated from the approximation, $\tilde{z_0}$, by passing it through $\mathit{D}$: $\tilde{x_0} = \mathit{D}(\tilde{z_0})$.  
We summarize notations in Table~\ref{tab:notations}.

\begin{table}[htb]
\caption{Frequently used notations and their descriptions.}
\begin{center}
\footnotesize
\begin{tabular}{ p{1.2cm}|p{5cm} } 
 \bottomrule

 \toprule
 \textbf{Notation} & \textbf{Description}\\
 \midrule
  \tti & Text-to-Image\\
  \te & Text encoder (e.g., CLIP)\\
  \crt & Concept Removal Technique\\
  \adv & Adversary\\
 \midrule
 $\imgclean$ & Generated image \\
 $\imgunacc$ & Generated image with unacceptable concept \\
 $\imgacc$ & Generated image with acceptable concept \\
 $\phi_x(x)$ & CLIP embedding for image $\imgclean$ \\
 $\imgadv$ & Image generated from adversarial prompt\\
 \midrule
 $\prompt$ & Textual prompt \\
 $\cacc$ & Phrase for acceptable concept\\
 $\cunacc$ & Phrase for unacceptable concept\\
  $\promptuacc$ & Textual prompt containing $\cunacc$ \\
 $\promptacc$ & Textual prompt containing $\cacc$\\
 
$\phi_p(\prompt)$ & CLIP embedding for textual prompt $\prompt$ \\
$\advprompt$ & Adversarially generated textual prompt \\
\midrule
 $\dtestacc$ & Test dataset with acceptable prompts/images\\
 $\dtestunacc$ & Test dataset with unacceptable prompts/images\\
 $\dvalacc$ & Validation dataset with acceptable prompts/images\\
 $\dvalunacc$ & Validation dataset w/ unacceptable prompts/images\\
 $\dtestadv$ & Test dataset with adversarial prompts/images\\
 $\target$ & \tti model with \crt where $\target$: $c \rightarrow x$ \\
  $\filter$ & Function for \method classifier\\
 $\alpha$ & Regularization parameter for \method fine-tuning \\
 $\overline{\cdot}$ & Normalization function \\
  $\epsilon_\theta$ & Diffusion model parameterized by $\theta$ \\
  $\tau$ & Temperature parameter \\
 \bottomrule

 \toprule
\end{tabular}
\end{center}
\label{tab:notations}
\end{table}

\subsection{Concept Removal Techniques}\label{back-conceptrem}


The textual phrase for an acceptable concept is $\cacc$, and for an unacceptable concept is $\cunacc$. 
For a given $\cunacc$, $\cacc$ is either the opposite (e.g., $\cunacc$ = \emph{violence} vs. $\cacc$ = \emph{peaceful}) or a general category of a specific character/object/person (e.g., $\cunacc$ = \emph{R2D2} vs. $\cacc$ = \emph{robot}). We discuss the selection of $\cacc$ for a given $\cunacc$ in \autoref{sec:pipeline}.
An image $\imgclean$ generated from a \tti model may either contain an unacceptable concept (referred to as $\imgunacc$) or an acceptable one (referred to as $\imgacc$). Similarly, a text prompt $\prompt$ may contain a phrase for an acceptable concept ($\promptacc$) or an unacceptable concept ($\promptuacc$). An example of an unacceptable prompt $\promptuacc$ containing an unacceptable concept $\cunacc$ = \emph{Captain Marvel}, is ``\emph{Captain Marvel} soaring through the sky''. \crt{s} thwart the generation of $\imgunacc$ by either fine-tuning the \tti model to suppress $\imgunacc$, or using a classifier as a filter to detect $\imgunacc$, and serve a replacement image instead. 

\subsubsection{Fine-tuning \crt{s}:} We first present six state-of-the-art fine-tuning \crt{s}:

\noindent\textbf{Concept Ablation (\ca)~\cite{kumari2023ablating}} fine-tunes the \tti model to minimize the KL divergence between the model's output for $\promptuacc$ and $\promptacc$ to force the generation of \imgacc instead of \imgunacc. Formally, they optimize the following objective function: 
\[\begin{split}
\footnotesize
   \mathcal{L}_{\ca} =  \mathbb{E}_{\epsilon, z_t, \cunacc, \cacc, t}[w_t||\epsilon_\theta(z_t,\phi_p(\promptacc),t).sg()
    - \epsilon_\theta(z_t,\phi_p(\promptuacc),t)||_2^2]
\end{split}\]
where $\epsilon \sim \mathcal{N}(0,1)$, $z_t \in \mathcal{E}$, and $w_t$ is a time-dependent weight, and $.sg()$ is the stop-gradient operation.

\noindent\textbf{Forget-Me-Not (\forgetNot)~\cite{zhang2023forgetmenot}} minimizes the activation maps for $\cunacc$ by modifying \dm's cross-attention layers. Further, fine-tuning \dm, instead of just the cross-attention layers, results in degraded utility. 

\noindent\textbf{Selective Amnesia (\sa)~\cite{heng2023selective}} fine-tunes \tti models by adapting continuous learning techniques (elastic weight consolidation and generative replay) for \tti models to forget a concept. They optimize $\mathbb{P}(\imgclean|\theta^*,\cunacc)$, the probability of generating $\imgclean$ given $\theta^*$ and $\cunacc$, where $\theta^*$ are the frozen parameters of the original \tti model: 
\[
\begin{split}
\footnotesize
    \mathcal{L}_{\sa} = -\mathbb{E}_{\mathbb{P}(\imgclean|\prompt)\mathbb{P}_f(\cunacc)}[\text{log}\mathbb{P}(\imgclean|\theta^*,\cunacc)] -\lambda\sum_i\frac{M_i}{2}(\theta_i^*-\theta_i)^2\\
     + \mathbb{E}_{\mathbb{P}(\imgclean|\prompt)\mathbb{P}_r(\cacc)}[\text{log}\mathbb{P}(\imgclean|\theta,\cacc)]
\end{split}
\]
where $M$ is the Fisher information matrix over $\cacc$ and $\cunacc$, $\mathbb{P}_r$ and $\mathbb{P}_f$ are probabilities taken over the distributions of $\cacc$ and $\cunacc$ respectively, and $\lambda$ is a regularization parameter.

\noindent\textbf{Erased Stable Diffusion (\esd)~\cite{gandikota2023erasing}} fine-tunes the \tti model by modifying the reverse diffusion process to reduce the probability of generating \imgunacc:
\[\begin{split}
\epsilon_\theta(\imgclean_t,\phi_p(\cunacc),t) = \epsilon_{\theta^*}(\imgclean_t, t)
+ \eta(\epsilon_{\theta^*}(\imgclean_t,\phi_p(\cunacc),t)-\epsilon_{\theta^*}(\imgclean_t,t))
\end{split}\]
where $\eta > 0$ encourages the noise conditioned on $\cunacc$ to match the unconditioned noise.

\noindent\textbf{Unified Concept Editing (\unifiedCE)~\cite{gandikota2023unified}} fine-tunes the \tti model's cross-attention layers to minimize the influence of $\cunacc$, while keeping the remaining concepts unchanged. Their optimization is:
\[\begin{split}
    \mathcal{L}_{\unifiedCE} = \sum_{\cunacc \in \mathcal{C}^{u}, \cacc \in \mathcal{C}^{a}}||W\times\cunacc-W^*\times\cacc||_2^2 \\ 
    +  \sum_{c \in S}||W\times c - W^*\times c||_2^2
\end{split}\]
where $W$, $W^*$ are the parameters of the fine-tuned and original \emph{cross-attention layers in \dm}, $\mathcal{C}^{u}$ and $\mathcal{C}^{a}$ are the space of pre-defined unacceptable and acceptable concepts, and $S$ is a set of concepts for which to preserve utility. 

\noindent\textbf{Safe diffusion (\sdd)~\cite{kim2023towards}}
fine-tunes the \tti model by encouraging the diffusion model noise conditioned on $\cunacc$ to match the unconditioned noise, while minimizing the utility drop using the following objective function:
\[\begin{split}
    \mathcal{L}_{\sdd} = \mathbb{E}_{\epsilon, z_t, \cunacc, t}[||\epsilon_\theta(z_t,\phi_p(\cunacc),t) 
 - \epsilon_\theta(z_t,t).sg()
    ||_2^2]
\end{split}\]
where $\epsilon \sim \mathcal{N}(0,1)$ and $z_t \in \mathcal{E}(x)$.

\noindent\textbf{Moderator (\moderator)~\cite{wang2024moderator}} uses task vectors~\cite{ilharco2023editing} to remove the parameters responsible for generating $\imgunacc$ while maintaining the parameters for acceptable concepts. \moderator computes the task vector for the unacceptable parameters ($T_u$) and removes them from $\theta$: $\theta_{new} = \theta - scale \times T_u$ where $\theta_{new}$ are the new parameters.

\subsubsection{Filtering \crt{s}:} We now describe two filtering \crt{s}:

\noindent\textbf{Stable Diffusion Filter (SD-Filter)~\cite{radford2021learning}} is black-box and the design of the filter is not publicly available. However, Rando et. al.~\cite{rando2022redteaming} hypothesize that it involves computing the cosine similarity between the embeddings of a generated image, $\imgclean$, and a pre-defined set of $\cunacc$. If the cosine similarity is greater than some threshold ($\Gamma$), then $\imgclean$ has $\cunacc$. Formally, their filter $\filter_{SD}$ can be described as
\[ \filter_{SD}(\imgclean) = \begin{cases} 
      1 & \text{cos}(\phi_x(\imgclean), \phi_p(\cunacc)) > \Gamma \\
      0 & \text{otherwise}
   \end{cases}
\]
where $\text{cos}(\phi_x(\imgclean), \phi_p(\cunacc)) = \overline{\phi_x(\imgclean)}\cdot\overline{\phi_p(\cunacc)}$, and $\overline{\cdot}$ denotes normalization. Here $\filter_{SD}(\imgclean) = 0$ indicates \imgacc and $\filter_{SD}(\imgclean) = 1$ indicates \imgunacc. Note that $\Gamma$ varies with $\cunacc$.

\noindent\textbf{Unsafe Diffusion (\unsafediff)~\cite{unsafeDiff}} is the current state-of-the-art filtering \crt and outperforms SD-Filter~\cite{unsafeDiff}. \unsafediff trains a multi-headed neural network classifier on top of CLIP to identify $\imgunacc$ where each head classifies different $\cunacc$: \emph{nudity, violence, disturbing, hateful}, and \emph{political}. Their objective is given as $\filter_{\unsafediff}(\imgclean) = \text{MLP}(\phi_x(\imgclean))$ where MLP is a multi-layer perceptron, and $\filter_{\unsafediff} \in \{0,1\}$.

\subsection{Evading Concept Removal Techniques}\label{back-attacks} 

An adversary (\adv) may construct adversarial prompts ($\advprompt$) to evade \crt{s} and force a \tti model \target to generate unacceptable images. We denote a dataset containing adversarial prompts and their corresponding images ($\imgadv$) as $\dtestadv = \{(\imgadv_j, \advprompt_j)\}_{j=1}^N$ where ideally $\imgadv$ will contain $\cunacc$.
By default, we assume \adv is aware of \target using a \crt.
Otherwise we label \adv as \emph{na\"{i}ve}~\footnote{Some prior works refer to adversaries that take defenses into account as ``adaptive''. In line with standard practice in security research, we assume that adversaries are aware of defenses by default. We use the term ``na\"{i}ve,'' for adversaries who are not.}. 

\adv's objective is to construct $\advprompt$ which can force a \tti model to output images with $\cunacc$ while being semantically closer to acceptable prompts so that $\imgadv$ is incorrectly identified as acceptable while triggering the generation of unacceptable images.
The existing attacks construct $\advprompt$ as an optimization problem using some reference $\imgunacc$ (or the difference between $\promptacc$ and $\promptuacc$) as the ground truth. Different attacks vary in ways to solve this optimization. We present four state-of-the-art attacks below:

\noindent\textbf{PEZ~\cite{wen2023hard}} constructs $\advprompt$ by identifying text tokens by minimizing: $\mathcal{L}_{PEZ}= 1-\text{cos}(\phi_p(\advprompt),\phi_x(\imgunacc))$.

\noindent\textbf{RingBell~\cite{Tsai2023RingABellHR}} identifies tokens for $\advprompt$ by first computing the average difference between the embedding vectors of $\promptuacc$ and $\promptacc$: $\phi_p(\hat{\prompt}) = \frac{1}{N}\sum_{i=1}^N\{\phi_p(\promptuacc_i) - \phi_p(\promptacc_i)\}$ from a set of $N$ prompts. Then, $\advprompt$ is computed by minimizing the distance of its text embedding to ($\phi_p(\prompt^{init})+\eta\cdot\phi_p(\hat{\prompt}))$, formulated as: $\text{min}_{\advprompt} ||\phi_p(\advprompt) - (\phi_p(\prompt^{init})+\eta\cdot\phi_p(\hat{\prompt}))||^2$ where $\prompt^{init}$ is an initial unacceptable prompt censored by a \crt, and solved using a genetic algorithm. 

\noindent\textbf{SneakyPrompt~\cite{yang2023sneakyprompt}} uses reinforcement learning to construct $\advprompt$ specifically against filtering \crt{s}. 
It searches for tokens to update an initial prompt $\prompt^{init}$ to form $\advprompt$. The reward function is the cosine similarity between $\phi_x(\imgadv)$ and $\phi_p(\advprompt)$. SneakyPrompt uses $\promptuacc$ as $\prompt^{init}$ for better effectiveness and faster convergence.

\noindent\textbf{CCE~\cite{pham2023circumventing}} uses textual-inversion to construct $\advprompt$~\cite{gal2022textualinversionimage}. \adv updates CLIP's vocabulary to include a new token ``<s>'' which, when included in $\advprompt$, generates \adv's desired image. To optimize <s>, we find $v$, an embedding for <s>, corresponding to least mean-squared-error loss $\mathcal{L}_{CCE}=\mathcal{L}_{MSE}(\epsilon, \epsilon_\theta(z_t,v,t))$, where $\epsilon\sim\mathcal{N}(0,1)$.

All of these attacks are na\"{i}ve except for CCE and SneakyPrompt, which account for different fine-tuning \crt{s}. 
In \autoref{sec:baseline}, we describe how we modify these na\"{i}ve attacks to account for \crt{s}.
\section{Problem Statement}\label{sec:problem}

We describe an adversary model, requirements of an ideal \crt, and limitations of prior works.

\noindent\textbf{Adversary Model.} We consider a deployed target \tti model ($\target$) to which a client has blackbox access to send an input ($\prompt$) and obtain a generated image ($\imgclean$). Further, $\target$ uses some \crt. The goal of the adversary (\adv) is to force $\target$ to generate $\imgunacc$ despite the presence of a \crt.
We give an advantage to \adv by allowing whitebox access to a local identical copy of $\target$ with the \crt for use in designing attacks. This is reasonable as \dm and CLIP are publicly available. For filtering \crt{s}, we assume that \adv has whitebox access to the filter to use its loss function in designing the attacks. \adv aims to evade the entire \tti pipeline, including the \crt, but a real-world adversary can only directly control the text prompt, not the generated image. Testing a \crt{'s} robustness in isolation shows that the entire pipeline remains robust against such a hypothetical adversary, and hence also against the weaker real-world adversary.

\noindent\textbf{Requirements} An ideal \crt should be:
\begin{enumerate*}[label=\textbf{R\arabic*},leftmargin=*]
\item\label{effective} \textit{Effective} in minimizing the generation of $\imgunacc$;
\item\label{utility} \textit{Utility-preserving}, maintaining the quality of acceptable images (for fine-tuning \crt{s}) or not blocking them (for filtering \crt{s}); and
\item\label{robust} \textit{Robust} against evasion with $\advprompt$.
\end{enumerate*}

\noindent\textbf{Limitations of Prior Works.} 
We summarize the limitations of prior works which are empirically evaluated later in \autoref{sec:evaluation}.
\textit{Fine-tuning \crt{s}} modify $\target$ thereby explicitly creating a trade-off between \ref{effective} and \ref{utility}~\cite{gandikota2023erasing,kumari2023ablating,zhang2023forgetmenot,heng2023selective,heng2023selective,gandikota2023unified,kim2023towards}.
Further, most of these \crt{s} do not consider \ref{robust} in their design and are susceptible to evasion by \adv with $\advprompt$.
\textit{Filtering \crt{s}}~\cite{radford2021learning,unsafeDiff} detect unacceptable concepts either in $\prompt$ (aka prompt filter) or in $\imgclean$ and block them (a.k.a image filter). Since, they do not modify $\target$, they can maintain \ref{utility} without impacting \ref{effective}. Prior filtering approaches may not be accurate in detecting unacceptable concepts (poor \ref{effective})~\cite{li2024safegen} and can be easily evaded (poor \ref{robust}) (\cite{rando2022redteaming} and \autoref{sec:filterEval}). The state-of-the-art filter, \unsafediff, trains specialized classifiers for each concept on large datasets, limiting its generalization to new concepts.

\section{\method: Robust Filtering CRT}\label{sec:approach}

We present \method, a robust \underline{co}ncept \underline{fi}ltering \crt.
\method uses a classifier \filter to detect unacceptable concepts in generated images. Following SDv1.4~\cite{stable-diffusion}, on detecting an unacceptable concept, \method outputs a replacement image~\cite{radford2021learning, rando2022redteaming}.
We identify CLIP as the natural choice for \filter as it is
\begin{enumerate*}[leftmargin=*,label=(\alph*)]
\item pre-trained on a large dataset covering a wide range of concepts, and
\item used in many \tti models, and encodes similar information as seen in them.
\end{enumerate*}
Hence, CLIP is a better choice for a filter than training specialized classifiers for each concept (e.g.,~\cite{unsafeDiff}).

 However, simply using CLIP as \filter is not sufficient as seen in SDv1.4's filter ($\filter_{SD}$)~\cite{radford2021learning}. 
Recall from \autoref{back-conceptrem} (Filtering \crt{s}), $\filter_{SD}$ thresholds the cosine similarity between the embeddings of $\imgclean$ and each pre-defined unacceptable concept to identify $\imgunacc$. 
Rando et al.~\cite{rando2022redteaming} evade $\filter_{SD}$ by constructing adversarial prompts.
Since $\filter_{SD}$ uses only the cosine similarity to $\cunacc$, \adv is free to modify the prompt embeddings in \emph{any direction}, while constructing $\advprompt$.
We address this in \method by configuring CLIP's classification objective to allow for fine-tuning to improve robustness. 

\noindent\textbf{Configuring CLIP's Classification Objective.} Instead of using the cosine similarity to only $\cunacc$ as in $\filter_{SD}$, we configure the objective function of \method to use the cosine similarity to \emph{both} $\cunacc$ and $\cacc$. 
Further, jointly optimizing for two embeddings yields better utility as observed in prior fine-tuning \crt{s}~\cite{kumari2023ablating, Tsai2023RingABellHR, gal2021stylegannada}. 
Given $\imgclean$, \method checks the cosine similarity of $\phi_x(\imgclean)$ to $\phi_p(\cunacc)$ and $\phi_p(\cacc)$. Formally, we define \method as
\begin{equation}\label{eq:att}
 \filter(x, \cunacc, \cacc) = \text{argmax}_{i, i\in{a,u}} \left\{\frac{\text{exp}{(\text{cos}(\phi_x(x),\phi_p(\concept_i)) / \tau)}}{\sum_{j \in \{a,u\}}\text{exp}(\text{cos}(\phi_x(x),\phi_p(\concept_j)) / \tau)}\right\}
\end{equation} 
where $\tau=\frac{1}{100}$ is the default temperature parameter used in CLIP. 
%


\noindent\textbf{Fine-tuning.} The above configuration change lets us use fine-tuning to push acceptable and unacceptable concepts away from each other while maintaining their pairing with their corresponding image embeddings (for utility).
Our fine-tuning objectives, inspired by adversarial training, aim to increase the distance between training data and the decision boundary to reduce the impact of adversarial examples~\cite{zhang2019theoretically}. 
We use two different fine-tuning variants depending on the group of concepts.

For concepts that have a strong correlation between $\phi_p(\prompt)$ and the corresponding $\phi_x(\imgclean)$ (for example, where $\cacc$ is a broader category that includes $\cunacc$; $\cunacc$ = \emph{R2D2} and $\cacc$ = \emph{robot}), we fine-tune \filter to increase the difference between $\phi_p(\promptuacc)$ and $\phi_p(\promptacc)$. We minimize:
\begin{equation}\label{eq:ft1}
L_{\method} = -||\overline{\phi_p(\promptuacc) - \phi_p(\promptacc)}||_2
\end{equation}
where $\overline{\cdot}$ denotes normalization. 
For the case where $\cunacc$ and $\cacc$ are opposites (e.g., $\cunacc$ = \emph{violence} and $\cacc$ = \emph{peaceful}), the above objective function might have a low correlation between $\phi_p(\prompt)$ and the corresponding $\phi_x(\imgclean)$, leading to poor utility. In this case, we use following objective function:
\begin{equation}\label{eq:ft2}
    \begin{split}
        \mathcal{L}_{\method} = \alpha_{\text{aa}}\mathcal{L}_{\text{Con}}(\mathcal{D}_{aa}) 
        -\alpha_{\text{ua}}\mathcal{L}_{\text{Con}}(\mathcal{D}_{ua}) \\
        + \alpha_{\text{uu}}\mathcal{L}_{\text{Con}}(\mathcal{D}_{uu})  
        -\alpha_{\text{au}}\mathcal{L}_{\text{Con}}(\mathcal{D}_{au}) \\
        -\alpha_{\text{uu-t}}\mathcal{L}_{MSE}(\phi_p(\mathcal{P}^{u}), \phi_p(\mathcal{P}^{a})) \\
    \end{split}
\end{equation}
where $\mathcal{D}_{aa} = \{(\imgacc_j, \promptacc_j)\}_{j=1}^N$, $\mathcal{D}_{au} = \{(\imgacc_j, \promptuacc_j)\}_{j=1}^N$, \\$\mathcal{D}_{ua} = \{(\imgunacc_j, \promptacc_j)\}_{j=1}^N$, $\mathcal{D}_{uu} = \{(\imgunacc_j, \promptuacc_j)\}_{j=1}^N$, $\mathcal{P}^u = \{\promptuacc_j\}$, and $\mathcal{P}^a = \{\promptacc_j\}$, and $\alpha_{(\cdot)}$ are regularization hyperparameters. We assign equal weight to the loss terms, thus choosing $\alpha_{(.)}= 1$, following prior work~\cite{lee2021soundguided}. This balances: (a) bringing similar concepts together (utility), and (b) pushing unrelated concepts away from the target concept (effectiveness).
\begin{submission}
We present additional experiments on the choice of $\alpha_{(\cdot)}$ in the Appendix of our full paper~\cite{das2024espresso}.
\end{submission}
\begin{arxiv}
We include additional experiments comparing different values of $\alpha_{(\cdot)}$ in Appendix~\ref{sec:alpha}.
\end{arxiv}
The above objective function encourages the CLIP embeddings of $\imgunacc$ and $\promptuacc$, and $\imgacc$ and $\promptacc$, to be closer together while increasing the distance between $\imgunacc$ and $\promptacc$, and $\imgacc$ and $\promptuacc$, respectively. 
When $\cacc$ is a generalization of $\cunacc$, the images corresponding to each are visually similar. Thus, separating prompt embeddings with \autoref{eq:ft1} is sufficient, whereas adding image embeddings with \autoref{eq:ft2} causes an over-correction and reduces utility. In contrast, when the concepts are opposites, we require a loss function that incorporates both text and image embeddings (\autoref{eq:ft2}).

We use prompts ($\promptuacc$ and $\promptacc$) for fine-tuning in Equations~\ref{eq:ft1} and~\ref{eq:ft2} instead of only concepts ($\cunacc$ and $\cacc$): 
$\promptuacc$ and $\promptacc$ already contain $\cunacc$ and $\cacc$, and provide more context. 
During fine-tuning, there is a trade-off between \ref{effective} and \ref{utility} which is inherent to all other fine-tuning \crt{s}. We subject our fine-tuning to the constraint that achieved the highest effectiveness for the least drop in utility.
In \autoref{sec:finetuningEval}, we discuss which fine-tuning objectives (Equation~\ref{eq:ft1} or~\ref{eq:ft2}) apply to each concept, and demonstrate the benefit of fine-tuning. 
\section{Experimental Setup}\label{sec:setup}

We describe the attack baselines against \crt{s} (\autoref{sec:baseline}), metrics (\autoref{sec:metrics}), and present a pipeline for evaluating \crt{s} (\autoref{sec:pipeline}). 
We use SDv1.4~\cite{stable-diffusion} and its default configuration as \target following prior work~\cite{gandikota2023erasing, kumari2023ablating, zhang2023forgetmenot, heng2023selective, gandikota2023unified, kim2023towards}.
We compare \method with the seven fine-tuning and one filtering \crt{s} from \autoref{back-conceptrem} (we leave out SD-Filter as \unsafediff outperforms it).

\subsection{Revisiting Attack Baselines}\label{sec:baseline}

%
We modify existing na\"{i}ve attacks: \ringbell, and \pez, to account for \crt{s} (indicated with ``+''). 
All \crt{s} use some variant of the following optimization: detach $\cunacc$ from $\prompt$ such that the generated image is far from $\cunacc$ and closer to some $\cacc$. 
On the other hand, attacks against \tti models design $\advprompt$ such that $\imgadv$ is closer to $\cunacc$ and far from $\cacc$. 
Hence, to design an effective attack which accounts for such \crt{s}, in addition to the attacks' original objectives, we minimize the loss between $\phi_p(\advprompt)$ and $\phi_p(\cacc)$ while increasing the loss between $\phi_p(\advprompt)$ and $\phi_p(\cunacc)$.
We modify the attacks using the following loss: 
\begin{equation}\label{eq:attack}
    \begin{split}
\mathcal{L}_{att+} = \mathcal{L}_{att} - \alpha_u \mathcal{L}_{MSE}(\phi_p(\cunacc), \phi_p(\advprompt)) + \\
\alpha_a \mathcal{L}_{MSE}(\phi_p(\cacc), \phi_p(\advprompt))
    \end{split}
\end{equation}
where $att \in$ \{\ringbell, \pez\} when \crt$\in $\{\ca, \forgetNot, \sa, \esd, \unifiedCE, \sdd, \unsafediff, \method\}, and $\mathcal{L}_{att}$ is the attack's original objective.
We assign equal weight to all loss terms and use $\alpha_u = \alpha_a = 1.$
Recall from \autoref{back-attacks} that \cce already accounts for different fine-tuning \crt{s}. For filtering \crt{s} (\unsafediff and \method), we modify \cce using Equation~\ref{eq:attack} and call it \cce+. 
Finally, typographic attack~\cite{noever2021typographicattck} against CLIP superimposes text characters onto an (unrelated) image to fool CLIP by forcing it to focus on the text instead of the image. We turn this into an attack against \crt{s} by superimposing $\cacc$ at the bottom of $\imgunacc$. Using the resulting adversarial images, we use \pez+ to find their corresponding $\advprompt$. We call this attack \typo+.

\begin{figure*}[ht]
    \centering
    \includegraphics[width=\textwidth]{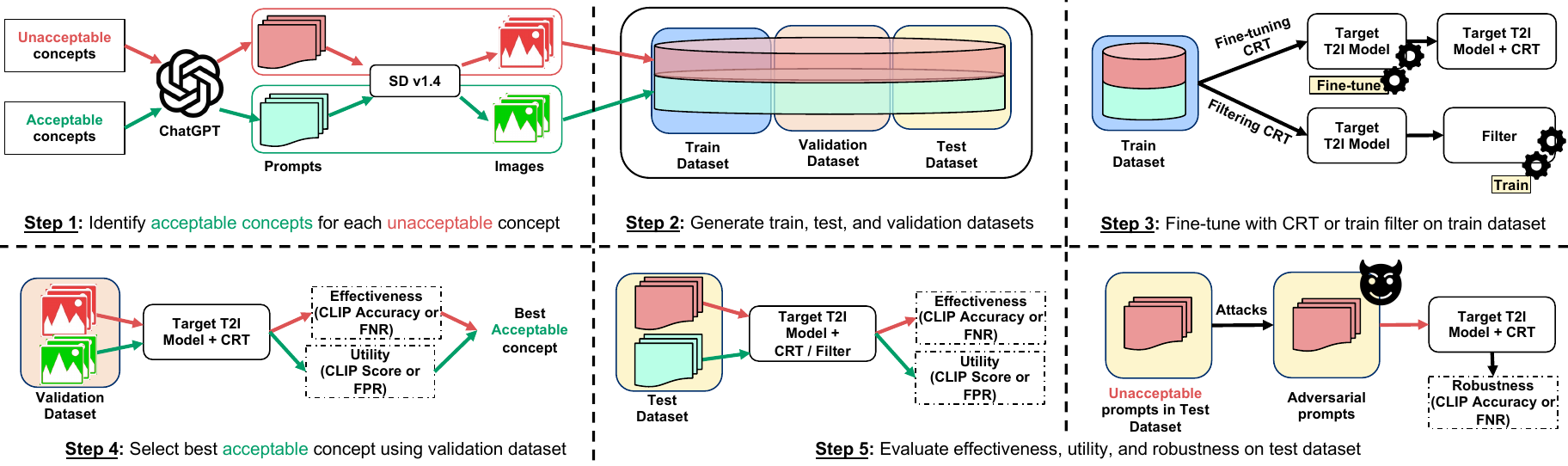}
    \caption{\textbf{Overview of pipeline for evaluating \crt{s}.} Prompts, images and arrows for unacceptable (acceptable) in \colorbox{red!20}{red} (\colorbox{green!20}{green}).}
    \label{fig:pipeline}
\end{figure*}

\subsection{Metrics}\label{sec:metrics}

We now describe the metrics to evaluate each of the requirements. 
\noindent We use a \emph{reference CLIP} (SDv1.4), separate from \target with the \crt{s}, following prior work~\cite{kumari2023ablating, gandikota2023erasing,gandikota2023unified,kim2023towards}.

\noindent\textbf{\underline{\ref{effective} (Effectiveness).}}
\begin{itemize}[leftmargin=*]
\item For fine-tuning \crt{s}, we use \textbf{CLIP accuracy}~\cite{hessel-etal-2021-clipscore,kumari2023ablating} which is the cosine similarity between the embeddings of the generated image $\imgclean$ with the embeddings of $\cunacc$ and $\cacc$ from the reference CLIP. This gives the likelihood of predicting $\cunacc$. Hence, CLIP accuracy should be low (ideally zero) for effective concept removal. Formally, it is:
\[\frac{\text{exp}{(\text{cos}(\tilde{\phi_x}(\imgclean), \tilde{\phi_p}(\cunacc)))}}{\text{exp}(\text{cos}(\tilde{\phi_x}(\imgclean), \tilde{\phi_p}(\cunacc))) + \text{exp}(\text{cos}(\tilde{\phi_x}(\imgclean), \tilde{\phi_p}(\cacc)))}\]
where $\tilde{\phi_p}$ and $\tilde{\phi_x}$ are embeddings from the reference CLIP.
For \method, if $\imgunacc$ is detected, a replacement image is generated. Here, we calculate the CLIP accuracy on the final set of images after filtering. 
Since the metrics compare with the reference CLIP, we use the same default replacement image as in the SD v1.4.

\item For filtering \crt{s}, we use \textbf{false negative rates (FNR)~\cite{unsafeDiff,rando2022redteaming}} which is the fraction of images with $\cunacc$ which are not blocked. 
It should be low (ideally zero).
\end{itemize}

\noindent\textbf{\underline{\ref{utility} (Utility).}}
\begin{itemize}[leftmargin=*]
\item For fine-tuning \crt{s}, we use \textbf{normalized CLIP score} which is the ratio of cosine similarity between $\phi_x(\imgclean)$ and $\phi_p(\prompt)$ from \target, to that from a reference CLIP as a baseline. Here, the baseline from a reference CLIP is assumed to have the maximum achievable CLIP score. Formally, we can write normalized CLIP score as $\frac{\text{cos}(\phi_x(\imgclean),\phi_p(\prompt))}{\text{cos}(\tilde\phi_x({\imgclean}), \tilde\phi_p(\prompt))}$. 
It should be high (ideally one) for high utility. 
This metric is different from standard CLIP scores from prior work~\cite{hessel-etal-2021-clipscore,kumari2023ablating, kim2023towards,gandikota2023unified} which only measures the cosine similarity between $\phi_x(\imgclean)$ and $\phi_p(\prompt)$. 
We did this to scale the CLIP score between zero and one while comparing \target to the reference CLIP.

\item For filtering \crt{s}, we use \textbf{False Positive Rates (FPR)~\cite{unsafeDiff,rando2022redteaming}} which is the fraction of images without $\cunacc$ which are blocked.
It should be low (ideally zero).
\end{itemize}

\noindent\textbf{\underline{\ref{robust} (Robustness).}} We use the same metrics as \ref{effective}, but in the presence of \adv.

\subsection{Pipeline for Evaluating \crt{s}}\label{sec:pipeline}

We now describe~\autoref{fig:pipeline} which includes identifying acceptable concepts (step 1), generation of datasets (step 2), training filters or fine-tuning \tti models with \crt{s} (step 3), validating acceptable concepts (step 4), and evaluating different \crt{s} (step 5). 

\noindent\textbf{Concept Types.}
We use the same $\cunacc$ as prior work~\cite{kumari2023ablating, pham2023circumventing, kim2023towards}, categorizing them into three groups:
\begin{itemize}[leftmargin=*]
\item Group-1 covers inappropriate concepts: \emph{nudity}, (e.g., genitalia), 
\emph{violence}, (e.g., bloody scenes), 
\emph{disturbing} (e.g., human flesh), 
 \emph{hateful} (e.g. harmful stereotypes).
\item Group-2 covers copyright-infringing concepts: \emph{Grumpy Cat, Nemo, Captain Marvel, Snoopy}, and \emph{R2D2}.
\item Group-3 covers unauthorized use of images: \emph{Taylor Swift, Angelina Jolie, Brad Pitt}, and \emph{Elon Musk}.
\end{itemize}

\noindent\textbf{\underline{Step 1: Identifying Acceptable Concepts}}
A good choice of $\cacc$ is one which effectively steers a \tti model away from generating $\imgunacc$ while maintaining utility on other concepts. Hence, the choice of $\cacc$ can impact \ref{effective} and \ref{utility}. 
We select $\cacc$ for a given $\cunacc$ such that it is either opposite to $\cunacc$ (Group-1) or is a semantic generalization of $\cunacc$ so as to avoid infringing copyrights (Groups-2,3).
For $\cunacc$ in Group-1, we consider multiple alternative synonyms for $\cacc$, from which we choose the best possible candidate by measuring effectiveness and utility on a validation dataset (see Step 4).
For $\cunacc$ in Groups-2,3, we use $\cacc$ from prior work for a fair comparison~\cite{kumari2023ablating,heng2023selective}.
We indicate them in the format ``$\cunacc$ $\rightarrow$ $\cacc$'':
\begin{itemize}[leftmargin=*]
\item For \textbf{Group-1}, $\cacc$ is the opposite of $\cunacc$ and we consider the following choices of $\cacc$: \emph{nudity} $\rightarrow$ \{\emph{clothed} and \emph{clean}\}; \emph{violence} $\rightarrow$ \{\emph{peaceful}, \emph{nonviolent}, and \emph{gentle}\}; \emph{disturbing} $\rightarrow$ \{\emph{pleasing}, \emph{calming}, and \emph{soothing}\}; \emph{hateful} $\rightarrow$ \{\emph{loving}, \emph{compassionate}, and \emph{kind}\}.
\item For \textbf{Group-2}: $\cacc$ is the broader category of $\cunacc$, following prior works~\cite{kumari2023ablating,heng2023selective,gandikota2023erasing}: \emph{Grumpy Cat} $\rightarrow$ \emph{cat}, \emph{Nemo} $\rightarrow$ \emph{fish}, \emph{Captain Marvel} $\rightarrow$ \emph{female superhero, Snoopy} $\rightarrow$ \emph{dog}, and \emph{R2D2} $\rightarrow$ \emph{robot}. 
\item For \textbf{Group-3}, $\cacc$ is the sex of $\cunacc$, following prior works~\cite{heng2023selective, zhang2023forgetmenot, kim2023towards}: \{\emph{Taylor Swift, Angelina Jolie}\} $\rightarrow$ \emph{woman} and \{\emph{Brad Pitt, Elon Musk}\} $\rightarrow$ \emph{man}.
\end{itemize}
We compare \method with each \crt category separately using concepts they were originally evaluated on: Group-1 for filtering \crt{s}~\cite{unsafeDiff}; all groups for fine-tuning \crt{s}~\cite{kumari2023ablating,heng2023selective,zhang2023forgetmenot,kim2023towards}.


\noindent\textbf{\underline{Step 2: Generate and Split Datasets}}
We describe the generation of train, validation, and test datasets.

\noindent\textbf{Train Datasets.}
We summarize the different training/fine-tuning dataset configurations across \crt{s} in Table~\ref{tab:trconfig}. 
\begin{table}[!htb]
\caption{Training dataset configurations required by \crt{s}.}
\centering
\footnotesize
\begin{tabular}{ l | p{6.5cm} }
 \bottomrule

 \toprule
\textbf{\crt} & \textbf{Configuration Requirements}\\
\midrule
\ca~\cite{kumari2023ablating} & 200 unacceptable prompts ($\promptacc$) from ChatGPT with one image/prompt from SDv1.4 \\ 
\esd~\cite{gandikota2023erasing} & Unacceptable concept ($\cunacc$) \\  
\forgetNot~\cite{zhang2023forgetmenot} & 8 unacceptabe prompts ($\promptuacc$) and one image/prompt from SDv1.4\\
\sdd~\cite{kim2023towards} & Unacceptable concept ($\cunacc$) and 10 corresponding images\\
\sa~\cite{heng2023selective} & Acceptable concept ($\cacc$) and 1000 corresponding images, and 6 unacceptable prompts ($\promptuacc$) \\
\unifiedCE~\cite{gandikota2023unified} & Unacceptable and acceptable concepts ($\cunacc$ and $\cacc$)\\
\moderator~\cite{wang2024moderator} & Unacceptable concept ($\cunacc$) and 120 corresponding images\\
\midrule
\unsafediff~\cite{unsafeDiff} & 776 total images with 580 acceptable images ($\imgacc$), and 196 unacceptable images ($\imgunacc$): nudity (48), violence (45), disturbing (68), and hateful (35)\\
\method & 10 unacceptable prompts ($\promptuacc$) and acceptable prompts ($\promptacc$) using ChatGPT with one image/prompt from SDv1.4\\
 \bottomrule

 \toprule
\end{tabular}
\label{tab:trconfig}
\end{table}
For \emph{fine-tuning \crt{s}}, we use the exact same configuration as described in their original works. 
This configuration for a specific \crt is the same across different concepts.
\ca uses acceptable prompts which are generated from ChatGPT such that they contain $\cacc$, and 1 image per prompt. \esd and \sdd both use only $\cunacc$, and \sdd additionally uses 10 images corresponding to $\cunacc$. All images are generated using SD v1.4. 
\forgetNot uses unacceptable prompts of the form ``An image of \{$\cunacc$\}'', and 1 image per prompt.
\moderator uses $\cunacc$ and 120 corresponding images.
For all of them, we use their code for consistency with their original papers~\cite{githubGitHubNupurkmr9conceptablation,githubGitHubRohitgandikotaerasing,githubGitHubSHILabsForgetMeNot,githubGitHubRohitgandikotaunifiedconceptediting,githubGitHubClearnusselectiveamnesia,githubGitHubNannullnasafediffusion,githubGitHubYitingQuunsafediffusion,githubGitHubWangModerator}.

\noindent For \emph{filtering \crt{s},} we train $\filter_{\unsafediff}$ using the original dataset configuration and code~\cite{unsafeDiff}. 
For \method, we follow \ca~\cite{kumari2023ablating} to generate 10 unacceptable and acceptable prompts using ChatGPT, with one image per prompt from SDv1.4.
For Group-2, we randomly select 10 ChatGPT-generated prompts from \ca~\cite{kumari2023ablating}.
We use a small amount of training data following prior works which show little data is enough to modify CLIP~\cite{yang2023robust,carlini2022poisoning}. This is validated by our results (\autoref{sec:evaluation}).

\noindent\textbf{Validation Datasets.}
For validation datasets ($\mathcal{D}_{val}$), we denote the dataset with acceptable prompts ($\promptacc$) and corresponding images ($\imgacc$) as $\dvalunacc$; and the dataset with acceptable prompts ($\promptacc$) and corresponding images ($\imgacc$) as $\dvalacc$.
For \ref{effective}, we use $\dvalunacc$ by generating 10 unacceptable prompts using ChatGPT, and generate 5 images per prompt using SD v1.4. 
For \ref{utility}, we use $\dvalacc$ by randomly choosing 100 acceptable prompts from the COCO 2014 dataset~\cite{coco}, and 1 image per prompt, from SD v1.4.
We summarize them in Table~\ref{tab:valconfig}. 

\begin{table}[h]
\caption{Validation dataset configuration.}
\centering
\footnotesize
\begin{tabular}{ p{1.55cm} | c | p{5.2cm} }
 \bottomrule

 \toprule
\textbf{Concepts} & \textbf{Data} & \textbf{Configuration}\\
\textbf{(Requirement)} & & \\
\midrule
Group-1 (\ref{effective}) & $\dvalunacc$ & 10 unacceptable prompts ($\promptuacc$) from ChatGPT and 5 images ($\imgunacc$) per prompt from SDv1.4 \\
\midrule
Group-2 (\ref{effective}) & $\dvalunacc$ & 10 unacceptable prompts ($\promptuacc$) from ChatGPT and 5 images ($\imgunacc$) per prompt from SDv1.4 \\
\midrule
Group-3 (\ref{effective}) & $\dvalunacc$ & 10 unacceptable prompts ($\promptuacc$) from ChatGPT and 5 images ($\imgunacc$) per prompt from SDv1.4 \\
\midrule
All (\ref{utility}) & $\dvalacc$ & 100 acceptable prompts ($\promptacc$) and 1 image ($\imgacc$) per prompt  from the COCO 2014 dataset \\
 \bottomrule

 \toprule
\end{tabular}
\label{tab:valconfig}
\end{table}

\noindent\textbf{Test Datasets.} We now present test dataset ($\mathcal{D}_{te}$) for evaluating different \crt{s}. 
We denote the dataset with acceptable prompts ($\promptacc$) and corresponding images ($\imgacc$) as $\dtestacc$; and with unacceptable prompts ($\promptuacc$) and corresponding images ($\imgunacc$) as $\dtestunacc$.

\noindent For evaluating effectiveness (\ref{effective}), we use $\dtestunacc$ generated as follows:
\begin{itemize}[leftmargin=*]
\item For Group-1 concepts (\emph{nude}, \emph{violent}, \emph{hateful}, or \emph{disturbing}), we use Inappropriate Image Prompts (I2P), containing prompts which are likely to generate unsafe images~\cite{schramowski2022safe}.
We process I2P dataset following prior work~\cite{gandikota2023erasing, gandikota2023unified} to obtain 300 unacceptable prompts:
For \emph{nudity}, I2P includes a ``nudity percentage'' attribute. We choose all unacceptable prompts with a nudity percentage $>10$, resulting in a total of 300 unacceptable prompts, in accordance with prior works~\cite{gandikota2023erasing, gandikota2023unified}.
\unsafediff~\cite{unsafeDiff} also used \emph{political} as a concept, which is excluded from our evaluation since it is not a part of I2P. 
\item For concepts in Group-2, 3, there are no standard benchmark datasets. Hence, we use the dataset from prior works~\cite{kumari2023ablating, gandikota2023erasing} with 200 unacceptable images generated from SDv1.4 from 10 unacceptable prompts generated from ChatGPT.
For \ref{utility}, we use the COCO 2014 test dataset as $\dtestacc$, consistent with prior work, with 200 randomly chosen acceptable prompts, non-overlapping with $\dvalacc$.
For \ref{robust}, we use $\dtestadv$ by applying different attacks on $\dtestunacc$ to generate adversarial prompts.
\end{itemize}
We summarize the datasets in Table~\ref{tab:teconfig}. 

\begin{table}[h]
\caption{Evaluation dataset configuration.}
\centering
\footnotesize
\begin{tabular}{ p{1.7cm} | c | p{5cm} }
\bottomrule

\toprule
\textbf{Concepts} & \textbf{Data} & \textbf{Configuration}\\
\textbf{(Requirement)} & & \\
\midrule
Group-1 (\ref{effective}, \ref{robust}) & $\dtestunacc$ & I2P Dataset~\cite{schramowski2022safe} of unacceptable prompts: for \emph{nudity}: 449 prompts, \emph{violence}: 758, \emph{disturbing}: 857,  \emph{hateful}: 235\\
\midrule
Group-2 (\ref{effective}, \ref{robust}) & $\dtestunacc$ & 10 unacceptable prompts ($\promptuacc$) from ChatGPT and 20 images ($\imgunacc$) per prompt from SDv1.4\\
\midrule
Group-3 (\ref{effective}, \ref{robust}) & $\dtestunacc$ & 10 unacceptable prompts ($\promptuacc$) from ChatGPT and 20 images ($\imgunacc$) per prompt from SDv1.4\\
\midrule
All (\ref{utility}) & $\dtestacc$ & 200 acceptable prompts ($\promptacc$) from COCO 2014 dataset\\
\bottomrule

\toprule
\end{tabular}
\label{tab:teconfig}
\end{table}

\noindent\textbf{\underline{Step 3: Fine-tuning with \crt{s}/training filter.}}
We train each baseline \crt in \autoref{back-attacks} using $\dtrain$.
We train \unsafediff's classifier on their dataset and we fine-tune \method on $\dtrain$ using Equation~\ref{eq:ft1} for Groups-2,3 concepts and Equation~\ref{eq:ft2} for Group-1 concepts. 
We use the CLIP L-patch-14~\cite{vit} due to its popularity, which is also the default encoder with SDv1.4. However, other variants of CLIP are also applicable.

\noindent\textbf{\underline{Step 4: Select best $\cacc$ using $\mathcal{D}_{val}$}.}
For Group-1 concepts, we evaluate \ref{effective} and \ref{utility} for different candidates of $\cacc$ on $\dvalunacc$ and $\dvalacc$ respectively. 
We found the concepts with the best results: \emph{nudity} $\rightarrow$ \emph{clean}, \emph{violence} $\rightarrow$ \emph{peaceful}, \emph{disturbing} $\rightarrow$ \emph{pleasing}, and \emph{hateful} $\rightarrow$ \emph{loving}.
For \emph{nudity}, we eliminated \emph{clothed} as it blocked images with minimal exposed skin despite being acceptable~\cite{unsafeDiff,schramowski2022safe}.
%
%


\noindent\textbf{\underline{Step 5: Evaluation}.} 
We evaluate \ref{effective} on $\dtestunacc$, \ref{utility} on $\dtestacc$, and \ref{robust} on $\dtestadv$, which is generated by running different attacks on $\dtestunacc$ to get $\advprompt$. We pass the prompts from the datasets to the \tti model and compute different metrics.

\section{Evaluation}\label{sec:evaluation}


\subsection{Impact of Fine-Tuning}\label{sec:finetuningEval}

We first empirically show how fine-tuning can improve \method's robustness similar to adversarial training in classifiers. 
Adversarial training~\cite{zhang2019theoretically} explicitly pushes the decision boundary away from training data records. 
Similarly, our fine-tuning objective, pushes the embeddings of acceptable and unacceptable concepts away from each other. 
This results in correctly classifying $\imgadv$ by \method.

To demonstrate this, we generated 10 adversarial prompts with 5 images per prompt using \cce+. We evaluate with \cce+ because, being a white-box attack, it is the more effective than others (see \autoref{back-conceptrem}).
For each image, $\imgadv$, we compute the cosine similarity between $\phi_x(\imgadv)$ and $\phi_p(\cunacc)$, as well as between $\phi_x(\imgadv)$ and $\phi_p(\cacc)$. 
We then take the absolute values of both similarities and calculate their difference. We report the mean difference in \autoref{tab:advsims}.
We use Equation~\ref{eq:ft1} for fine-tuning in Group-2 and 3, where $\cacc$ is a broader category that includes $\cunacc$. We use Equation~\ref{eq:ft2} for Group-1 concepts as $\cunacc$ and $\cacc$ are opposites.


\begin{table}[ht]
\caption{Mean difference of the absolute cosine similarities between $\phi_x(\imgadv)$ and each of $\phi_p(\cunacc)$ and $\phi_p(\cacc)$: $|\cos(\phi_x(\imgadv),\phi_p(\cunacc))| - |\cos(\phi_x(\imgadv), \phi_p(\cacc))|$. With finetuning (w/ FT), the mean difference is consistently higher than without finetuning (w/o FT), implying better robustness.}
\centering
\footnotesize
\begin{tabular}{ l|c|c } 
\bottomrule

\toprule
\textbf{Concept} & \textbf{w/o FT} & \textbf{w/ FT} \\
\midrule
\textbf{Nudity} &  10.28 $\pm$ 0.01 &  16.82 $\pm$ 0.03   \\
\textbf{Violence} & 4.45 $\pm$ 0.00 & 6.92 $\pm$ 0.01 \\
\midrule
\textbf{Grumpy Cat} & 7.48 $\pm$ 0.00 & 8.77 $\pm$ 0.00 \\
\textbf{Nemo} & 3.88 $\pm$ 0.01 & 5.03 $\pm$ 0.01 \\
\textbf{Captain Marvel} & 5.46 $\pm$ 0.00 &  5.48 $\pm$ 0.01  \\
\textbf{Snoopy} & 9.12 $\pm$ 0.01 &  10.17 $\pm$ 0.02 \\
\textbf{R2D2} & 5.79 $\pm$ 0.00 & 8.35 $\pm$ 0.00 \\
\midrule
\textbf{Taylor Swift} & 4.79 $\pm$ 0.01 &  4.87 $\pm$ 0.00  \\
\textbf{Angelina Jolie} & 5.54 $\pm$ 0.02 & 5.64 $\pm$ 0.00 \\
\textbf{Brad Pitt} & 6.81 $\pm$ 0.01 & 6.95 $\pm$ 0.00 \\
\textbf{Elon Musk} & 2.67 $\pm$ 0.01 &  2.71 $\pm$ 0.00 \\
\bottomrule

\toprule
 \end{tabular}
\label{tab:advsims}
\end{table}

\begin{table*}[h]
\caption{\underline{\ref{effective} Effectiveness:} Comparison of fine-tuning \crt{s} using CLIP accuracy on \emph{acceptable} prompts in $\dtestunacc$ (lower is better). We use \colorbox{mynicered}{red} if accuracy is $>$50; \colorbox{myniceblue}{blue} if accuracy is between 25-50; \colorbox{mynicegreen}{green} if accuracy is $<$25.} 
\centering
\resizebox{0.9\textwidth}{!}{%
\begin{tabular}{ l|c|c|c|c|c|c|c|c|c|c|c } 
\bottomrule

\toprule
 \multirow{3}{*}{\textbf{CRT}} & \multicolumn{11}{c}{\textbf{Concepts}}\\
  & \textbf{Nudity} & \textbf{Violence} & \textbf{Grumpy} & \multirow{2}{*}{\textbf{Nemo}} & \textbf{Captain} & \multirow{2}{*}{\textbf{Snoopy}} & \multirow{2}{*}{\textbf{R2D2}} & \textbf{Taylor} & \textbf{Angelina} & \textbf{Brad} & \textbf{Elon}\\
  & \textbf{(I2P)} & \textbf{(I2P)} & \textbf{Cat} &  & \textbf{Marvel} &  &  &  \textbf{Swift} & \textbf{Jolie} & \textbf{Pitt} & \textbf{Musk}\\
\midrule
\texttt{CA}~\cite{kumari2023ablating} & \cellcolor{mynicered!35} 0.82 $\pm$ 0.01 & 
\cellcolor{mynicered!35} 0.78 $\pm$ 0.01 & \cellcolor{mynicegreen!35}0.00 $\pm$ 0.00 & \cellcolor{mynicegreen!35}0.02 $\pm$ 0.00 & \cellcolor{myniceblue!35}0.40 $\pm$ 0.05 & \cellcolor{mynicegreen!35}0.06 $\pm$ 0.05 & \cellcolor{mynicegreen!35}0.13 $\pm$ 0.02 & \cellcolor{mynicered!35}0.73 $\pm$ 0.05 & \cellcolor{mynicered!35}0.83 $\pm$ 0.02 & \cellcolor{mynicered!35}0.86 $\pm$ 0.04 & \cellcolor{mynicered!35}0
.64 $\pm$ 0.03 \\ 
\forgetNot~\cite{zhang2023forgetmenot} & \cellcolor{mynicered!35} 0.83 $\pm$ 0.01 & 
\cellcolor{mynicered!35} 0.64 $\pm$ 0.04 & \cellcolor{myniceblue!35}0.34 $\pm$ 0.02 & \cellcolor{mynicered!35}0.61 $\pm$ 0.01 & \cellcolor{mynicered!35}0.82 $\pm$ 0.03 & \cellcolor{mynicegreen!35}0.16 $\pm$ 0.00 & \cellcolor{mynicered!35}0.89 $\pm$ 0.03 & \cellcolor{myniceblue!35}0.45 $\pm$ 0.02 & \cellcolor{mynicered!35}0.59 $\pm$ 0.06 & \cellcolor{mynicered!35}0.79 $\pm$ 0.04 &\cellcolor{mynicered!35} 0.56 $\pm$ 0.22 \\ 
\sa~\cite{heng2023selective} & 
\cellcolor{mynicered!35} 0.69 $\pm$ 0.09 & 
\cellcolor{mynicered!35} 0.69 $\pm$ 0.00 & \cellcolor{mynicegreen!35}0.16 $\pm$ 0.00 & \cellcolor{mynicered!35}0.87 $\pm$ 0.04 & \cellcolor{mynicered!35}0.93 $\pm$ 0.02 & \cellcolor{mynicered!35}0.55 $\pm$ 0.07 & \cellcolor{mynicered!35}0.98 $\pm$ 0.01 & \cellcolor{mynicered!35}0.82 $\pm$ 0.05 & \cellcolor{myniceblue!35}0.49 $\pm$ 0.04 & \cellcolor{mynicered!35}0.63 $\pm$ 0.05 & \cellcolor{mynicered!35}0.75 $\pm$ 0.04 \\ 
\esd~\cite{gandikota2023erasing} & 
\cellcolor{mynicered!35} 0.62 $\pm$ 0.06 & 
\cellcolor{mynicered!35} 0.63 $\pm$ 0.01 & \cellcolor{myniceblue!35}0.28 $\pm$ 0.06 & \cellcolor{mynicered!35}0.64 $\pm$ 0.06 & \cellcolor{myniceblue!35}0.37 $\pm$ 0.04 & \cellcolor{mynicegreen!35}0.20 $\pm$ 0.02 & \cellcolor{myniceblue!35}0.41 $\pm$ 0.04 & \cellcolor{mynicegreen!35}0.11 $\pm$ 0.02 & \cellcolor{myniceblue!35}0.29 $\pm$ 0.05 & \cellcolor{mynicegreen!35}0.17 $\pm$ 0.02 & \cellcolor{mynicegreen!35}0.17 $\pm$ 0.02 \\ 
\unifiedCE~\cite{gandikota2023unified} &
\cellcolor{mynicered!35}0.70 $\pm$ 0.01 & 
\cellcolor{mynicered!35} 0.71 $\pm$ 0.01 & \cellcolor{mynicegreen!35}0.05 $\pm$ 0.00 & \cellcolor{myniceblue!35}0.43 $\pm$ 0.00 & \cellcolor{mynicegreen!35}0.04 $\pm$ 0.00 & \cellcolor{mynicegreen!35}0.03 $\pm$ 0.00 & \cellcolor{myniceblue!35}0.40 $\pm$ 0.01 & \cellcolor{mynicegreen!35}0.02 $\pm$ 0.01 & \cellcolor{mynicegreen!35}0.06 $\pm$ 0.00 & \cellcolor{mynicegreen!35}0.05 $\pm$ 0.00 & \cellcolor{mynicegreen!35}0.10 $\pm$ 0.01 \\ 
\sdd~\cite{kim2023towards} & 
\cellcolor{mynicered!35} 0.57 $\pm$ 0.02 & 
\cellcolor{mynicered!35}0.55 $\pm$ 0.02 & 
\cellcolor{mynicegreen!35}0.20 $\pm$ 0.02 & \cellcolor{mynicegreen!35}0.20 $\pm$ 0.03 & \cellcolor{myniceblue!35}0.41 $\pm$ 0.03 & \cellcolor{myniceblue!35}0.37 $\pm$ 0.03 & \cellcolor{myniceblue!35}0.39 $\pm$ 0.02 & \cellcolor{mynicegreen!35}0.05 $\pm$ 0.02 & \cellcolor{mynicegreen!35}0.06 $\pm$ 0.01 & \cellcolor{mynicegreen!35}0.04 $\pm$ 0.01 & \cellcolor{mynicegreen!35}0.06 $\pm$ 0.01 \\ 
\moderator~\cite{wang2024moderator} & 
\cellcolor{mynicered!35} 0.87 $\pm$ 0.02 & 
\cellcolor{mynicered!35}0.92 $\pm$ 0.05 & 
\cellcolor{mynicegreen!35}0.02 $\pm$ 0.00 & \cellcolor{mynicegreen!35}0.25 $\pm$ 0.02 & \cellcolor{mynicered!35}0.73 $\pm$ 0.03 & \cellcolor{mynicegreen!35}0.01 $\pm$ 0.00 & \cellcolor{myniceblue!35}0.28 $\pm$ 0.02 & \cellcolor{mynicegreen!35}0.03 $\pm$ 0.01 & \cellcolor{mynicegreen!35}0.09 $\pm$ 0.01 & \cellcolor{mynicegreen!35}0.00 $\pm$ 0.01 & \cellcolor{mynicegreen!35}0.13 $\pm$ 0.04 \\ 
\midrule
 \textbf{\method} & 
  \cellcolor{mynicegreen!35}0.15 $\pm$ 0.06& 
   \cellcolor{mynicegreen!35}0.20 $\pm$ 0.05 & 
  \cellcolor{mynicegreen!35} 0.00 $\pm$ 0.01 &
  \cellcolor{mynicegreen!35} 0.10 $\pm$ 0.02 & 
  \cellcolor{mynicegreen!35} 0.03 $\pm$ 0.01& 
  \cellcolor{mynicegreen!35} 0.08 $\pm$ 0.02& 
  \cellcolor{mynicegreen!35} 0.00 $\pm$ 0.00 & 
   \cellcolor{mynicegreen!35}0.02 $\pm$ 0.00 & \cellcolor{mynicegreen!35}0.03 $\pm$ 0.00 & \cellcolor{mynicegreen!35}0.00 $\pm$ 0.00 & \cellcolor{mynicegreen!35}0.03 $\pm$ 0.00 \\ 
\bottomrule

\toprule
 \end{tabular}
}
\label{tab:effectiveness}
\end{table*}
\begin{table*}[ht]
\caption{\underline{\ref{utility} (Utility):} Comparison with fine-tuning \crt{s} using normalized CLIP scores on $\dtestacc$ (higher is better). We use \colorbox{mynicered}{red} if score is between 50-70, \colorbox{myniceblue}{blue} if between 70-90; \colorbox{mynicegreen}{green} if $>$90. \textbf{\underline{Note:}} \method's utility against Group-2 concepts like \emph{Nemo}, \emph{Captain Marvel}, \emph{Snoopy}, and \emph{R2D2} can be improved as described in Section~\ref{sec:finetuneEval}.}
\centering
\resizebox{0.9\textwidth}{!}{%
\begin{tabular}{ l|c|c|c|c|c|c|c|c|c|c|c } 
\bottomrule

\toprule
 \multirow{3}{*}{\textbf{CRT}} & \multicolumn{11}{c}{\textbf{Concepts}}\\
  & \multirow{2}{*}{\textbf{Nudity}} & \multirow{2}{*}{\textbf{Violence}} & \textbf{Grumpy} & \multirow{2}{*}{\textbf{Nemo}} & \textbf{Captain} & \multirow{2}{*}{\textbf{Snoopy}} & \multirow{2}{*}{\textbf{R2D2}} & \textbf{Taylor} & \textbf{Angelina} & \textbf{Brad} & \textbf{Elon}\\
  &  & & \textbf{Cat} &  & \textbf{Marvel} &  &  &  \textbf{Swift} & \textbf{Jolie} & \textbf{Pitt} & \textbf{Musk}\\
\midrule
\texttt{CA}~\cite{kumari2023ablating}  & \cellcolor{mynicegreen!35}0.93 $\pm$ 0.00 & \cellcolor{mynicegreen!35}0.93 $\pm$ 0.00 & \cellcolor{mynicegreen!35}0.93 $\pm$ 0.00 & \cellcolor{mynicegreen!35}0.93 $\pm$ 0.00 & \cellcolor{mynicegreen!35}0.93 $\pm$ 0.00 & \cellcolor{mynicegreen!35}0.93 $\pm$ 0.00 & \cellcolor{mynicegreen!35}0.93 $\pm$ 0.00 & \cellcolor{mynicegreen!35}0.93 $\pm$ 0.00 & \cellcolor{mynicegreen!35}0.93 $\pm$ 0.00 & \cellcolor{mynicegreen!35}0.93 $\pm$ 0.00 & \cellcolor{mynicegreen!35}0.93 $\pm$ 0.00 \\ 
\forgetNot~\cite{zhang2023forgetmenot} & \cellcolor{myniceblue!35}0.79 $\pm$ 0.00 & \cellcolor{myniceblue!35}0.79 $\pm$ 0.00 & \cellcolor{myniceblue!35}0.79 $\pm$ 0.00 & \cellcolor{myniceblue!35}0.79 $\pm$ 0.00 & \cellcolor{myniceblue!35}0.79 $\pm$ 0.00 & \cellcolor{myniceblue!35}0.79 $\pm$ 0.00 & \cellcolor{myniceblue!35}0.79 $\pm$ 0.00 & \cellcolor{myniceblue!35}0.79 $\pm$ 0.00 & \cellcolor{myniceblue!35}0.79 $\pm$ 0.00  & \cellcolor{myniceblue!35}0.79 $\pm$ 0.00 & \cellcolor{myniceblue!35}0.79 $\pm$ 0.00 \\ 
\sa~\cite{heng2023selective} & \cellcolor{myniceblue!35}0.79 $\pm$ 0.00 & \cellcolor{myniceblue!35}0.79 $\pm$ 0.00 & \cellcolor{myniceblue!35}0.79 $\pm$ 0.00 & \cellcolor{myniceblue!35}0.79 $\pm$ 0.00 & \cellcolor{myniceblue!35}0.79 $\pm$ 0.00 & \cellcolor{myniceblue!35}0.79 $\pm$ 0.00 & \cellcolor{myniceblue!35}0.79 $\pm$ 0.00 & \cellcolor{myniceblue!35}0.79 $\pm$ 0.00 & \cellcolor{myniceblue!35}0.79 $\pm$ 0.00 & \cellcolor{myniceblue!35}0.79 $\pm$ 0.01 & \cellcolor{myniceblue!35}0.79 $\pm$ 0.00 \\  
\esd~\cite{gandikota2023erasing} & \cellcolor{myniceblue!35}0.82 $\pm$ 0.01 & \cellcolor{myniceblue!35}0.82 $\pm$ 0.00 & \cellcolor{myniceblue!35}0.82 $\pm$ 0.01 & \cellcolor{myniceblue!35}0.82 $\pm$ 0.00  & \cellcolor{myniceblue!35}0.82 $\pm$ 0.01 & \cellcolor{myniceblue!35}0.82 $\pm$ 0.01 & \cellcolor{myniceblue!35}0.82 $\pm$ 0.01 & \cellcolor{myniceblue!35}0.82 $\pm$ 0.01 & \cellcolor{myniceblue!35}0.82 $\pm$ 0.01 & \cellcolor{myniceblue!35}0.82 $\pm$ 0.01 & \cellcolor{myniceblue!35}0.82 $\pm$ 0.01 \\ 
\unifiedCE~\cite{gandikota2023unified} & \cellcolor{mynicegreen!35}0.96 $\pm$ 0.02 & \cellcolor{mynicegreen!35}0.96 $\pm$ 0.00 & \cellcolor{mynicegreen!35}0.97 $\pm$ 0.03 & \cellcolor{mynicegreen!35}0.96 $\pm$ 0.00 & \cellcolor{mynicegreen!35}0.96 $\pm$ 0.00 & \cellcolor{mynicegreen!35}0.97 $\pm$ 0.02 & \cellcolor{mynicegreen!35}0.96 $\pm$ 0.00 & \cellcolor{mynicegreen!35}0.97 $\pm$ 0.02 & \cellcolor{mynicegreen!35}0.97 $\pm$ 0.01 & \cellcolor{mynicegreen!35}0.96 $\pm$ 0.03 & \cellcolor{mynicegreen!35}0.97 $\pm$ 0.00 \\ 
\sdd~\cite{kim2023towards} & \cellcolor{myniceblue!35}0.86 $\pm$ 0.00 & \cellcolor{myniceblue!35}0.75 $\pm$ 0.00 & \cellcolor{mynicegreen!35}0.93 $\pm$ 0.00 & \cellcolor{myniceblue!35}0.86 $\pm$ 0.00 & \cellcolor{myniceblue!35}0.82 $\pm$ 0.00 & \cellcolor{myniceblue!35}0.82 $\pm$ 0.00 & \cellcolor{mynicered!35}0.64 $\pm$ 0.00 & \cellcolor{myniceblue!35}0.82 $\pm$ 0.00 & \cellcolor{myniceblue!35}0.82 $\pm$ 0.00 & \cellcolor{mynicered!35}0.61 $\pm$ 0.00 & \cellcolor{myniceblue!35}0.89 $\pm$ 0.00 \\ 
\moderator~\cite{wang2024moderator} & \cellcolor{mynicegreen!35}0.93 $\pm$ 0.01 & \cellcolor{mynicered!35}0.53 $\pm$ 0.00 & \cellcolor{mynicered!35}0.57 $\pm$ 0.00 & \cellcolor{mynicered!35}0.55 $\pm$ 0.00 & \cellcolor{mynicered!35}0.53 $\pm$ 0.00 & \cellcolor{mynicered!35}0.50 $\pm$ 0.01 & \cellcolor{mynicered!35}0.46 $\pm$ 0.00 & \cellcolor{myniceblue!35}0.86 $\pm$ 0.00 & \cellcolor{mynicered!35}0.61 $\pm$ 0.00 & \cellcolor{mynicered!35}0.68 $\pm$ 0.00 & \cellcolor{myniceblue!35}0.71 $\pm$ 0.00 \\
\midrule
\textbf{\method} & 
  \cellcolor{mynicegreen!35}0.94 $\pm$ 0.08 & 
 \cellcolor{mynicered!35}0.59 $\pm$ 0.11&  
 \cellcolor{mynicegreen!35}0.98 $\pm$ 0.04 &  
 \cellcolor{mynicered!35} 0.37 $\pm$ 0.02 & 
 \cellcolor{mynicered!35} 0.37 $\pm$ 0.03 &  
 \cellcolor{myniceblue!35} 0.66 $\pm$ 0.03& 
 \cellcolor{myniceblue!35} 0.66 $\pm$ 0.02 & \cellcolor{mynicegreen!35}0.98 $\pm$ 0.02  &  \cellcolor{mynicegreen!35}0.98 $\pm$ 0.03 & 
\cellcolor{mynicegreen!35}0.98 $\pm$ 0.01 & \cellcolor{mynicegreen!35}0.97 $\pm$ 0.01 \\
\bottomrule

\toprule
 \end{tabular}
}
\label{tab:utility}
\end{table*}

\begin{table*}[!htb]
\centering
\caption{\underline{\ref{robust} (Robustness):} Comparison with fine-tuning \crt{s} using CLIP accuracy on \emph{adversarial} prompts in $\dtestadv$ (lower is better). We evaluate fine-tuning \crt's against \cce and \method against \cce+ since \cce is already adapted to fine-tuning \crt's. We use \colorbox{mynicered}{red} if accuracy is $>$50; \colorbox{myniceblue}{blue} if accuracy is between 25-50; \colorbox{mynicegreen}{green} if accuracy is $<$25.}
\resizebox{0.9\textwidth}{!}{
\begin{tabular}{  l | c c | c c c c c | c c c c  }
\bottomrule

\toprule
\multirow{3}{*}{\textbf{CRT}} & \multicolumn{11}{c}{\textbf{Concepts}}\\
& \textbf{Nudity} & \textbf{Violence} & \textbf{Grumpy} & \multirow{2}{*}{\textbf{Nemo}} & \textbf{Captain} & \multirow{2}{*}{\textbf{Snoopy}} & \multirow{2}{*}{\textbf{R2D2}} & \textbf{Taylor} & \textbf{Angelina} & \textbf{Brad} & \textbf{Elon}\\
 & \textbf{(I2P)} & \textbf{(I2P)} & \textbf{Cat} &  & \textbf{Marvel} &  &  &  \textbf{Swift} & \textbf{Jolie} & \textbf{Pitt} & \textbf{Musk}\\
\bottomrule

\toprule
\multicolumn{12}{c}{\textbf{\typo+}}\\
\bottomrule

\toprule
\texttt{CA}~\cite{kumari2023ablating} & 
\cellcolor{mynicered!35} 0.58 $\pm$ 0.02 &
\cellcolor{mynicered!35} 0.75 $\pm$ 0.01&
\cellcolor{myniceblue!35}0.26 $\pm$ 0.02 & \cellcolor{myniceblue!35}0.27 $\pm$ 0.01 & \cellcolor{myniceblue!35}0.42 $\pm$ 0.01 & \cellcolor{myniceblue!35}0.29 $\pm$ 0.02 & \cellcolor{mynicegreen!35}0.23 $\pm$ 0.02 & \cellcolor{mynicegreen!35}0.09 $\pm$ 0.02 & \cellcolor{mynicegreen!35}0.24 $\pm$ 0.01 & \cellcolor{mynicegreen!35}0.05 $\pm$ 0.01 & \cellcolor{myniceblue!35}0.31 $\pm$ 0.06 \\ 
\forgetNot~\cite{zhang2023forgetmenot} &
\cellcolor{mynicered!35}0.61 $\pm$ 0.02& \cellcolor{mynicered!35}0.75 $\pm$ 0.02 & \cellcolor{mynicegreen!35}0.21 $\pm$ 0.01 & \cellcolor{myniceblue!35}0.31 $\pm$ 0.01 & \cellcolor{myniceblue!35}0.49 $\pm$ 0.02 &\cellcolor{myniceblue!35}0.27 $\pm$ 0.02 & \cellcolor{mynicegreen!35}0.22 $\pm$ 0.02 & \cellcolor{mynicegreen!35}0.03 $\pm$ 0.01 & \cellcolor{mynicegreen!35}0.17 $\pm$ 0.01 & \cellcolor{mynicegreen!35}0.06 $\pm$ 0.01 & \cellcolor{myniceblue!35}0.34 $\pm$ 0.01 \\
\sa~\cite{heng2023selective} & \cellcolor{myniceblue!35} 0.31 $\pm$ 0.01 & 
\cellcolor{mynicered!35}0.71 $\pm$ 0.02  & \cellcolor{mynicered!35}0.99 $\pm$ 0.01 & \cellcolor{mynicered!35}0.94 $\pm$ 0.01 & \cellcolor{mynicered!35}0.89 $\pm$ 0.02 & \cellcolor{mynicered!35}0.73 $\pm$ 0.03 & \cellcolor{mynicered!35}0.99 $\pm$ 0.00 & \cellcolor{mynicegreen!35}0.20 $\pm$ 0.02 & \cellcolor{mynicegreen!35}0.05 $\pm$ 0.01 & \cellcolor{myniceblue!35}0.43 $\pm$ 0.04 & \cellcolor{mynicered!35}0.65 $\pm$ 0.05 \\ 
\esd~\cite{gandikota2023erasing} & 
\cellcolor{myniceblue!35}0.39 $\pm$ 0.01 & 
\cellcolor{mynicered!35} 0.70 $\pm$ 0.01 & \cellcolor{myniceblue!35}0.27 $\pm$ 0.02 & \cellcolor{mynicegreen!35}0.25 $\pm$ 0.05 & \cellcolor{myniceblue!35}0.40 $\pm$ 0.03 & \cellcolor{mynicegreen!35}0.23 $\pm$ 0.02 & \cellcolor{mynicegreen!35}0.25 $\pm$ 0.05 & \cellcolor{mynicegreen!35}0.03 $\pm$ 0.01 & \cellcolor{mynicegreen!35}0.08 $\pm$ 0.07 & \cellcolor{mynicegreen!35}0.04 $\pm$ 0.03 & \cellcolor{mynicegreen!35}0.23 $\pm$ 0.05 \\ 
\unifiedCE~\cite{gandikota2023unified} & \cellcolor{myniceblue!35} 0.41 $\pm$ 0.00 &
\cellcolor{mynicered!35} 0.60 $\pm$ 0.00 & \cellcolor{myniceblue!35}0.28 $\pm$ 0.02 & \cellcolor{myniceblue!35}0.29 $\pm$ 0.02 & \cellcolor{myniceblue!35}0.34 $\pm$ 0.02 & \cellcolor{mynicegreen!35}0.21 $\pm$ 0.03 & \cellcolor{mynicegreen!35}0.17 $\pm$ 0.02 & \cellcolor{mynicegreen!35}0.00 $\pm$ 0.00 & \cellcolor{mynicegreen!35}0.05 $\pm$ 0.00 & \cellcolor{mynicegreen!35}0.02 $\pm$ 0.00 & \cellcolor{mynicegreen!35}0.12 $\pm$ 0.00 \\ 
\sdd~\cite{kim2023towards} & 
\cellcolor{mynicegreen!35} 0.20 $\pm$ 0.02 &
\cellcolor{myniceblue!35}0.50 $\pm$ 0.04 & \cellcolor{myniceblue!35}0.27 $\pm$ 0.02 & \cellcolor{mynicegreen!35}0.21 $\pm$ 0.02 & \cellcolor{myniceblue!35}0.48 $\pm$ 0.01 & \cellcolor{mynicegreen!35}0.19 $\pm$ 0.01 & \cellcolor{myniceblue!35}0.31 $\pm$ 0.00 & \cellcolor{mynicegreen!35}0.05 $\pm$ 0.01 & \cellcolor{mynicegreen!35}0.06 $\pm$ 0.00 & \cellcolor{mynicegreen!35}0.05 $\pm$ 0.00 & \cellcolor{mynicegreen!35}0.10 $\pm$ 0.01 \\ 
\moderator~\cite{wang2024moderator} & 
\cellcolor{mynicered!35} 0.83 $\pm$ 0.01 &
\cellcolor{myniceblue!35}0.90 $\pm$ 0.01 & \cellcolor{mynicegreen!35}0.01 $\pm$ 0.00 & \cellcolor{mynicegreen!35}0.21 $\pm$ 0.02 & \cellcolor{myniceblue!35}0.69 $\pm$ 0.04 & \cellcolor{mynicegreen!35}0.06 $\pm$ 0.01 & \cellcolor{myniceblue!35}0.30 $\pm$ 0.00 & \cellcolor{mynicegreen!35}0.01 $\pm$ 0.01 & \cellcolor{mynicegreen!35}0.05 $\pm$ 0.00 & \cellcolor{mynicegreen!35}0.00 $\pm$ 0.00 & \cellcolor{mynicegreen!35}0.13 $\pm$ 0.02 \\ 
\midrule
\textbf{\method} & \cellcolor{mynicegreen!35} 0.14 $\pm$ 0.01 & \cellcolor{mynicegreen!35} 0.20 $\pm$ 0.01 & \cellcolor{mynicegreen!35}0.10 $\pm$ 0.01 &  
\cellcolor{mynicegreen!35}0.06 $\pm$ 0.01 &  \cellcolor{mynicegreen!35}0.09 $\pm$ 0.01 &   \cellcolor{mynicegreen!35}0.09 $\pm$ 0.01  &   \cellcolor{mynicegreen!35}0.08 $\pm$ 0.01   & \cellcolor{mynicegreen!35} 0.00 $\pm$ 0.01 & \cellcolor{mynicegreen!35} 0.00 $\pm$ 0.01 & \cellcolor{mynicegreen!35} 0.01 $\pm$ 0.01 & \cellcolor{mynicegreen!35} 0.01 $\pm$ 0.01   \\ 
\bottomrule

\toprule
\multicolumn{12}{c}{\textbf{\pez+}}\\
\bottomrule

\toprule
 \texttt{CA}~\cite{kumari2023ablating} & \cellcolor{mynicered!35} 0.75 $\pm$ 0.01 &
 \cellcolor{mynicered!35} 0.84 $\pm$ 0.02 & \cellcolor{myniceblue!35}0.33 $\pm$ 0.04 & \cellcolor{mynicered!35}0.52 $\pm$ 0.01 & \cellcolor{mynicered!35}0.70 $\pm$ 0.02 & \cellcolor{mynicegreen!35}0.20 $\pm$ 0.01 & \cellcolor{mynicegreen!35}0.25 $\pm$ 0.03 & \cellcolor{myniceblue!35}0.46 $\pm$ 0.01 & \cellcolor{mynicered!35}0.64 $\pm$ 0.01 & \cellcolor{mynicered!35}0.63 $\pm$ 0.02 & \cellcolor{mynicered!35}0.72 $\pm$ 0.01 \\
\forgetNot~\cite{zhang2023forgetmenot} & 
\cellcolor{mynicered!35}0.74 $\pm$ 0.01&
\cellcolor{mynicered!35}0.72 $\pm$ 0.02 & \cellcolor{myniceblue!35}0.43 $\pm$ 0.02 & \cellcolor{myniceblue!35}0.41 $\pm$ 0.01 & \cellcolor{mynicered!35}0.85 $\pm$ 0.03 & \cellcolor{myniceblue!35}0.45 $\pm$ 0.01 & \cellcolor{mynicered!35}0.93 $\pm$ 0.06 & \cellcolor{mynicegreen!35}0.04 $\pm$ 0.01 & \cellcolor{mynicegreen!35}0.16 $\pm$ 0.01 & \cellcolor{mynicegreen!35}0.08 $\pm$ 0.01 & \cellcolor{mynicegreen!35}0.23 $\pm$ 0.01 \\ 
\sa~\cite{heng2023selective} & \cellcolor{mynicered!35}0.55 $\pm$ 0.03& 
\cellcolor{mynicered!35}0.82 $\pm$ 0.01& \cellcolor{mynicegreen!35}0.14 $\pm$ 0.00 & \cellcolor{mynicegreen!35}0.14 $\pm$ 0.00 & \cellcolor{mynicegreen!35}0.14 $\pm$ 0.01 & \cellcolor{mynicegreen!35}0.15 $\pm$ 0.01 & \cellcolor{mynicegreen!35}0.15 $\pm$ 0.00 & \cellcolor{mynicegreen!35}0.15 $\pm$ 0.01 & \cellcolor{mynicegreen!35}0.15 $\pm$ 0.01 & \cellcolor{mynicegreen!35}0.14 $\pm$ 0.00 & \cellcolor{mynicegreen!35}0.15 $\pm$ 0.01 \\ 
\esd~\cite{gandikota2023erasing} & 
\cellcolor{mynicered!35} 0.69 $\pm$ 0.01 & 
\cellcolor{mynicered!35} 0.88 $\pm$ 0.01 & \cellcolor{myniceblue!35}0.36 $\pm$ 0.06 & \cellcolor{myniceblue!35}0.40 $\pm$ 0.04 & \cellcolor{myniceblue!35}0.44 $\pm$ 0.02 & \cellcolor{myniceblue!35}0.34 $\pm$ 0.03 & \cellcolor{myniceblue!35}0.26 $\pm$ 0.03 & \cellcolor{mynicegreen!35}0.05 $\pm$ 0.02 & \cellcolor{mynicegreen!35}0.11 $\pm$ 0.04 & \cellcolor{mynicegreen!35}0.17 $\pm$ 0.02 & \cellcolor{mynicegreen!35}0.23 $\pm$ 0.03 \\ 
\unifiedCE~\cite{gandikota2023unified} & 
\cellcolor{mynicered!35} 0.59 $\pm$ 0.00 & 
\cellcolor{mynicered!35} 0.82 $\pm$ 0.00 & \cellcolor{mynicegreen!35}0.23 $\pm$ 0.01 & \cellcolor{mynicered!35}0.52 $\pm$ 0.01 & \cellcolor{mynicered!35}0.59 $\pm$ 0.03 & \cellcolor{mynicegreen!35}0.14 $\pm$ 0.02 & \cellcolor{mynicegreen!35}0.25 $\pm$ 0.02 & \cellcolor{mynicegreen!35}0.00 $\pm$ 0.00 & \cellcolor{mynicegreen!35}0.06 $\pm$ 0.01 & \cellcolor{mynicegreen!35}0.06 $\pm$ 0.01 & \cellcolor{mynicegreen!35}0.15 $\pm$ 0.02 \\ 
\sdd~\cite{kim2023towards} & \cellcolor{myniceblue!35} 0.30 $\pm$ 0.01 & 
\cellcolor{mynicered!35} 0.60 $\pm$ 0.01 & \cellcolor{myniceblue!35}0.28 $\pm$ 0.05 & \cellcolor{myniceblue!35}0.28 $\pm$ 0.01 & \cellcolor{myniceblue!35}0.50 $\pm$ 0.03 & \cellcolor{myniceblue!35}0.34 $\pm$ 0.03 & \cellcolor{myniceblue!35}0.30 $\pm$ 0.03 & \cellcolor{mynicegreen!35}0.04 $\pm$ 0.01 & \cellcolor{mynicegreen!35}0.09 $\pm$ 0.02 & \cellcolor{mynicegreen!35}0.06 $\pm$ 0.01 & \cellcolor{mynicegreen!35}0.12 $\pm$ 0.01 \\ 
\moderator~\cite{wang2024moderator} & 
\cellcolor{mynicered!35} 0.84 $\pm$ 0.01 &
\cellcolor{myniceblue!35}0.90 $\pm$ 0.01 & \cellcolor{mynicegreen!35}0.01 $\pm$ 0.00 & \cellcolor{mynicegreen!35}0.27 $\pm$ 0.02 & \cellcolor{myniceblue!35}0.71 $\pm$ 0.02 & \cellcolor{mynicegreen!35}0.02 $\pm$ 0.01 & \cellcolor{myniceblue!35}0.33 $\pm$ 0.00 & \cellcolor{mynicegreen!35}0.06 $\pm$ 0.01 & \cellcolor{mynicegreen!35}0.07 $\pm$ 0.00 & \cellcolor{mynicegreen!35}0.01 $\pm$ 0.00 & \cellcolor{mynicegreen!35}0.15 $\pm$ 0.02 \\ 
\midrule
\textbf{\method} & \cellcolor{mynicegreen!35}0.15 $\pm$ 0.01  & \cellcolor{mynicegreen!35}0.25 $\pm$ 0.05 & \cellcolor{mynicegreen!35}0.10 $\pm$ 0.01 &  \cellcolor{mynicegreen!35}0.12 $\pm$ 0.01 &   \cellcolor{mynicegreen!35}0.11 $\pm$ 0.03  &  \cellcolor{mynicegreen!35}0.08 $\pm$ 0.01 &   \cellcolor{mynicegreen!35}0.03 $\pm$ 0.00 
& \cellcolor{mynicegreen!35}0.00 $\pm$ 0.01  & \cellcolor{mynicegreen!35}0.03 $\pm$ 0.00  & \cellcolor{mynicegreen!35}0.04 $\pm$ 0.00  & \cellcolor{mynicegreen!35}0.04 $\pm$ 0.00  \\ 
\bottomrule

\toprule
\multicolumn{12}{c}{\textbf{\ringbell+}}\\
\bottomrule

\toprule
\texttt{CA}~\cite{kumari2023ablating} & 
\cellcolor{mynicered!35} 0.97 $\pm$ 0.01 & 
\cellcolor{mynicered!35} 0.96 $\pm$ 0.01 & 
\cellcolor{mynicered!35} 0.79 $\pm$ 0.01 & 
\cellcolor{mynicered!35} 0.76 $\pm$ 0.02 & 
\cellcolor{mynicered!35} 0.88 $\pm$ 0.02 & 
\cellcolor{myniceblue!35} 0.38 $\pm$ 0.02 &
\cellcolor{mynicered!35} 0.65 $\pm$ 0.05 & 
\cellcolor{mynicegreen!35} 0.03 $\pm$ 0.03 & 
\cellcolor{mynicegreen!35} 0.00 $\pm$ 0.01 & 
\cellcolor{mynicered!35} 0.88 $\pm$ 0.01 & 
\cellcolor{mynicered!35} 1.00 $\pm$ 0.01 \\ 
\forgetNot~\cite{zhang2023forgetmenot} & 
\cellcolor{mynicered!35} 0.96 $\pm$ 0.01 & 
\cellcolor{mynicered!35} 0.95 $\pm$ 0.02 & 
\cellcolor{mynicered!35} 0.75 $\pm$ 0.00 & 
\cellcolor{mynicered!35} 0.57 $\pm$ 0.01 & 
\cellcolor{mynicered!35} 0.91 $\pm$ 0.00 & 
\cellcolor{myniceblue!35} 0.45 $\pm$ 0.01 & 
\cellcolor{mynicered!35} 0.59 $\pm$ 0.01 &  
\cellcolor{myniceblue!35} 0.26 $\pm$ 0.01 & 
\cellcolor{mynicered!35} 0.85 $\pm$ 0.02 & 
\cellcolor{mynicered!35} 0.88 $\pm$ 0.01 & 
\cellcolor{mynicered!35} 0.99 $\pm$ 0.02 \\ 
\sa~\cite{heng2023selective} & 
\cellcolor{mynicered!35} 0.80 $\pm$ 0.02 & 
\cellcolor{mynicered!35} 0.98 $\pm$ 0.02 & 
\cellcolor{mynicered!35} 0.93 $\pm$ 0.02 & 
\cellcolor{mynicered!35} 0.98 $\pm$ 0.01 & 
\cellcolor{mynicered!35} 0.96 $\pm$  0.03 & 
\cellcolor{mynicered!35} 0.97 $\pm$ 0.03 & 
\cellcolor{mynicered!35} 0.88 $\pm$ 0.02 & 
\cellcolor{mynicegreen!35} 0.00 $\pm$ 0.01 & 
\cellcolor{mynicegreen!35} 0.03 $\pm$ 0.02 & 
\cellcolor{mynicered!35} 0.77 $\pm$ 0.10 & 
\cellcolor{mynicered!35} 1.00 $\pm$ 0.01 \\ 
\esd~\cite{gandikota2023erasing} & \cellcolor{mynicered!35} 0.77 $\pm$ 0.03
& 
\cellcolor{mynicered!35} 0.95 $\pm$ 0.02 & 
\cellcolor{mynicered!35} 0.63 $\pm$ 0.06 & 
\cellcolor{mynicered!35} 0.66 $\pm$ 0.12 & 
\cellcolor{mynicered!35} 0.56 $\pm$ 0.06 & 
\cellcolor{mynicered!35} 0.66 $\pm$ 0.07 & 
\cellcolor{mynicered!35} 0.69 $\pm$ 0.01 & 
\cellcolor{mynicegreen!35} 0.00 $\pm$ 0.00 & 
\cellcolor{mynicegreen!35} 0.03 $\pm$ 0.02 & 
\cellcolor{myniceblue!35} 0.27 $\pm$ 0.03 & 
\cellcolor{mynicered!35} 0.55 $\pm$ 0.08 \\ 
\unifiedCE~\cite{gandikota2023unified} & 
\cellcolor{mynicered!35} 0.84 $\pm$ 0.00 & 
\cellcolor{mynicered!35} 0.67 $\pm$ 0.00 & 
\cellcolor{myniceblue!35} 0.38 $\pm$ 0.05 & 
\cellcolor{mynicered!35} 0.74 $\pm$ 0.01 & 
\cellcolor{mynicegreen!35} 0.07 $\pm$ 0.00 &
\cellcolor{mynicegreen!35} 0.16 $\pm$ 0.01 & 
\cellcolor{mynicered!35} 0.50 $\pm$ 0.01 & 
\cellcolor{mynicegreen!35} 0.05 $\pm$ 0.00 & 
\cellcolor{mynicegreen!35} 0.01 $\pm$ 0.00 & 
\cellcolor{mynicegreen!35} 0.02 $\pm$ 0.01 & 
\cellcolor{myniceblue!35} 0.34 $\pm$ 0.01 \\ 
\sdd~\cite{kim2023towards} & 
\cellcolor{myniceblue!35} 0.33 $\pm$ 0.02 &
 \cellcolor{mynicered!35} 0.60 $\pm$ 0.03 & 
\cellcolor{mynicegreen!35} 0.22 $\pm$ 0.01 & 
\cellcolor{myniceblue!35} 0.31 $\pm$ 0.01 &
\cellcolor{mynicered!35} 0.62 $\pm$ 0.01 &
\cellcolor{myniceblue!35} 0.42 $\pm$ 0.03 & 
\cellcolor{myniceblue!35} 0.41 $\pm$ 0.01 & 
\cellcolor{mynicegreen!35} 0.07 $\pm$ 0.02 & 
\cellcolor{mynicegreen!35} 0.07 $\pm$ 0.02 & 
\cellcolor{mynicegreen!35} 0.07 $\pm$ 0.01 & 
\cellcolor{mynicegreen!35} 0.17 $\pm$ 0.02 \\ 
\moderator~\cite{wang2024moderator} & 
\cellcolor{mynicered!35} 0.98 $\pm$ 0.03 &
 \cellcolor{mynicered!35} 0.95 $\pm$ 0.03 & 
\cellcolor{mynicegreen!35} 0.02 $\pm$ 0.01 & 
\cellcolor{myniceblue!35} 0.26 $\pm$ 0.01 &
\cellcolor{mynicered!35} 0.73 $\pm$ 0.00 &
\cellcolor{mynicegreen!35} 0.10 $\pm$ 0.03 & 
\cellcolor{myniceblue!35} 0.36 $\pm$ 0.02 & 
\cellcolor{mynicegreen!35} 0.07 $\pm$ 0.02 & 
\cellcolor{mynicegreen!35} 0.10 $\pm$ 0.00 & 
\cellcolor{mynicegreen!35} 0.00 $\pm$ 0.01 & 
\cellcolor{mynicegreen!35} 0.18 $\pm$ 0.01 \\
\midrule
\textbf{\method} & \cellcolor{mynicegreen!35} 0.05 $\pm$ 0.01 & \cellcolor{mynicegreen!35} 0.08 $\pm$ 0.01 & \cellcolor{mynicegreen!35} 0.20 $\pm$ 0.08 & \cellcolor{mynicegreen!35} 0.15 $\pm$ 0.03 &
\cellcolor{mynicegreen!35} 0.04 $\pm$ 0.02  & \cellcolor{mynicegreen!35} 0.01 $\pm$ 0.01 & 
\cellcolor{mynicegreen!35} 0.15 $\pm$ 0.05 & \cellcolor{mynicegreen!35} 0.00 $\pm$ 0.02 &  \cellcolor{mynicegreen!35} 0.03 $\pm$ 0.02 &  \cellcolor{mynicegreen!35} 0.01 $\pm$ 0.02 &   \cellcolor{mynicegreen!35} 0.02 $\pm$ 0.02 \\ 
\bottomrule

\toprule
 \multicolumn{12}{c}{\textbf{\cce or \cce+ (against \method)}}\\
\bottomrule

\toprule
\texttt{CA}~\cite{kumari2023ablating} & 
\cellcolor{mynicered!35} 1.00 $\pm$ 0.00 & 
 \cellcolor{mynicered!35} 1.00 $\pm$ 0.00 & 
\cellcolor{mynicered!35} 1.00 $\pm$ 0.00 &  
\cellcolor{mynicered!35} 0.99 $\pm$ 0.00 & 
\cellcolor{mynicered!35} 0.97 $\pm$ 0.01 &  
\cellcolor{mynicered!35} 1.00 $\pm$ 0.00 &  
\cellcolor{mynicered!35} 0.99 $\pm$ 0.01 &
\cellcolor{mynicered!35} 1.00 $\pm$ 0.00 & 
\cellcolor{mynicered!35}1.00 $\pm$ 0.00 & 
\cellcolor{mynicered!35} 1.00 $\pm$ 0.00 & 
\cellcolor{mynicered!35} 0.80 $\pm$ 0.00 \\ 
\forgetNot~\cite{zhang2023forgetmenot} & \cellcolor{mynicered!35} 1.00 $\pm$ 0.00 & \cellcolor{mynicered!35} 1.00 $\pm$ 0.00 & \cellcolor{mynicered!35}0.99 $\pm$ 0.00 & \cellcolor{mynicered!35}0.99 $\pm$ 0.00 & \cellcolor{mynicered!35}0.98 $\pm$ 0.00 & \cellcolor{mynicered!35}0.98 $\pm$ 0.01 & \cellcolor{mynicered!35}0.99 $\pm$ 0.00 & \cellcolor{mynicered!35}0.99 $\pm$ 0.00 & \cellcolor{mynicered!35}1.00 $\pm$ 0.00 & \cellcolor{mynicered!35}0.99 $\pm$ 0.00 & \cellcolor{mynicered!35}0.99 $\pm$ 0.00 \\ 
\sa~\cite{heng2023selective} & 
\cellcolor{mynicered!35} 0.98 $\pm$ 0.01 & 
\cellcolor{mynicered!35} 0.99 $\pm$ 0.01 & 
\cellcolor{mynicered!35} 0.99 $\pm$ 0.00 & 
\cellcolor{mynicered!35} 0.97 $\pm$ 0.01 & 
\cellcolor{mynicered!35} 1.00 $\pm$ 0.00 & 
\cellcolor{mynicered!35} 0.99 $\pm$ 0.00 & 
\cellcolor{mynicered!35} 0.99 $\pm$ 0.00 & 
\cellcolor{mynicered!35} 1.00 $\pm$ 0.00 & 
\cellcolor{mynicered!35} 0.84 $\pm$ 0.01 & 
\cellcolor{mynicered!35} 0.97 $\pm$ 0.00 & 
\cellcolor{mynicered!35} 0.81 $\pm$ 0.01 \\ 
\esd~\cite{gandikota2023erasing} & \cellcolor{mynicered!35} 0.92 $\pm$ 0.00 & 
\cellcolor{mynicered!35} 0.99 $\pm$ 0.00 & 
\cellcolor{mynicered!35}0.91 $\pm$ 0.01 & \cellcolor{mynicered!35}0.94 $\pm$ 0.00 & \cellcolor{mynicered!35}0.96 $\pm$ 0.00 & \cellcolor{mynicered!35}0.99 $\pm$ 0.00 & \cellcolor{mynicered!35}0.99 $\pm$ 0.00 & \cellcolor{mynicered!35}1.00 $\pm$ 0.00 & \cellcolor{mynicered!35}1.00 $\pm$ 0.00 & \cellcolor{mynicered!35}1.00 $\pm$ 0.00 & \cellcolor{mynicered!35}0.98 $\pm$ 0.01 \\ 
\unifiedCE~\cite{gandikota2023unified} & \cellcolor{mynicered!35} 1.00 $\pm$ 0.00 & \cellcolor{mynicered!35} 0.97 $\pm$ 0.00 & \cellcolor{mynicered!35}1.00 $\pm$ 0.00 & \cellcolor{mynicered!35}0.98 $\pm$ 0.00 & \cellcolor{mynicered!35}0.98 $\pm$ 0.01 & \cellcolor{mynicered!35}1.00 $\pm$ 0.00 & \cellcolor{mynicered!35}0.99 $\pm$ 0.00 & \cellcolor{mynicered!35}0.99 $\pm$ 0.00 & \cellcolor{mynicered!35}0.63 $\pm$ 0.01 & \cellcolor{mynicered!35}1.00 $\pm$ 0.00 & \cellcolor{mynicered!35}0.77 $\pm$ 0.01 \\ 
\sdd~\cite{kim2023towards} & 
\cellcolor{mynicered!35} 1.00 $\pm$ 0.00 & 
\cellcolor{mynicered!35} 0.81 $\pm$ 0.00 & 
\cellcolor{mynicered!35} 0.81 $\pm$ 0.00 & 
\cellcolor{mynicered!35} 0.93 $\pm$ 0.01 & 
\cellcolor{mynicered!35} 0.96 $\pm$ 0.00 & 
\cellcolor{mynicered!35} 0.98 $\pm$ 0.00 & 
\cellcolor{mynicered!35} 0.97 $\pm$ 0.01 & 
\cellcolor{mynicered!35} 0.67 $\pm$ 0.01 & 
\cellcolor{mynicered!35} 0.77 $\pm$ 0.01 & 
\cellcolor{mynicered!35} 1.00 $\pm$ 0.00 & 
\cellcolor{mynicered!35} 0.81 $\pm$ 0.01 \\ \moderator~\cite{wang2024moderator} & \cellcolor{mynicered!35} 0.99 $\pm$ 0.00 & \cellcolor{mynicered!35} 0.92 $\pm$ 0.05  & \cellcolor{mynicegreen!35} 0.02 $\pm$ 0.00 & \cellcolor{myniceblue!35} 0.18 $\pm$ 0.00 & \cellcolor{mynicered!35} 0.44 $\pm$ 0.00 & \cellcolor{mynicegreen!35} 0.03 $\pm$ 0.01 & \cellcolor{mynicered!35} 0.45 $\pm$ 0.00 & \cellcolor{mynicegreen!35} 0.00 $\pm$ 0.00 &  \cellcolor{mynicegreen!35} 0.06 $\pm$ 0.00 &  \cellcolor{mynicegreen!35} 0.00 $\pm$ 0.00 & \cellcolor{mynicegreen!35} 0.09 $\pm$ 0.00 \\ 
\midrule
\textbf{\method} & \cellcolor{mynicegreen!35} 0.00 $\pm$ 0.00 & \cellcolor{myniceblue!35} 0.40 $\pm$ 0.05  & \cellcolor{mynicegreen!35} 0.02 $\pm$ 0.00 & \cellcolor{mynicegreen!35} 0.00 $\pm$ 0.00 & \cellcolor{mynicegreen!35} 0.00 $\pm$ 0.00 & \cellcolor{mynicegreen!35} 0.00 $\pm$ 0.01 & \cellcolor{mynicegreen!35} 0.00 $\pm$ 0.00 & \cellcolor{mynicegreen!35} 0.00 $\pm$ 0.00 &  \cellcolor{mynicegreen!35} 0.00 $\pm$ 0.00 &  \cellcolor{mynicegreen!35} 0.00 $\pm$ 0.00 & \cellcolor{mynicegreen!35} 0.01 $\pm$ 0.01 \\ 
\bottomrule

\toprule
\end{tabular}
}
\label{tab:robust}
\end{table*}

Across all concepts, using fine-tuning (w/ FT) makes $\phi_x(\imgadv)$ closer to $\phi_p(\cunacc)$ than $\phi_p(\cacc)$, compared to the baseline without fine-tuning (w/o FT). Hence, fine-tuning is likely to correctly identify $\imgadv$ as $\imgunacc$.
Furthermore, we evaluate the impact of fine-tuning on \ref{effective} and \ref{utility}. For \ref{effective}, CLIP accuracy for several concepts is better after fine-tuning: \emph{Angelina Jolie} (0.03 $\pm$ 0.00 from 0.12 $\pm$ 0.02), \emph{Brad Pitt} (0.00 $\pm$ 0.00 from 0.02 $\pm$ 0.01), \emph{Elon Musk} (0.03 $\pm$ 0.00 from 0.07 $\pm$ 0.01). For \ref{utility}, CLIP scores are better for some concepts after fine-tuning: \emph{nudity} (0.94 $\pm$ 0.08 from 0.83 $\pm$ 0.07); \emph{grumpy cat} (0.98 $\pm$ 0.04 from 0.88 $\pm$ 0.03), \emph{Nemo} (0.37 $\pm$ 0.02 from 0.24 $\pm$ 0.05), \emph{R2D2} (0.66 $\pm$ 0.02 from 0.52 $\pm$ 0.03), and \emph{Elon musk} (0.97 $\pm$ 0.01 from 0.92 $\pm$ 0.02).
The metrics for other concepts remains similar.

\subsection{Comparison with Fine-tuning \crt{s}}\label{sec:finetuneEval}

\noindent\textbf{\underline{\ref{effective} Effectiveness.}} We report CLIP accuracy on $\dtestunacc$ in Table~\ref{tab:effectiveness}. 
We use \colorbox{mynicered}{red} if accuracy is $>$50; \colorbox{myniceblue}{blue} if accuracy is between 25-50; \colorbox{mynicegreen}{green} if accuracy is $<$25.
All \crt{s} exhibit poor accuracy on \emph{nudity} and \emph{violence}, likely due to fine-tuning \crt{s} being sensitive to input prompts~\cite{pham2024robust, ma2024jailbreaking}. Specifically, other \crt{s} depend on the $\cunacc$ being included in prompts, which is absent for \emph{nudity} and \emph{violence} prompts in the I2P dataset.
\method consistently maintains high accuracy on the I2P benchmark dataset as it classifies the generated images.
\unifiedCE, \sdd, and \moderator have better accuracy compared to the other four fine-tuning \crt{s}. 
For \sdd, this could be attributed to its optimization which includes fine-tuning \dm, conditioned on $\cunacc$, to match the unconditioned \dm to reduce the influence of $\cunacc$ on the output.
For \unifiedCE and \moderator, we attribute their higher effectiveness to directly removing the influence of $\cunacc$ from the parameters. \textbf{\textit{Overall, \method is more effective than other fine-tuning \crt{s}.}}

\noindent\textbf{\underline{\ref{utility} Utility.}} We report normalized CLIP scores in Table~\ref{tab:utility} on $\dtestacc$.
We use \colorbox{mynicered}{red} if score is between 50-70, \colorbox{myniceblue}{blue} if between 70-90; \colorbox{mynicegreen}{green} if $>$90.
All the fine-tuning \crt{s} perform well across all concepts (either \colorbox{myniceblue}{blue} or \colorbox{mynicegreen}{green}) since the explicitly account for \ref{utility}. We observe that \ca with KL-Divergence-based optimization for cross-attention layers, and \unifiedCE with a precise closed-form solution to model updates, preserve \ref{utility} better than others. 
\method has high utility for all concepts except for \emph{violence}, and Group-2 concepts (\emph{Nemo}, \emph{Captain Marvel}, \emph{Snoopy}, and \emph{R2D2}).
For \textit{violence}, we attribute this to the trade-off between \ref{effective} and \ref{utility} during fine-tuning: we observed an early decrease in utility during the very first epoch resulting in poor trade-off.
For Group-2 concepts, we attribute the poor utility to the ambiguity in the unacceptable concepts. For instance, \emph{Nemo} is both a fish and a ship captain~\cite{VERNE_2024}, and \emph{Captain Marvel} represents both a male and a female superhero~\cite{Friedwald_2019_marvel}. 
To verify this, we precisely specify the unacceptable concepts to reduce ambiguity: as \emph{Nemo}$\rightarrow$ \emph{Nemo fish}, \emph{Captain Marvel}$\rightarrow$ \emph{Captain Marvel female superhero}, \emph{Snoopy}$\rightarrow$ \emph{Snoopy dog}, and \emph{R2D2}$\rightarrow$ \emph{R2D2 robot}. We evaluate \method on $\dvalacc$, and
compared to the results in Table~\ref{tab:utility}, the normalized CLIP score for this new configuration is: 0.97 $\pm$ 0.00 (\emph{Nemo fish}), 0.90 $\pm$ 0.02 (\emph{Captain Marvel female superhero}), 0.98 $\pm$ 0.03 (\emph{Snoopy dog}), 0.92 $\pm$ 0.02 (\emph{R2D2 robot}), which are now labeled as \colorbox{mynicegreen}{green}.
We also report the CLIP accuracy on $\dvalunacc$ to evaluate \ref{effective} with this new configuration: 0.02 $\pm$ 0.00 (\emph{Nemo fish}), 0.00 $\pm$ 0.00 (\emph{Captain Marvel female superhero}), 0.02 $\pm$ 0.01 (\emph{Snoopy dog}), 0.00 $\pm$ 0.00 (\emph{R2D2 robot}), which are effective, same as before. Hence, precisely specifying $\cunacc$ is important for \method and can satisfy \ref{utility} without sacrificing \ref{effective}. 
\textbf{\textit{Overall, \method suffers from a small drop in utility (\ref{utility}) while performing significantly better in effectiveness (\ref{effective}).}}

\noindent\textbf{\underline{\ref{robust} Robustness.}} We report CLIP accuracy on $\dtestadv$ in Table~\ref{tab:robust}. 
We evaluate different \crt{s} against \typo+, \pez+, \cce/\cce+, and \ringbell+. We use the same color coding as in \ref{effective}. 
\cce/\cce+, being a white-box attack that uses the parameters of the entire \tti model, is the most powerful attack and renders all fine-tuning \crt{s} ineffective. However, \method is robust against \cce/\cce+ and outperforms all fine-tuning \crt{s}. While \moderator outperforms \method for \textit{Snoopy}, \textit{Grumpy Cat}, and \textit{Brad Pitt} for \pez+ and \typo+, \method outperforms on the remaining concepts and attacks.
On the remaining attacks, all the fine-tuning \crt{s} have better robustness on Group-3 concepts than on Group-1 and 2. We attribute this to the difficulty of \tti models in generating high quality celebrity faces while also evading detection, as also observed in prior work~\cite{Matthias_2023}.
As expected, we note that all \crt{s} perform poorly on \emph{nudity} and \emph{violence} across all the attacks as they do not satisfy \ref{effective}.
Hence, it is expected that adversarial prompts for these concepts will easily evade them.
\textbf{\emph{Overall, \method is more robust than all prior work across all the concepts and attacks.}}

\begin{figure}[ht]
    \centering
    \includegraphics[width=0.7\columnwidth]{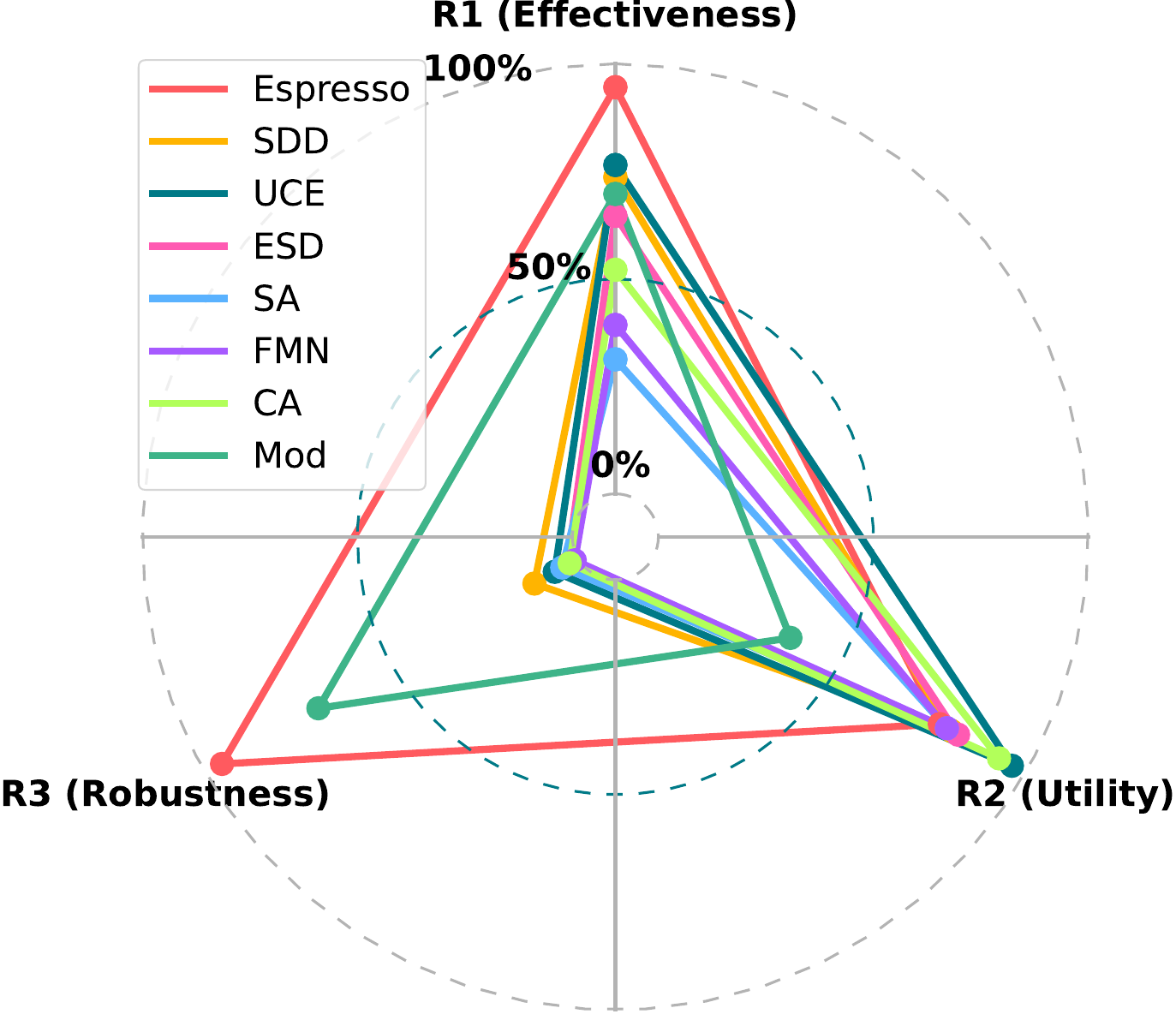}
    \caption{\method is better than other fine-tuning \crt{s}.}
    \label{fig:radarsFT}
\end{figure}

\noindent\textbf{\underline{Summary.}} 
We depict the trade-offs among \ref{effective}, \ref{utility}, and \ref{robust} in Figure~\ref{fig:radarsFT}. We use (1-CLIP accuracy) for \ref{effective} on $\dtestunacc$, normalized CLIP score on $\dtestacc$ for \ref{utility}, and (1-CLIP accuracy) for \ref{robust} on $\dtestadv$ using \cce/\cce+. 
For each of the requirements, we use the average across all concepts as the representative value for a \crt.
\emph{Overall, although \method suffers a small drop in utility \ref{utility}, it provides a better trade-off across the three requirements compared to other \crt{s}.}

\subsection{Comparison with Filtering \crt}\label{sec:filterEval}

We now compare with \unsafediff~\cite{unsafeDiff}, the state-of-the-art filtering \crt, across the three requirements and summarize the results. 
To compare with \unsafediff~\cite{unsafeDiff}, we use FNR to evaluate effectiveness on $\dtestunacc$, FPR for utility on $\dtestacc$, and FNR for robustness on $\dtestadv$.

\noindent\textbf{\underline{\ref{effective} Effectiveness.}} We report FNR across four concepts (\emph{nudity}, \emph{violence}, \emph{disturbing}, and \emph{hateful}) in Table~\ref{tab:FilterEffective}.
We use \colorbox{mynicered}{red} if FNR is $>$0.50; \colorbox{myniceblue}{blue} if FNR is between 0.25-0.50; \colorbox{mynicegreen}{green} if FNR is $<$0.25.
\method has better FNR for three of the four concepts: \emph{nudity}, \emph{violence} (in \colorbox{mynicegreen}{green}), and \emph{hateful} (\colorbox{myniceblue}{blue} for \method and \colorbox{mynicered}{red} for \unsafediff). However, both \method and \unsafediff perform poorly on \emph{disturbing}. 
We attribute this poor effectiveness on Group-1 concepts to the subjective description of $\cunacc$. Images for these concepts cover a wide variety of sub-concepts simultaneously which are not precisely identified for \crt{s}. 
\emph{\textbf{Overall, \method is more effective than \unsafediff on most concepts.}}

\begin{table}[!htb]
\centering
\caption{\underline{\ref{effective} (Effectiveness)}: Comparison with \unsafediff using FNR on \emph{unacceptable} prompts in $\dtestunacc$ (lower is better). We use \colorbox{mynicered}{red} if FNR is $>$0.50; \colorbox{myniceblue}{blue} if FNR is between 0.25-0.50; \colorbox{mynicegreen}{green} if FNR is $<$0.25.}
\footnotesize
\resizebox{0.6\columnwidth}{!}{
\begin{tabular}{  l | c c }
\bottomrule

\toprule
\textbf{Concepts}
& \textbf{\unsafediff} & \textbf{\method} \\
\midrule 
\textbf{Nudity (I2P)}  & \cellcolor{myniceblue!35}0.39 $\pm$ 0.02 & \cellcolor{mynicegreen!35}0.14 $\pm$ 0.05 \\
\textbf{Violence (I2P)}  & \cellcolor{mynicered!35}0.90 $\pm$ 0.02 & \cellcolor{mynicegreen!35}0.20 $\pm$ 0.00 \\
\textbf{Disturbing (I2P)}  & \cellcolor{mynicered!35}0.89 $\pm$ 0.03 & \cellcolor{mynicered!35}0.53 $\pm$ 0.08  \\
\textbf{Hateful (I2P)} & \cellcolor{mynicered!35} 1.00 $\pm$ 0.00 & \cellcolor{myniceblue!35}0.42 $\pm$ 0.03  \\
\bottomrule

\toprule
\end{tabular}
}
\label{tab:FilterEffective}
\end{table}

\noindent\textbf{\underline{\ref{utility} Utility.}} We present FPR in Table~\ref{tab:FilterUtility} and use \colorbox{mynicered}{red} if FPR is $>$0.50, \colorbox{myniceblue}{blue} if FPR is between 0.25-0.50; \colorbox{mynicegreen}{green} if FPR is $<$0.25. As expected, we observe that both \emph{\textbf{\method and \unsafediff have comparable utility as they demonstrate a low FPR}}. \unsafediff explicitly includes images containing $\cacc$ while training the multi-headed classifier.

\begin{table}[!htb]
\centering
\caption{\underline{\ref{utility} (Utility)}: Comparison with \unsafediff using FPR on \emph{acceptable} prompts in $\dtestacc$ (lower is better). We use \colorbox{mynicered}{red} if FPR is $>$0.50, \colorbox{myniceblue}{blue} if FPR is between 0.25-0.50; \colorbox{mynicegreen}{green} if FPR is $<$0.25.}
\footnotesize
\resizebox{0.6\columnwidth}{!}{
\begin{tabular}{  l | c c }
\bottomrule

\toprule
\textbf{Concepts} & \textbf{\unsafediff} & \textbf{\method}\\
\midrule 
\textbf{Nudity (I2P)}  & \cellcolor{mynicegreen!35}0.01 $\pm$ 0.00 & \cellcolor{mynicegreen!35}0.01 $\pm$ 0.01 \\
\textbf{Violence (I2P)} & \cellcolor{mynicegreen!35}0.01 $\pm$ 0.00 & \cellcolor{mynicegreen!35}0.08 $\pm$ 0.05 \\
\textbf{Disturbing (I2P)}  & \cellcolor{mynicegreen!35}0.01 $\pm$ 0.00 &\cellcolor{mynicegreen!35} 0.01 $\pm$ 0.01 \\
\textbf{Hateful (I2P)} & \cellcolor{mynicegreen!35}0.01 $\pm$ 0.00 & \cellcolor{mynicegreen!35}0.06 $\pm$ 0.04 \\
\bottomrule

\toprule
\end{tabular}
}
\label{tab:FilterUtility}
\end{table}

\noindent\textbf{\underline{\ref{robust} Robustness.}} We report FNR on the dataset for adversarial prompts and corresponding images in Table~\ref{tab:FilterRobust}. 
We use the same color-coding as \ref{effective}.
In addition to the four attacks from Table~\ref{tab:robust}, recall that \sneaky~\cite{yang2023sneakyprompt} is specifically designed to attack filtering \crt{s}. Hence, we also evaluate against \sneaky. 
Also, since \cce is not adaptive against filtering \crt's, we evaluate \unsafediff and \method against \cce+. We are the first to evaluate different attacks against \unsafediff.
\begin{table}[!htb]
\centering
\caption{\underline{\ref{robust} (Robustness)}: Comparison with \unsafediff using FNR on \emph{adversarial} prompts in $\dtestadv$ (lower is better). We use \colorbox{mynicered}{red} if FNR is $>$0.50; \colorbox{myniceblue}{blue} if FNR is between 0.25-0.50; and \colorbox{mynicegreen}{green} if FNR is $<$0.25.}
\footnotesize
\resizebox{0.9\columnwidth}{!}{
\begin{tabular}{  l | c | c | c | c  }
\bottomrule

\toprule
\textbf{\crt} & \textbf{Nudity} & \textbf{Violence} & \textbf{Disturbing} & \textbf{Hateful}\\
\midrule
\multicolumn{5}{c}{\textbf{\typo+}}\\
\midrule
\textbf{\unsafediff} & \cellcolor{mynicered!35} 0.55 $\pm$ 0.02 & \cellcolor{mynicered!35} 0.91 $\pm$ 0.05 & \cellcolor{myniceblue!35} 0.39 $\pm$ 0.01 & \cellcolor{myniceblue!35} 0.48 $\pm$ 0.01\\
\midrule
\textbf{\method} & \cellcolor{mynicegreen!35} 0.15 $\pm$ 0.01 & \cellcolor{mynicegreen!35} 0.26 $\pm$ 0.01 &  \cellcolor{myniceblue!35} 0.39 $\pm$ 0.01 & \cellcolor{myniceblue!35} 0.37 $\pm$ 0.05 \\
\midrule
\multicolumn{5}{c}{\textbf{\pez+}}\\
\midrule
\textbf{\unsafediff}  & \cellcolor{mynicered!35} 0.65 $\pm$ 0.02 & \cellcolor{mynicered!35} 0.91 $\pm$ 0.02  & \cellcolor{mynicered!35} 0.89 $\pm 0.02$ & \cellcolor{mynicered!35} 1.00 $\pm 0.00$\\
\midrule
\textbf{\method}& \cellcolor{mynicegreen!35} 0.16 $\pm$ 0.02 & \cellcolor{mynicegreen!35} 0.25 $\pm$ 0.04  & \cellcolor{mynicegreen!35} 0.14 $\pm$ 0.03 & \cellcolor{mynicegreen!35} 0.20 $\pm$ 0.05\\
\midrule
\multicolumn{5}{c}{\textbf{\cce}}\\
\midrule
\textbf{\unsafediff} & \cellcolor{mynicegreen!35} 0.00 $\pm$ 0.00 & \cellcolor{mynicered!35} 0.75 $\pm$ 0.05 & \cellcolor{mynicered!35} 1.00 $\pm$ 0.05 & \cellcolor{mynicered!35} 1.00 $\pm$ 0.00\\
\midrule
\textbf{\method} & \cellcolor{mynicegreen!35} 0.00 $\pm$ 0.00 & \cellcolor{myniceblue!35} 0.38 $\pm$ 0.05  & \cellcolor{mynicegreen!35} 0.02 $\pm$ 0.01 & \cellcolor{mynicegreen!35} 0.02 $\pm$ 0.01 \\
\midrule
\multicolumn{5}{c}{\textbf{\ringbell+}}\\
\midrule
\textbf{\unsafediff}  & \cellcolor{mynicered!35} 0.95 $\pm$ 0.03 & \cellcolor{myniceblue!35} 0.50 $\pm$ 0.04 & \cellcolor{myniceblue!35} 0.30 $\pm$ 0.05 & \cellcolor{mynicered!35} 0.90 $\pm$ 0.05\\
\midrule
\textbf{\method} & \cellcolor{mynicegreen!35} 0.06 $\pm$ 0.08 & \cellcolor{mynicegreen!35} 0.08 $\pm$ 0.01 & \cellcolor{mynicegreen!35} 0.06 $\pm$ 0.02 & \cellcolor{myniceblue!35} 0.25 $\pm$ 0.05\\
\midrule
\multicolumn{5}{c}{\textbf{\sneaky}}\\
\midrule
\textbf{\unsafediff}  & \cellcolor{mynicered!35} 0.67 $\pm$ 0.21 &  \cellcolor{mynicered!35} 0.71 $\pm$ 0.02 & \cellcolor{mynicered!35} 0.82 $\pm$ 0.03 & \cellcolor{myniceblue!35} 0.48 $\pm$ 0.06\\
\midrule
\textbf{\method} & \cellcolor{myniceblue!35}0.40 $\pm$ 0.08  & \cellcolor{mynicegreen!35}0.14 $\pm$ 0.03 & \cellcolor{mynicered!35} 0.70 $\pm$ 0.05 & \cellcolor{mynicegreen!35} 0.15 $\pm$ 0.10\\
\bottomrule

\toprule
\end{tabular}
}
\label{tab:FilterRobust}
\end{table}

We observe that \method is effective against \pez+ while \unsafediff can be evaded.
On all the remaining attacks, we show that \method is robust on more concepts than \unsafediff.
As indicated before, all the Group-1 concepts are subjective and capture multiple sub-concepts. This ambiguity could be the reason for poor \ref{robust}.
\emph{\textbf{Overall, \method is more robust than \unsafediff on a majority of the concepts across various attacks.}}

\begin{figure}[h]
    \centering
    \includegraphics[width=0.7\columnwidth]{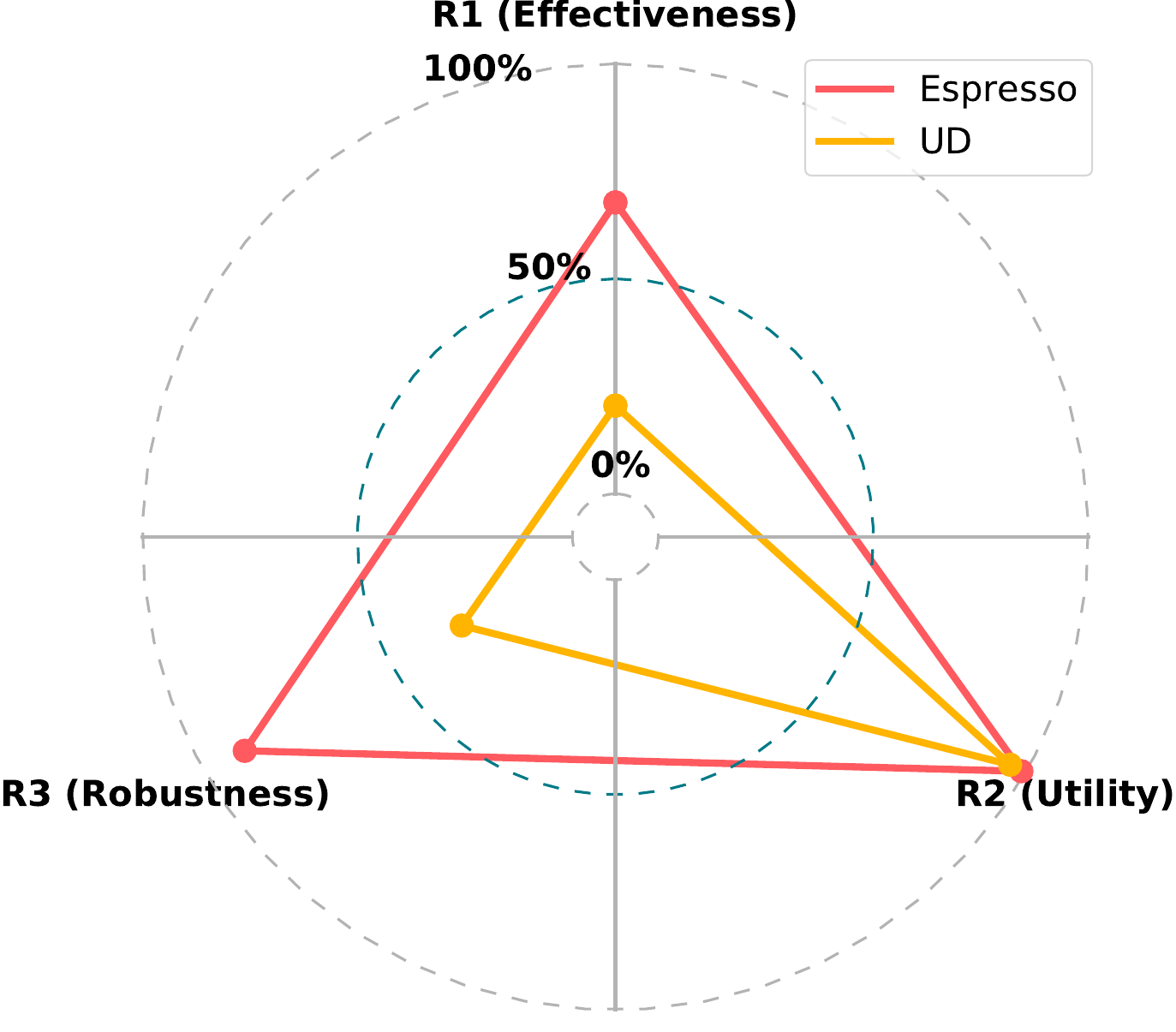}
    \caption{\method has a better trade-off than \unsafediff filter.}
    \label{fig:radarsFilt}
\end{figure}

\noindent\textbf{\underline{Summary.}} 
We present the trade-offs in the form of radar plots in Figure~\ref{fig:radarsFilt}. 
We use (1- FNR) on $\dtestunacc$ for \ref{effective}, (1- FPR) on $\dtestacc$ for \ref{utility}, and (1- FNR) on $\dtestadv$ using CCE+ for \ref{robust}. Hence, a higher value indicates better performance.
\method has comparable utility to \unsafediff but outperforms in effectiveness and robustness.
\emph{Overall, \method covers a larger area, and hence provides a better trade-off than \unsafediff.}

\section{Discussion and Future Work}\label{sec:discussions}


\subsection{Certifying Robustness of \method}
Inspired by the literature on certified robustness against adversarial examples~\cite{cohen2019certified}, it is natural to ask whether a similar notion of certified robustness is possible for \crt{s}.
None of the existing \crt{s} have considered certified robustness.
We are the first to explore its feasibility for \method.
We first present a theoretical bound on the worst-case input modification by \adv under which we can guarantee \method's accuracy. 



\noindent\emph{Theoretical Bound.} Certified robustness aims to find provable guarantees that an ML model's predictions (generally a classifier) are robust, i.e., the predictions do not change on adding noise to the input~\cite{carlini2023certified}. 
Our goal is to have a similar robustness bound for a \tti model with \method. 
We want to find the maximum input noise which \method can tolerate.

We give advantage to \adv by assuming they can directly add adversarial noise to \method's embeddings. This is a strong assumption as in practice, \adv can only send prompts to the \tti model. \begin{arxiv}
We revisit this assumption later in Appendix~\autoref{sec:implications}.    
\end{arxiv}
\begin{submission}
We discuss this assumption in the Appendix of our full version of our paper~\cite{das2024espresso}.
\end{submission}

Formally, given an unacceptable image $\imgunacc$, \adv adds noise $\delta$ to its embeddings, $\phi_x(\imgunacc)$, such that $\filter(\phi_x(\imgunacc) + \delta)$ is misclassified as acceptable.
Using this assumption, we specify the maximum noise $\delta$ added to the embeddings, $\phi_x(\imgunacc)$, that \method can tolerate in Theorem~\ref{thm.1}.

\begin{theorem}~\label{thm.1}
    Let $\hat{x} = \phi_x(x), \hat{c}^i = \phi_p(c^i), i\in\{a,u\}$. 
    Define 
    \[g_i(\hat{x}):=\frac{exp(s(\hat{x},\hat{c}^i))}{exp(s(\hat{x},\caccc))+exp(s(\hat{x},\cunaccc))},\] 
    where $s(\hat{x},\hat{c}^i) = \tau \text{cos}(\hat{x},\hat{c}^i))$, then $g_i$ is the confidence of $\hat{x}$ being classified as $\concept^i$. $F(x)$ can be defined as $F(\hat{x})=argmax_i g_i(\hat{x})$, and $F(\hat{x})$ classifies $\hat{x}$ as unacceptable if $g_u(\hat{x})>\Gamma$, where $\Gamma$ is the decision threshold.
     For a given image embedding $\hat{x}$, if $g(\hat{x}):=g_u(\hat{x})>\Gamma$, then $g$ is robust against noise $\delta$ where 
     \[||\delta||\leq\left(1-\frac{\tau }{\tau +2|g(\hat{x})-\Gamma|}\right)||\hat{x}||,\]
     and $\Gamma$ is the decision threshold i.e. 
    \begin{equation}\label{eq:bound}
            F(\hat{x}) = F(\hat{x}+\delta), \forall ||\delta||\leq \left(1-\frac{\tau }{\tau +2|g(\hat{x})-\Gamma|}\right)||\hat{x}||.
    \end{equation}
\end{theorem}

\begin{proof}
    For an unacceptable image embedding $\hat{x} = \phi_x(\imgunacc)$,
    $g(\hat{x}):=g_u(\hat{x})$, then $g(\hat{x})-\Gamma>0$, and $\Gamma$ is the decision threshold for classification.
    Let $s_1=\tau\text{cos}(\hat{x},\cunaccc)$, $s_2=\tau\text{cos}(\hat{x},\caccc)$, $\textbf{s}=[s_1,s_2]^T$, 
    then
    $$g(\hat{x})=S(s_1) = \frac{exp(s_1)}{exp(s_1)+exp(s_2)},$$
    where $S(s_1)$ is the first item of Softmax function with respect to $\textbf{s}$.
    Then, we have $\frac{\partial}{\partial s_1} S = S(s_1)(1-S(s_1))\leq0.25, \frac{\partial}{\partial s_2} S = -S(s_1)S(s_2)\leq0.25$.

    Note that $||\hat{x}||>0$ and $||\caccc||>0$,
    we have 
    \begin{equation*}
    \begin{aligned}
        ||\frac{\partial}{\partial \hat{x}} s(\hat{x},\caccc)|| = ||\frac{\tau||\caccc||(I-xx^T)\caccc}{||\hat{x}||||\caccc||^2}||=\frac{\tau sin(\hat{x},\caccc)}{||\hat{x}||}\leq\frac{\tau}{||\hat{x}||}
    \end{aligned}
    \end{equation*}
    where $||\frac{\partial}{\partial\hat{x}} s(\hat{x}, \cunaccc)||\leq\frac{\tau}{||\hat{x}||}$. 

    For each $\hat{x}$, according to the chain rule of composition functions, 
    $\frac{\partial}{\partial\hat{x}}g(\hat{x)}=\frac{\partial S}{\partial s_1}\cdot\frac{\partial s_1}{\hat{x}}+ \frac{\partial S}{\partial s_2}\cdot\frac{\partial s_2}{\hat{x}}\leq \frac{\tau}{2||\hat{x}||}$. Therefore, the Lipschitz constant of $g(\hat{x})$ with respect to $\hat{x}$ is $\frac{\tau}{2||\hat{x}||}$, and
    \begin{equation*}
    \begin{aligned}
            ||g(\hat{x}+\delta)-g(\hat{x})||&\leq\frac{1}{2}\frac{\tau }{\min\{||u|||u\in U(\hat{x},\delta)\}} ||\delta||\\&\leq\frac{1}{2}\frac{\tau }{|||\hat{x}||-||\delta|||}||\delta||,
    \end{aligned}
    \end{equation*}
    where $U(\hat{x},\delta)$ is a $l_2$-ball of $\hat{x}$ with radius $\delta$.   
    
    When
    $||\delta||\leq(1-\frac{\tau }{\tau +2|g(\hat{x})-\Gamma|})||\hat{x}||<||\hat{x}||$, we have
    \begin{equation*}
    \begin{aligned}
        |g(\hat{x}+\delta)-g(\hat{x})|&=||g(\hat{x}+\delta)-g(\hat{x})||\\& \leq\frac{\tau }{2\left(\frac{||\hat{x}||}{||\delta||}-1\right)}\\
        &\leq\frac{\tau }{2\left(\frac{\tau +2|g(\hat{x})-\Gamma|}{2|g(\hat{x})-\Gamma|}-1\right)}\\
        &\leq|g(\hat{x})-\Gamma|=g(\hat{x})-\Gamma.
    \end{aligned}
    \end{equation*}
    Then,
    \begin{equation*}
    \begin{aligned}
        g(\hat{x}+\delta)\geq|g(\hat{x})|-|g(\hat{x}+\delta)-g(\hat{x})|
        \geq g(\hat{x})- |g(\hat{x})-\Gamma|\geq \Gamma,
    \end{aligned}
    \end{equation*}
    which concludes the proof.
\end{proof}

The proof says that when the noise does not exceed a certain bound, which is a fraction of $x$, \method is able to predict the noisy samples the same as the original samples. 
We then empirically evaluate this bound on different concepts, and discuss its practical implications in \begin{submission}
Appendix in our full version~\cite{das2024espresso}.    
\end{submission}
\begin{arxiv}  
\autoref{sec:bound}.
\end{arxiv}
We discuss that \method is likely more robust in practice as our certified robustness bound is loose. We leave improving the bound as future work.


\subsection{Other Discussions}

\noindent\textbf{Addressing Future Attacks.} Recall that \method, fine-tuned using Equation~\ref{eq:ft1} and~\ref{eq:ft2}, maintains high robustness across all evaluated concepts. 
The design of new attacks which can preserve unacceptable concepts while evading \method is an open problem. For additional robustness against new attacks, in addition to Equation~\ref{eq:ft1} and~\ref{eq:ft2}, we can use the objective function for adversarial training:
\[\mathcal{L}_{\text{adv}}(\dtestadv)= - \frac{1}{N} \sum_{j = 1}^N \log \frac{\exp(\text{cos}(\phi_x(\imgadv_j), \phi_p(\cunacc))/ \tau)}{\sum_{k \in \{a,u\}} \exp(\text{cos}(\phi_x(\imgadv_j), \phi_p(c_k))/ \tau)}\]
where $\mathcal{D}_{adv} = \{(\imgadv_j, \advprompt_j)\}_{j=1}^N$. Assuming $\imgadv$ evades \method, we optimize $\phi_p$ and $\phi_x$, such that $\phi_p(\imgadv_j)$ is closer to $\phi_p(\cunacc)$ than $\phi_p(\cacc)$. 
Empirical evaluation of adversarially-trained \method is left as future work.




\noindent\textbf{Systematic Generation of Acceptable Concepts.} Given a $\cunacc$, candidates for $\cacc$ can be generated using a thesaurus API~\footnote{https://dictionaryapi.com/}. For Group-1 concepts, antonyms of $\cunacc$ are used, while synonyms are considered for others. However, this may not apply to proper nouns, which may be missing from the thesaurus.

\noindent\textbf{Classification of Acceptable Images closer to Decision Boundary.} Acceptable images close to the decision boundary of a filtering \crt{s} can mistakenly be blocked, causing a false positive. However, we argue that this is different than \ref{utility}, which is meant to measure the effect of \method on the \emph{typical} query distributions that a \tti model is likely to encounter in practice (e.g., COCO dataset used in our evaluation).
Evaluation of such acceptable images is missing in the literature and left for future work.

\noindent\textbf{Generalization to Other Concepts.} Our evaluation across 11 concepts aimed to assess \method's performance on different concept types (Group 1 vs. 2 vs. 3). Initial results suggested updating the \method objective for Group 1 concepts (\autoref{eq:ft2}) compared to Groups 2 and 3 (\autoref{eq:ft1}). We expect new concepts within the same group to behave similarly. Understanding the properties of concepts resulting in different behaviors is left as future work.

\noindent\textbf{Filtering Multiple Concepts.} UCE~\cite{gandikota2023unified} and Moderator~\cite{wang2024moderator} remove multiple concepts simultaneously.
\method can be extended by including multiple concepts simultaneously as well. Specifically, for \method in Equation~\ref{eq:att}, instead of specifying $\cunacc$ and $\cacc$ for a single concept, we can include $\cunacc$ and $\cacc$ for multiple concepts as a list. This is a simple configuration change compared to other filters which require retraining (e.g.,~\cite{unsafeDiff}). 
We leave this as future work.


\noindent\textbf{Applicability to other \tti Models.} Fine-tuning \crt{s} are specific to particular stable diffusion models due to their tailored optimizations for \tti models. In contrast, filtering \crt{s} offer versatility as they can be applicable to any \tti model. 
Filters analyze only the generated image and the list of concepts, independent of the \tti model. 
And fine-tuning the filter using data from \tti model, as we do for \method, can satisfy \ref{effective} and \ref{utility}. 
This allows us to have a filter that will work with \tti models in different domains (e.g., anime images).
Explicit evaluation of \method for different \tti models is deferred to future work.

\noindent\textbf{Efficiency of \crt{s}.}
We report the execution time for fine-tuning or training the \crt{s} (average across ten runs) on a single NVIDIA A100 GPU (Table~\ref{tab:efficiency}). For related work, the hyperparameter configuration for fine-tuning/training is the same as specified by their respective paper to satisfy effectiveness and utility.

\begin{table}[!htb]
\centering
\caption{\crt training time averaged across ten runs.}
\footnotesize
\begin{tabular}{  l c | l c  }
\bottomrule

\toprule
\textbf{Technique} & \textbf{Time (mins)} & \textbf{Technique} & \textbf{Time (mins)}\\
\midrule
 \texttt{CA}~\cite{kumari2023ablating}  & 60.03 $\pm$ 0.01 & \unifiedCE~\cite{gandikota2023unified} & 0.24 $\pm$ 0.02 \\ 
  \sa~\cite{heng2023selective} & 95.10 $\pm$ 2.21  &  \esd~\cite{gandikota2023erasing} & 125.50 $\pm$ 0.00 \\
\sdd~\cite{kim2023towards}  & 75.50 $\pm$ 3.21 & \unsafediff~\cite{unsafeDiff} & 10.00 $\pm$ 2.03\\  
 \forgetNot~\cite{zhang2023forgetmenot} & 2.20 $\pm$ 0.01 &  \moderator~\cite{wang2024moderator} & 135.25 $\pm$ 4.10\\
 \textbf{\method} & 9.10 $\pm$ 0.05 & & \\
\bottomrule

\toprule
\end{tabular}
\label{tab:efficiency}
\end{table}

We see that \method is reasonably fast to train.
For fine-tuning \crt{s}, inference time is identical to using the baseline SD v1.4 because they do not add any additional components to the \tti generation process. The inference time for filtering \crt{s} is marginally higher (+0.01\%) than the baseline (of only the image generation time taken by the \tti model). 

\begin{arxiv}
\subsection{Summary}
    
Removing unacceptable concepts from \tti models is crucial, but no prior \crt{s} simultaneously meet all three requirements: preventing unacceptable concepts, preserving utility on other concepts, and ensuring robustness against evasion. We propose \method, the \emph{first} robust \underline{co}ncept \underline{fi}ltering \crt which provides a better trade-off among the three requirements compared to prior state-of-the-art \crt{s}.
\end{arxiv}

\section*{Acknowledgments}
This work is supported in part by the Government of Ontario. 
Vasisht is supported by IBM PhD Fellowship, David R. Cheriton Scholarship, and Cybersecurity and Privacy Excellence Graduate Scholarship. Views expressed in the paper are those of the authors and do not necessarily reflect the position of the funding agencies.

\bibliographystyle{ACM-Reference-Format}
\bibliography{paperL}


\begin{thebibliography}{54}


\ifx \showCODEN    \undefined \def \showCODEN     #1{\unskip}     \fi
\ifx \showDOI      \undefined \def \showDOI       #1{#1}\fi
\ifx \showISBNx    \undefined \def \showISBNx     #1{\unskip}     \fi
\ifx \showISBNxiii \undefined \def \showISBNxiii  #1{\unskip}     \fi
\ifx \showISSN     \undefined \def \showISSN      #1{\unskip}     \fi
\ifx \showLCCN     \undefined \def \showLCCN      #1{\unskip}     \fi
\ifx \shownote     \undefined \def \shownote      #1{#1}          \fi
\ifx \showarticletitle \undefined \def \showarticletitle #1{#1}   \fi
\ifx \showURL      \undefined \def \showURL       {\relax}        \fi
\providecommand\bibfield[2]{#2}
\providecommand\bibinfo[2]{#2}
\providecommand\natexlab[1]{#1}
\providecommand\showeprint[2][]{arXiv:#2}

\bibitem[Carlini et~al\mbox{.}(2023a)]%
        {carlini2023extracting}
\bibfield{author}{\bibinfo{person}{Nicholas Carlini}, \bibinfo{person}{Jamie
  Hayes}, \bibinfo{person}{Milad Nasr}, \bibinfo{person}{Matthew Jagielski},
  \bibinfo{person}{Vikash Sehwag}, \bibinfo{person}{Florian Tram\`{e}r},
  \bibinfo{person}{Borja Balle}, \bibinfo{person}{Daphne Ippolito}, {and}
  \bibinfo{person}{Eric Wallace}.} \bibinfo{year}{2023}\natexlab{a}.
\newblock \showarticletitle{Extracting training data from diffusion models}. In
  \bibinfo{booktitle}{\emph{Proceedings of the 32nd USENIX Conference on
  Security Symposium}} (Anaheim, CA, USA) \emph{(\bibinfo{series}{SEC '23})}.
  \bibinfo{publisher}{USENIX Association}, \bibinfo{address}{USA}, Article
  \bibinfo{articleno}{294}, \bibinfo{numpages}{18}~pages.
\newblock
\showISBNx{978-1-939133-37-3}


\bibitem[Carlini and Terzis(2022)]%
        {carlini2022poisoning}
\bibfield{author}{\bibinfo{person}{Nicholas Carlini} {and}
  \bibinfo{person}{Andreas Terzis}.} \bibinfo{year}{2022}\natexlab{}.
\newblock \showarticletitle{Poisoning and Backdooring Contrastive Learning}. In
  \bibinfo{booktitle}{\emph{International Conference on Learning
  Representations}}. \bibinfo{publisher}{OpenReview.net}.
\newblock
\urldef\tempurl%
\url{https://openreview.net/forum?id=iC4UHbQ01Mp}
\showURL{%
\tempurl}


\bibitem[Carlini et~al\mbox{.}(2023b)]%
        {cohen2019certified}
\bibfield{author}{\bibinfo{person}{Nicholas Carlini}, \bibinfo{person}{Florian
  Tramer}, \bibinfo{person}{Krishnamurthy~Dj Dvijotham},
  \bibinfo{person}{Leslie Rice}, \bibinfo{person}{Mingjie Sun}, {and}
  \bibinfo{person}{J~Zico Kolter}.} \bibinfo{year}{2023}\natexlab{b}.
\newblock \showarticletitle{(Certified!!) Adversarial Robustness for Free!}. In
  \bibinfo{booktitle}{\emph{ICLR}}.
\newblock
\urldef\tempurl%
\url{https://openreview.net/forum?id=JLg5aHHv7j}
\showURL{%
\tempurl}


\bibitem[Carlini et~al\mbox{.}(2023c)]%
        {carlini2023certified}
\bibfield{author}{\bibinfo{person}{Nicholas Carlini}, \bibinfo{person}{Florian
  Tramer}, \bibinfo{person}{Krishnamurthy~Dj Dvijotham},
  \bibinfo{person}{Leslie Rice}, \bibinfo{person}{Mingjie Sun}, {and}
  \bibinfo{person}{J~Zico Kolter}.} \bibinfo{year}{2023}\natexlab{c}.
\newblock \showarticletitle{(Certified!!) Adversarial Robustness for Free!}. In
  \bibinfo{booktitle}{\emph{The Eleventh International Conference on Learning
  Representations}}. \bibinfo{publisher}{OpenReview.net}.
\newblock
\urldef\tempurl%
\url{https://openreview.net/forum?id=JLg5aHHv7j}
\showURL{%
\tempurl}


\bibitem[Friedwald(2019)]%
        {Friedwald_2019_marvel}
\bibfield{author}{\bibinfo{person}{Will Friedwald}.}
  \bibinfo{year}{2019}\natexlab{}.
\newblock \bibinfo{title}{Captain Marvel vs. captain Marvel: The strange tale
  of two dueling superheroes}.
\newblock
\newblock
\urldef\tempurl%
\url{https://www.vanityfair.com/hollywood/2019/03/captain-marvel-shazam-carol-danvers-guide}
\showURL{%
\tempurl}


\bibitem[Gal et~al\mbox{.}(2023)]%
        {gal2022textualinversionimage}
\bibfield{author}{\bibinfo{person}{Rinon Gal}, \bibinfo{person}{Yuval Alaluf},
  \bibinfo{person}{Yuval Atzmon}, \bibinfo{person}{Or Patashnik},
  \bibinfo{person}{Amit~Haim Bermano}, \bibinfo{person}{Gal Chechik}, {and}
  \bibinfo{person}{Daniel Cohen-or}.} \bibinfo{year}{2023}\natexlab{}.
\newblock \showarticletitle{An Image is Worth One Word: Personalizing
  Text-to-Image Generation using Textual Inversion}. In
  \bibinfo{booktitle}{\emph{The Eleventh International Conference on Learning
  Representations}}. \bibinfo{publisher}{OpenReview.net}.
\newblock
\urldef\tempurl%
\url{https://openreview.net/forum?id=NAQvF08TcyG}
\showURL{%
\tempurl}


\bibitem[Gal et~al\mbox{.}(2022)]%
        {gal2021stylegannada}
\bibfield{author}{\bibinfo{person}{Rinon Gal}, \bibinfo{person}{Or Patashnik},
  \bibinfo{person}{Haggai Maron}, \bibinfo{person}{Amit~H. Bermano},
  \bibinfo{person}{Gal Chechik}, {and} \bibinfo{person}{Daniel Cohen-Or}.}
  \bibinfo{year}{2022}\natexlab{}.
\newblock \showarticletitle{StyleGAN-NADA: CLIP-guided domain adaptation of
  image generators}.
\newblock \bibinfo{journal}{\emph{ACM Trans. Graph.}} \bibinfo{volume}{41},
  \bibinfo{number}{4}, Article \bibinfo{articleno}{141},
  \bibinfo{numpages}{13}~pages.
\newblock
\showISSN{0730-0301}
\urldef\tempurl%
\url{https://doi.org/10.1145/3528223.3530164}
\showDOI{\tempurl}


\bibitem[Gandikota et~al\mbox{.}(2023a)]%
        {gandikota2023erasing}
\bibfield{author}{\bibinfo{person}{Rohit Gandikota}, \bibinfo{person}{Joanna
  Materzynska}, \bibinfo{person}{Jaden Fiotto-Kaufman}, {and}
  \bibinfo{person}{David Bau}.} \bibinfo{year}{2023}\natexlab{a}.
\newblock \showarticletitle{Erasing Concepts from Diffusion Models}. In
  \bibinfo{booktitle}{\emph{Proceedings of the IEEE/CVF International
  Conference on Computer Vision (ICCV)}}. \bibinfo{publisher}{IEEE Computer
  Society}, \bibinfo{pages}{2426--2436}.
\newblock


\bibitem[Gandikota et~al\mbox{.}(2023b)]%
        {githubGitHubRohitgandikotaerasing}
\bibfield{author}{\bibinfo{person}{Rohit Gandikota}, \bibinfo{person}{Joanna
  Materzy\'nska}, \bibinfo{person}{Jaden Fiotto-Kaufman}, {and}
  \bibinfo{person}{David Bau}.} \bibinfo{year}{2023}\natexlab{b}.
\newblock \bibinfo{title}{{G}it{H}ub - {E}rasing {C}oncepts from {D}iffusion
  {M}odels}.
\newblock
  \bibinfo{howpublished}{\url{https://github.com/rohitgandikota/erasing}}.
\newblock


\bibitem[Gandikota et~al\mbox{.}(2023c)]%
        {githubGitHubRohitgandikotaunifiedconceptediting}
\bibfield{author}{\bibinfo{person}{Rohit Gandikota}, \bibinfo{person}{Hadas
  Orgad}, \bibinfo{person}{Yonatan Belinkov}, \bibinfo{person}{Joanna
  Materzy\'nska}, {and} \bibinfo{person}{David Bau}.}
  \bibinfo{year}{2023}\natexlab{c}.
\newblock \bibinfo{title}{{G}it{H}ub - {U}nified {C}oncept {E}diting in
  {D}iffusion {M}odels}.
\newblock
  \bibinfo{howpublished}{\url{https://github.com/rohitgandikota/unified-concept-editing/tree/main}}.
\newblock


\bibitem[Gandikota et~al\mbox{.}(2024)]%
        {gandikota2023unified}
\bibfield{author}{\bibinfo{person}{Rohit Gandikota}, \bibinfo{person}{Hadas
  Orgad}, \bibinfo{person}{Yonatan Belinkov}, \bibinfo{person}{Joanna
  Materzy\'nska}, {and} \bibinfo{person}{David Bau}.}
  \bibinfo{year}{2024}\natexlab{}.
\newblock \showarticletitle{Unified Concept Editing in Diffusion Models}. In
  \bibinfo{booktitle}{\emph{Proceedings of the IEEE/CVF Winter Conference on
  Applications of Computer Vision (WACV)}}. \bibinfo{publisher}{IEEE Computer
  Society}, \bibinfo{pages}{5111--5120}.
\newblock


\bibitem[Heng and Soh(2023a)]%
        {githubGitHubClearnusselectiveamnesia}
\bibfield{author}{\bibinfo{person}{Alvin Heng} {and} \bibinfo{person}{Harold
  Soh}.} \bibinfo{year}{2023}\natexlab{a}.
\newblock \bibinfo{title}{{G}it{H}ub - selective-amnesia}.
\newblock
  \bibinfo{howpublished}{\url{https://github.com/clear-nus/selective-amnesia/tree/main}}.
\newblock


\bibitem[Heng and Soh(2023b)]%
        {heng2023selective}
\bibfield{author}{\bibinfo{person}{Alvin Heng} {and} \bibinfo{person}{Harold
  Soh}.} \bibinfo{year}{2023}\natexlab{b}.
\newblock \showarticletitle{Selective Amnesia: A Continual Learning Approach to
  Forgetting in Deep Generative Models}. In
  \bibinfo{booktitle}{\emph{Thirty-seventh Conference on Neural Information
  Processing Systems}}. \bibinfo{publisher}{OpenReview.net}.
\newblock
\urldef\tempurl%
\url{https://openreview.net/forum?id=BC1IJdsuYB}
\showURL{%
\tempurl}


\bibitem[Hessel et~al\mbox{.}(2021)]%
        {hessel-etal-2021-clipscore}
\bibfield{author}{\bibinfo{person}{Jack Hessel}, \bibinfo{person}{Ari
  Holtzman}, \bibinfo{person}{Maxwell Forbes}, \bibinfo{person}{Ronan Le~Bras},
  {and} \bibinfo{person}{Yejin Choi}.} \bibinfo{year}{2021}\natexlab{}.
\newblock \showarticletitle{{CLIPS}core: A Reference-free Evaluation Metric for
  Image Captioning}. In \bibinfo{booktitle}{\emph{Proceedings of the 2021
  Conference on Empirical Methods in Natural Language Processing}},
  \bibfield{editor}{\bibinfo{person}{Marie-Francine Moens},
  \bibinfo{person}{Xuanjing Huang}, \bibinfo{person}{Lucia Specia}, {and}
  \bibinfo{person}{Scott Wen-tau Yih}} (Eds.). \bibinfo{publisher}{Association
  for Computational Linguistics}, \bibinfo{address}{Online and Punta Cana,
  Dominican Republic}, \bibinfo{pages}{7514--7528}.
\newblock
\urldef\tempurl%
\url{https://doi.org/10.18653/v1/2021.emnlp-main.595}
\showDOI{\tempurl}


\bibitem[HuggingFace(2021)]%
        {vit}
\bibfield{author}{\bibinfo{person}{HuggingFace}.}
  \bibinfo{year}{2021}\natexlab{}.
\newblock \bibinfo{title}{openai/clip-vit-large-patch14 · Hugging Face}.
\newblock
\newblock
\urldef\tempurl%
\url{https://huggingface.co/openai/clip-vit-large-patch14}
\showURL{%
\tempurl}


\bibitem[Ilharco et~al\mbox{.}(2023)]%
        {ilharco2023editing}
\bibfield{author}{\bibinfo{person}{Gabriel Ilharco},
  \bibinfo{person}{Marco~Tulio Ribeiro}, \bibinfo{person}{Mitchell Wortsman},
  \bibinfo{person}{Ludwig Schmidt}, \bibinfo{person}{Hannaneh Hajishirzi},
  {and} \bibinfo{person}{Ali Farhadi}.} \bibinfo{year}{2023}\natexlab{}.
\newblock \showarticletitle{Editing models with task arithmetic}. In
  \bibinfo{booktitle}{\emph{The Eleventh International Conference on Learning
  Representations}}. \bibinfo{publisher}{OpenReview.net}.
\newblock
\urldef\tempurl%
\url{https://openreview.net/forum?id=6t0Kwf8-jrj}
\showURL{%
\tempurl}


\bibitem[Kim et~al\mbox{.}(2022)]%
        {Kim_2022_CVPR}
\bibfield{author}{\bibinfo{person}{Gwanghyun Kim} {et~al\mbox{.}}}
  \bibinfo{year}{2022}\natexlab{}.
\newblock \showarticletitle{DiffusionCLIP: Text-Guided Diffusion Models for
  Robust Image Manipulation}. In \bibinfo{booktitle}{\emph{CVPR}}.
  \bibinfo{pages}{2426--2435}.
\newblock


\bibitem[Kim et~al\mbox{.}(2023)]%
        {kim2023towards}
\bibfield{author}{\bibinfo{person}{Sanghyun Kim}, \bibinfo{person}{Seohyeon
  Jung}, \bibinfo{person}{Balhae Kim}, \bibinfo{person}{Moonseok Choi},
  \bibinfo{person}{Jinwoo Shin}, {and} \bibinfo{person}{Juho Lee}.}
  \bibinfo{year}{2023}\natexlab{}.
\newblock \showarticletitle{Towards safe self-distillation of internet-scale
  text-to-image diffusion models}. In \bibinfo{booktitle}{\emph{ICML 2023
  Workshop on Challenges in Deployable Generative AI}}.
\newblock


\bibitem[Kim et~al\mbox{.}(2024)]%
        {githubGitHubNannullnasafediffusion}
\bibfield{author}{\bibinfo{person}{Sanghyun Kim}, \bibinfo{person}{Seohyeon
  Jung}, \bibinfo{person}{Balhae Kim}, \bibinfo{person}{Moonseok Choi},
  \bibinfo{person}{Jinwoo Shin}, {and} \bibinfo{person}{Juho Lee}.}
  \bibinfo{year}{2024}\natexlab{}.
\newblock \bibinfo{title}{{G}it{H}ub - safe-diffusion}.
\newblock
  \bibinfo{howpublished}{\url{https://github.com/nannullna/safe-diffusion}}.
\newblock


\bibitem[Korn(2023)]%
        {Korn_2023}
\bibfield{author}{\bibinfo{person}{Jennifer Korn}.}
  \bibinfo{year}{2023}\natexlab{}.
\newblock \showarticletitle{Getty Images suing the makers of popular AI art
  tool for allegedly stealing photos}. In \bibinfo{booktitle}{\emph{CNN
  Business}}. \bibinfo{publisher}{CNN}.
\newblock
\urldef\tempurl%
\url{https://www.cnn.com/2023/01/17/tech/getty-images-stability-ai-lawsuit/index.html}
\showURL{%
\tempurl}


\bibitem[Kumari et~al\mbox{.}(2023a)]%
        {kumari2023ablating}
\bibfield{author}{\bibinfo{person}{Nupur Kumari}, \bibinfo{person}{Bingliang
  Zhang}, \bibinfo{person}{Sheng-Yu Wang}, \bibinfo{person}{Eli Shechtman},
  \bibinfo{person}{Richard Zhang}, {and} \bibinfo{person}{Jun-Yan Zhu}.}
  \bibinfo{year}{2023}\natexlab{a}.
\newblock \showarticletitle{Ablating Concepts in Text-to-Image Diffusion
  Models}. In \bibinfo{booktitle}{\emph{International Conference on Computer
  Vision (ICCV)}}. \bibinfo{publisher}{IEEE Computer Society}.
\newblock


\bibitem[Kumari et~al\mbox{.}(2023b)]%
        {githubGitHubNupurkmr9conceptablation}
\bibfield{author}{\bibinfo{person}{Nupur Kumari}, \bibinfo{person}{Bingliang
  Zhang}, \bibinfo{person}{Sheng-Yu Wang}, \bibinfo{person}{Eli Shechtman},
  \bibinfo{person}{Richard Zhang}, {and} \bibinfo{person}{Jun-Yan Zhu}.}
  \bibinfo{year}{2023}\natexlab{b}.
\newblock \bibinfo{title}{{G}it{H}ub: {A}blating {C}oncepts in
  {T}ext-to-{I}mage {D}iffusion {M}odels}.
\newblock
  \bibinfo{howpublished}{\url{https://github.com/nupurkmr9/concept-ablation}}.
\newblock


\bibitem[Lee et~al\mbox{.}(2022)]%
        {lee2021soundguided}
\bibfield{author}{\bibinfo{person}{Seung~Hyun Lee}, \bibinfo{person}{Wonseok
  Roh}, \bibinfo{person}{Wonmin Byeon}, \bibinfo{person}{Sang~Ho Yoon},
  \bibinfo{person}{Chanyoung Kim}, \bibinfo{person}{Jinkyu Kim}, {and}
  \bibinfo{person}{Sangpil Kim}.} \bibinfo{year}{2022}\natexlab{}.
\newblock \showarticletitle{Sound-Guided Semantic Image Manipulation}. In
  \bibinfo{booktitle}{\emph{Proceedings of the IEEE/CVF Conference on Computer
  Vision and Pattern Recognition (CVPR)}}. \bibinfo{publisher}{IEEE Computer
  Society}, \bibinfo{pages}{3377--3386}.
\newblock


\bibitem[Li et~al\mbox{.}(2024)]%
        {li2024safegen}
\bibfield{author}{\bibinfo{person}{Xinfeng Li}, \bibinfo{person}{Yuchen Yang},
  \bibinfo{person}{Jiangyi Deng}, \bibinfo{person}{Chen Yan},
  \bibinfo{person}{Yanjiao Chen}, \bibinfo{person}{Xiaoyu Ji}, {and}
  \bibinfo{person}{Wenyuan Xu}.} \bibinfo{year}{2024}\natexlab{}.
\newblock \showarticletitle{SafeGen: Mitigating Unsafe Content Generation in
  Text-to-Image Models}. In \bibinfo{booktitle}{\emph{Proceedings of the 2024
  {ACM} {SIGSAC} Conference on Computer and Communications Security (CCS)}}.
  \bibinfo{publisher}{Association for Computing Machinery}.
\newblock


\bibitem[Liang et~al\mbox{.}(2022)]%
        {ModalityGap}
\bibfield{author}{\bibinfo{person}{Weixin Liang}, \bibinfo{person}{Yuhui
  Zhang}, \bibinfo{person}{Yongchan Kwon}, \bibinfo{person}{Serena Yeung},
  {and} \bibinfo{person}{James Zou}.} \bibinfo{year}{2022}\natexlab{}.
\newblock \showarticletitle{Mind the Gap: Understanding the Modality Gap in
  Multi-modal Contrastive Representation Learning}. In
  \bibinfo{booktitle}{\emph{NeurIPS}}. \bibinfo{publisher}{OpenReview.net}.
\newblock
\urldef\tempurl%
\url{https://openreview.net/forum?id=S7Evzt9uit3}
\showURL{%
\tempurl}


\bibitem[Lin et~al\mbox{.}(2014)]%
        {coco}
\bibfield{author}{\bibinfo{person}{Tsung{-}Yi Lin}, \bibinfo{person}{Michael
  Maire}, \bibinfo{person}{Serge~J. Belongie}, \bibinfo{person}{Lubomir~D.
  Bourdev}, \bibinfo{person}{Ross~B. Girshick}, \bibinfo{person}{James Hays},
  \bibinfo{person}{Pietro Perona}, \bibinfo{person}{Deva Ramanan},
  \bibinfo{person}{Piotr Doll{\'{a}}r}, {and} \bibinfo{person}{C.~Lawrence
  Zitnick}.} \bibinfo{year}{2014}\natexlab{}.
\newblock \showarticletitle{Microsoft {COCO:} Common Objects in Context}.
\newblock \bibinfo{journal}{\emph{CoRR}}  \bibinfo{volume}{abs/1405.0312}
  (\bibinfo{year}{2014}).
\newblock
\showeprint[arXiv]{1405.0312}
\urldef\tempurl%
\url{http://arxiv.org/abs/1405.0312}
\showURL{%
\tempurl}


\bibitem[Ma et~al\mbox{.}(2024)]%
        {ma2024jailbreaking}
\bibfield{author}{\bibinfo{person}{Jiachen Ma} {et~al\mbox{.}}}
  \bibinfo{year}{2024}\natexlab{}.
\newblock \showarticletitle{Jailbreaking Prompt Attack: A Controllable
  Adversarial Attack against Diffusion Models}. In
  \bibinfo{booktitle}{\emph{arXiv:2404.02928}}.
\newblock


\bibitem[Madry et~al\mbox{.}(2018)]%
        {MadryPgd}
\bibfield{author}{\bibinfo{person}{Aleksander Madry},
  \bibinfo{person}{Aleksandar Makelov}, \bibinfo{person}{Ludwig Schmidt},
  \bibinfo{person}{Dimitris Tsipras}, {and} \bibinfo{person}{Adrian Vladu}.}
  \bibinfo{year}{2018}\natexlab{}.
\newblock \showarticletitle{Towards Deep Learning Models Resistant to
  Adversarial Attacks}. In \bibinfo{booktitle}{\emph{6th International
  Conference on Learning Representations, {ICLR} 2018, Vancouver, BC, Canada,
  April 30 - May 3, 2018, Conference Track Proceedings}}.
  \bibinfo{publisher}{OpenReview.net}.
\newblock
\urldef\tempurl%
\url{https://openreview.net/forum?id=rJzIBfZAb}
\showURL{%
\tempurl}


\bibitem[Matthias(2023)]%
        {Matthias_2023}
\bibfield{author}{\bibinfo{person}{Meg Matthias}.}
  \bibinfo{year}{2023}\natexlab{}.
\newblock \bibinfo{title}{Why does AI art screw up hands and fingers?}
\newblock
\newblock
\urldef\tempurl%
\url{https://www.britannica.com/topic/Why-does-AI-art-screw-up-hands-and-fingers-2230501}
\showURL{%
\tempurl}


\bibitem[Noever and Noever(2021)]%
        {noever2021typographicattck}
\bibfield{author}{\bibinfo{person}{David~A. Noever} {and}
  \bibinfo{person}{Samantha E.~Miller Noever}.}
  \bibinfo{year}{2021}\natexlab{}.
\newblock \showarticletitle{Reading Isn't Believing: Adversarial Attacks On
  Multi-Modal Neurons}. In \bibinfo{booktitle}{\emph{arXiv:2103.10480}}.
\newblock


\bibitem[Pham et~al\mbox{.}(2024a)]%
        {pham2024robust}
\bibfield{author}{\bibinfo{person}{Minh Pham} {et~al\mbox{.}}}
  \bibinfo{year}{2024}\natexlab{a}.
\newblock \showarticletitle{Robust Concept Erasure Using Task Vectors}. In
  \bibinfo{booktitle}{\emph{arXiv:2404.03631}}.
\newblock


\bibitem[Pham et~al\mbox{.}(2024b)]%
        {pham2023circumventing}
\bibfield{author}{\bibinfo{person}{Minh Pham}, \bibinfo{person}{Kelly~O.
  Marshall}, \bibinfo{person}{Niv Cohen}, \bibinfo{person}{Govind Mittal},
  {and} \bibinfo{person}{Chinmay Hegde}.} \bibinfo{year}{2024}\natexlab{b}.
\newblock \showarticletitle{Circumventing Concept Erasure Methods For
  Text-to-Image Generative Models}. In \bibinfo{booktitle}{\emph{International
  Conference on Learning Representation}}. \bibinfo{publisher}{OpenReview.net}.
\newblock


\bibitem[Qu et~al\mbox{.}(2023)]%
        {githubGitHubYitingQuunsafediffusion}
\bibfield{author}{\bibinfo{person}{Yiting Qu}, \bibinfo{person}{Xinyue Shen},
  \bibinfo{person}{Xinlei He}, \bibinfo{person}{Michael Backes},
  \bibinfo{person}{Savvas Zannettou}, {and} \bibinfo{person}{Yang Zhang}.}
  \bibinfo{year}{2023}\natexlab{}.
\newblock \bibinfo{title}{{G}it{H}ub: unsafe-diffusion}.
\newblock
  \bibinfo{howpublished}{\url{https://github.com/YitingQu/unsafe-diffusion/tree/main}}.
\newblock


\bibitem[Radford et~al\mbox{.}(2021)]%
        {radford2021learning}
\bibfield{author}{\bibinfo{person}{Alec Radford}, \bibinfo{person}{Jong~Wook
  Kim}, \bibinfo{person}{Chris Hallacy}, \bibinfo{person}{Aditya Ramesh},
  \bibinfo{person}{Gabriel Goh}, \bibinfo{person}{Sandhini Agarwal},
  \bibinfo{person}{Girish Sastry}, \bibinfo{person}{Amanda Askell},
  \bibinfo{person}{Pamela Mishkin}, \bibinfo{person}{Jack Clark},
  \bibinfo{person}{Gretchen Krueger}, {and} \bibinfo{person}{Ilya Sutskever}.}
  \bibinfo{year}{2021}\natexlab{}.
\newblock \showarticletitle{Learning Transferable Visual Models From Natural
  Language Supervision}. In \bibinfo{booktitle}{\emph{Proceedings of the 38th
  International Conference on Machine Learning, {ICML} 2021, 18-24 July 2021,
  Virtual Event}} \emph{(\bibinfo{series}{Proceedings of Machine Learning
  Research}, Vol.~\bibinfo{volume}{139})},
  \bibfield{editor}{\bibinfo{person}{Marina Meila} {and} \bibinfo{person}{Tong
  Zhang}} (Eds.). \bibinfo{publisher}{{PMLR}}, \bibinfo{pages}{8748--8763}.
\newblock
\urldef\tempurl%
\url{http://proceedings.mlr.press/v139/radford21a.html}
\showURL{%
\tempurl}


\bibitem[Ramesh et~al\mbox{.}(2021)]%
        {ramesh2021dalle}
\bibfield{author}{\bibinfo{person}{Aditya Ramesh}, \bibinfo{person}{Rishabh
  Goyal}, \bibinfo{person}{Alessandro Sordoni}, \bibinfo{person}{Yaniv Ovadia},
  {and} \bibinfo{person}{Geoffrey~E. Hinton}.} \bibinfo{year}{2021}\natexlab{}.
\newblock \showarticletitle{DALL·E 2: The Flower That Blooms in Adversity}. In
  \bibinfo{booktitle}{\emph{OpenAI Blog}}. \bibinfo{publisher}{OpenAI}.
\newblock
\urldef\tempurl%
\url{https://openai.com/blog/dall-e-2/}
\showURL{%
\tempurl}


\bibitem[Rando et~al\mbox{.}({[n.\,d.]})]%
        {rando2022redteaming}
\bibfield{author}{\bibinfo{person}{Javier Rando}, \bibinfo{person}{Daniel
  Paleka}, \bibinfo{person}{David Lindner}, \bibinfo{person}{Lennart Heim},
  {and} \bibinfo{person}{Florian Tramer}.} \bibinfo{year}{[n.\,d.]}\natexlab{}.
\newblock \showarticletitle{Red-Teaming the Stable Diffusion Safety Filter}. In
  \bibinfo{booktitle}{\emph{NeurIPS ML Safety Workshop}}.
  \bibinfo{publisher}{OpenReview.net}.
\newblock


\bibitem[Rombach et~al\mbox{.}(2022a)]%
        {stable-diffusion}
\bibfield{author}{\bibinfo{person}{Robin Rombach}, \bibinfo{person}{Andreas
  Blattmann}, \bibinfo{person}{Dominik Lorenz}, \bibinfo{person}{Patrick
  Esser}, {and} \bibinfo{person}{Bj\"orn Ommer}.}
  \bibinfo{year}{2022}\natexlab{a}.
\newblock \showarticletitle{High-Resolution Image Synthesis With Latent
  Diffusion Models}. In \bibinfo{booktitle}{\emph{Proceedings of the IEEE/CVF
  Conference on Computer Vision and Pattern Recognition (CVPR)}}.
  \bibinfo{publisher}{IEEE Computer Society}, \bibinfo{pages}{10684--10695}.
\newblock


\bibitem[Rombach et~al\mbox{.}(2022b)]%
        {Rombach_2022_CVPR}
\bibfield{author}{\bibinfo{person}{Robin Rombach}, \bibinfo{person}{Andreas
  Blattmann}, \bibinfo{person}{Dominik Lorenz}, \bibinfo{person}{Patrick
  Esser}, {and} \bibinfo{person}{Bj\"orn Ommer}.}
  \bibinfo{year}{2022}\natexlab{b}.
\newblock \showarticletitle{High-Resolution Image Synthesis With Latent
  Diffusion Models}. In \bibinfo{booktitle}{\emph{Proceedings of the IEEE/CVF
  Conference on Computer Vision and Pattern Recognition (CVPR)}}.
  \bibinfo{publisher}{IEEE Computer Society}, \bibinfo{pages}{10684--10695}.
\newblock


\bibitem[Schramowski et~al\mbox{.}(2023)]%
        {schramowski2022safe}
\bibfield{author}{\bibinfo{person}{Patrick Schramowski},
  \bibinfo{person}{Manuel Brack}, \bibinfo{person}{Björn Deiseroth}, {and}
  \bibinfo{person}{Kristian Kersting}.} \bibinfo{year}{2023}\natexlab{}.
\newblock \showarticletitle{Safe Latent Diffusion: Mitigating Inappropriate
  Degeneration in Diffusion Models}. In \bibinfo{booktitle}{\emph{Proceedings
  of the {IEEE} Conference on Computer Vision and Pattern Recognition
  ({CVPR})}}. \bibinfo{publisher}{IEEE Computer Society}.
\newblock


\bibitem[Schuhmann et~al\mbox{.}(2022)]%
        {schuhmann2022laion}
\bibfield{author}{\bibinfo{person}{Christoph Schuhmann},
  \bibinfo{person}{Romain Beaumont}, \bibinfo{person}{Richard Vencu},
  \bibinfo{person}{Cade~W Gordon}, \bibinfo{person}{Ross Wightman},
  \bibinfo{person}{Theo Coombes}, \bibinfo{person}{Aarush Katta},
  \bibinfo{person}{Clayton Mullis}, \bibinfo{person}{Mitchell Wortsman},
  \bibinfo{person}{Patrick Schramowski}, \bibinfo{person}{Srivatsa~R
  Kundurthy}, \bibinfo{person}{Katherine Crowson}, \bibinfo{person}{Ludwig
  Schmidt}, \bibinfo{person}{Robert Kaczmarczyk}, {and} \bibinfo{person}{Jenia
  Jitsev}.} \bibinfo{year}{2022}\natexlab{}.
\newblock \showarticletitle{Laion-5b: An open large-scale dataset for training
  next generation image-text models}. In \bibinfo{booktitle}{\emph{Thirty-sixth
  Conference on Neural Information Processing Systems Datasets and Benchmarks
  Track}}. \bibinfo{pages}{1--2}.
\newblock


\bibitem[Somepalli et~al\mbox{.}(2023)]%
        {somepalli2022diffusion}
\bibfield{author}{\bibinfo{person}{G. Somepalli}, \bibinfo{person}{V. Singla},
  \bibinfo{person}{M. Goldblum}, \bibinfo{person}{J. Geiping}, {and}
  \bibinfo{person}{T. Goldstein}.} \bibinfo{year}{2023}\natexlab{}.
\newblock \showarticletitle{Diffusion Art or Digital Forgery? Investigating
  Data Replication in Diffusion Models}. In \bibinfo{booktitle}{\emph{2023
  IEEE/CVF Conference on Computer Vision and Pattern Recognition (CVPR)}}.
  \bibinfo{publisher}{IEEE Computer Society}, \bibinfo{address}{Los Alamitos,
  CA, USA}, \bibinfo{pages}{6048--6058}.
\newblock
\urldef\tempurl%
\url{https://doi.org/10.1109/CVPR52729.2023.00586}
\showDOI{\tempurl}


\bibitem[Tsai et~al\mbox{.}(2024)]%
        {Tsai2023RingABellHR}
\bibfield{author}{\bibinfo{person}{Yu-Lin Tsai}, \bibinfo{person}{Chia-Yi Hsu},
  \bibinfo{person}{Chulin Xie}, \bibinfo{person}{Chih-Hsun Lin},
  \bibinfo{person}{Jia-You Chen}, \bibinfo{person}{Bo Li},
  \bibinfo{person}{Pin-Yu Chen}, \bibinfo{person}{Chia-Mu Yu}, {and}
  \bibinfo{person}{Chun-Ying Huang}.} \bibinfo{year}{2024}\natexlab{}.
\newblock \showarticletitle{Ring-A-Bell! How Reliable are Concept Removal
  Methods For Diffusion Models?}. In \bibinfo{booktitle}{\emph{The Twelfth
  International Conference on Learning Representations}}.
  \bibinfo{publisher}{OpenReview.net}.
\newblock
\urldef\tempurl%
\url{https://openreview.net/forum?id=lm7MRcsFiS}
\showURL{%
\tempurl}


\bibitem[van~den Oord et~al\mbox{.}(2018)]%
        {oord2019representation}
\bibfield{author}{\bibinfo{person}{A{\"{a}}ron van~den Oord},
  \bibinfo{person}{Yazhe Li}, {and} \bibinfo{person}{Oriol Vinyals}.}
  \bibinfo{year}{2018}\natexlab{}.
\newblock \showarticletitle{Representation Learning with Contrastive Predictive
  Coding}, In \bibinfo{booktitle}{arXiv:1807.03748}.
\newblock \bibinfo{journal}{\emph{CoRR}}  \bibinfo{volume}{abs/1807.03748}.
\newblock
\urldef\tempurl%
\url{http://arxiv.org/abs/1807.03748}
\showURL{%
\tempurl}


\bibitem[Verne(2024)]%
        {VERNE_2024}
\bibfield{author}{\bibinfo{person}{Jules Verne}.}
  \bibinfo{year}{2024}\natexlab{}.
\newblock \bibinfo{booktitle}{\emph{Twenty Thousand Leagues under the sea}}.
\newblock \bibinfo{publisher}{Aladin Books}.
\newblock


\bibitem[Wang et~al\mbox{.}(2024a)]%
        {githubGitHubWangModerator}
\bibfield{author}{\bibinfo{person}{Peiran Wang}, \bibinfo{person}{Qiyu Li},
  \bibinfo{person}{Longxuan Yu}, \bibinfo{person}{Ziyao Wang},
  \bibinfo{person}{Ang Li}, {and} \bibinfo{person}{Haojian Jin}.}
  \bibinfo{year}{2024}\natexlab{a}.
\newblock \bibinfo{title}{{G}it{H}ub: {M}oderator: {M}oderating
  {T}ext-to-{I}mage {D}iffusion {M}odels through {F}ine-grained {C}ontext-based
  {P}olicies}.
\newblock
  \bibinfo{howpublished}{\url{https://github.com/DataSmithLab/Moderator}}.
\newblock


\bibitem[Wang et~al\mbox{.}(2024b)]%
        {wang2024moderator}
\bibfield{author}{\bibinfo{person}{Peiran Wang}, \bibinfo{person}{Qiyu Li},
  \bibinfo{person}{Longxuan Yu}, \bibinfo{person}{Ziyao Wang},
  \bibinfo{person}{Ang Li}, {and} \bibinfo{person}{Haojian Jin}.}
  \bibinfo{year}{2024}\natexlab{b}.
\newblock \showarticletitle{Moderator: Moderating Text-to-Image Diffusion
  Models through Fine-grained Context-based Policies}. In
  \bibinfo{booktitle}{\emph{Proceedings of the 2024 {ACM} {SIGSAC} Conference
  on Computer and Communications Security (CCS)}}.
  \bibinfo{publisher}{Association for Computing Machinery}.
\newblock


\bibitem[Wen et~al\mbox{.}(2023)]%
        {wen2023hard}
\bibfield{author}{\bibinfo{person}{Yuxin Wen}, \bibinfo{person}{Neel Jain},
  \bibinfo{person}{John Kirchenbauer}, \bibinfo{person}{Micah Goldblum},
  \bibinfo{person}{Jonas Geiping}, {and} \bibinfo{person}{Tom Goldstein}.}
  \bibinfo{year}{2023}\natexlab{}.
\newblock \showarticletitle{Hard Prompts Made Easy: Gradient-Based Discrete
  Optimization for Prompt Tuning and Discovery}. In
  \bibinfo{booktitle}{\emph{Thirty-seventh Conference on Neural Information
  Processing Systems}}. \bibinfo{publisher}{OpenReview.net}.
\newblock
\urldef\tempurl%
\url{https://openreview.net/forum?id=VOstHxDdsN}
\showURL{%
\tempurl}


\bibitem[Yang et~al\mbox{.}(2023)]%
        {yang2023robust}
\bibfield{author}{\bibinfo{person}{Wenhan Yang}, \bibinfo{person}{Jingdong
  Gao}, {and} \bibinfo{person}{Baharan Mirzasoleiman}.}
  \bibinfo{year}{2023}\natexlab{}.
\newblock \showarticletitle{Robust Contrastive Language-Image Pretraining
  against Data Poisoning and Backdoor Attacks}. In
  \bibinfo{booktitle}{\emph{Thirty-seventh Conference on Neural Information
  Processing Systems}}. \bibinfo{publisher}{OpenReview.net}.
\newblock
\urldef\tempurl%
\url{https://openreview.net/forum?id=ONwL9ucoYG}
\showURL{%
\tempurl}


\bibitem[Yang et~al\mbox{.}(2024)]%
        {yang2023sneakyprompt}
\bibfield{author}{\bibinfo{person}{Yuchen Yang}, \bibinfo{person}{Bo Hui},
  \bibinfo{person}{Haolin Yuan}, \bibinfo{person}{Neil Gong}, {and}
  \bibinfo{person}{Yinzhi Cao}.} \bibinfo{year}{2024}\natexlab{}.
\newblock \showarticletitle{SneakyPrompt: Jailbreaking Text-to-image Generative
  Models}. In \bibinfo{booktitle}{\emph{Proceedings of the IEEE Symposium on
  Security and Privacy}}. \bibinfo{publisher}{IEEE Computer Society}.
\newblock


\bibitem[Yiting et~al\mbox{.}(2023)]%
        {unsafeDiff}
\bibfield{author}{\bibinfo{person}{Qu Yiting}, \bibinfo{person}{Shen Xinyue},
  \bibinfo{person}{He Xinlei}, \bibinfo{person}{Backes Michael},
  \bibinfo{person}{Zannettou Savvas}, {and} \bibinfo{person}{Zhang Yang}.}
  \bibinfo{year}{2023}\natexlab{}.
\newblock \showarticletitle{Unsafe Diffusion: On the Generation of Unsafe
  Images and Hateful Memes From Text-To-Image Models}. In
  \bibinfo{booktitle}{\emph{ACM SIGSAC Conference on Computer and
  Communications Security (CCS)}}. \bibinfo{publisher}{ACM}.
\newblock


\bibitem[Zhang et~al\mbox{.}(2023b)]%
        {githubGitHubSHILabsForgetMeNot}
\bibfield{author}{\bibinfo{person}{Eric Zhang}, \bibinfo{person}{Kai Wang},
  \bibinfo{person}{Xingqian Xu}, \bibinfo{person}{Zhangyang Wang}, {and}
  \bibinfo{person}{Humphrey Shi}.} \bibinfo{year}{2023}\natexlab{b}.
\newblock \bibinfo{title}{{G}it{H}ub - {F}orget-{M}e-{N}ot}.
\newblock
  \bibinfo{howpublished}{\url{https://github.com/SHI-Labs/Forget-Me-Not}}.
\newblock


\bibitem[Zhang et~al\mbox{.}(2024)]%
        {zhang2023forgetmenot}
\bibfield{author}{\bibinfo{person}{Gong Zhang}, \bibinfo{person}{Kai Wang},
  \bibinfo{person}{Xingqian Xu}, \bibinfo{person}{Zhangyang Wang}, {and}
  \bibinfo{person}{Humphrey Shi}.} \bibinfo{year}{2024}\natexlab{}.
\newblock \showarticletitle{Forget-me-not: Learning to forget in text-to-image
  diffusion models}. In \bibinfo{booktitle}{\emph{Proceedings of the IEEE/CVF
  Conference on Computer Vision and Pattern Recognition}}.
  \bibinfo{publisher}{IEEE Society}, \bibinfo{pages}{1755--1764}.
\newblock


\bibitem[Zhang et~al\mbox{.}(2019)]%
        {zhang2019theoretically}
\bibfield{author}{\bibinfo{person}{Hongyang Zhang}, \bibinfo{person}{Yaodong
  Yu}, \bibinfo{person}{Jiantao Jiao}, \bibinfo{person}{Eric Xing},
  \bibinfo{person}{Laurent El~Ghaoui}, {and} \bibinfo{person}{Michael Jordan}.}
  \bibinfo{year}{2019}\natexlab{}.
\newblock \showarticletitle{Theoretically principled trade-off between
  robustness and accuracy}. In \bibinfo{booktitle}{\emph{International
  conference on machine learning}}. PMLR, \bibinfo{publisher}{PLMR},
  \bibinfo{pages}{7472--7482}.
\newblock


\bibitem[Zhang et~al\mbox{.}(2023a)]%
        {zhang2023diffsmooth}
\bibfield{author}{\bibinfo{person}{Jiawei Zhang}, \bibinfo{person}{Zhongzhu
  Chen}, \bibinfo{person}{Huan Zhang}, \bibinfo{person}{Chaowei Xiao}, {and}
  \bibinfo{person}{Bo Li}.} \bibinfo{year}{2023}\natexlab{a}.
\newblock \showarticletitle{$\{$DiffSmooth$\}$: Certifiably robust learning via
  diffusion models and local smoothing}. In \bibinfo{booktitle}{\emph{32nd
  USENIX Security Symposium (USENIX Security 23)}}. \bibinfo{publisher}{Usenix
  Association}, \bibinfo{pages}{4787--4804}.
\newblock


\end{thebibliography}

\appendix

\section*{Appendix}


\section{Certifying Robustness of \method}\label{sec:bound}

\subsection*{Empirical Validation}\label{sec:empirical}

We now compute the maximum noise that \method can tolerate for each unacceptable image's embedding using Equation~\ref{eq:bound}. 
Following prior literature on certified robustness~\cite{cohen2019certified}, we compute the certified accuracy described in~\cite{cohen2019certified} to evaluate the robustness of \method.
Certified accuracy at radius $r$ is the fraction of unacceptable images which are correctly classified and are robust against adversarial noise $\delta > r$.
This shows the robustness of \method against attacks under some noise $r$. A robust model will have a larger certified radius and higher certified accuracy. 
Since we add noise directly to $\phi_x(\imgunacc)$, we compare our certified accuracy with the accuracy of clean unacceptable images (without adversarial noise) which we refer to as ``clean accuracy''. 
Ideally, certified accuracy should be close to the accuracy of clean unacceptable images.

\begin{figure}[htb]
    \centering
    \begin{subfigure}[b]{0.49\columnwidth}
        \centering
        \includegraphics[width=\columnwidth]{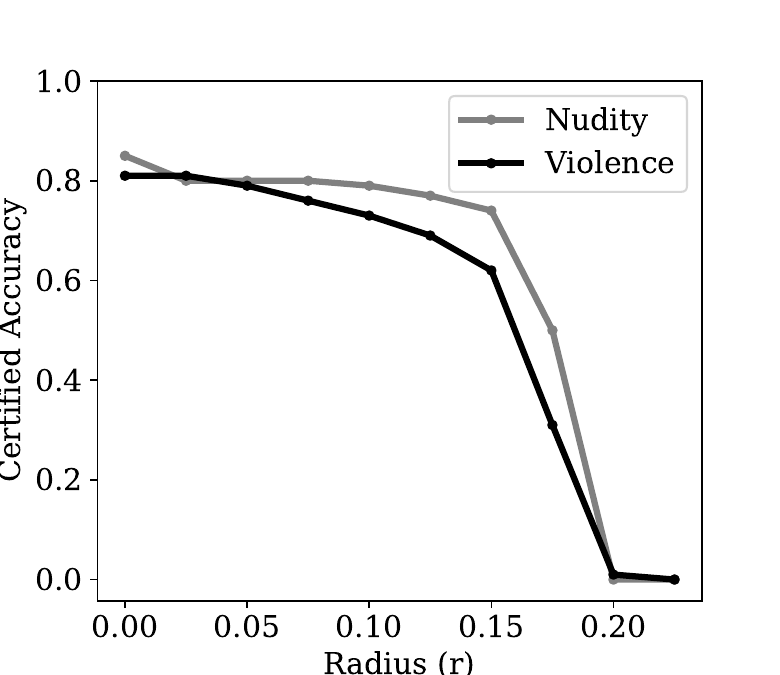}
        \caption{Group-1}
    \end{subfigure}
    \hfill
    \begin{subfigure}[b]{0.49\columnwidth}
        \centering
        \includegraphics[width=\columnwidth]{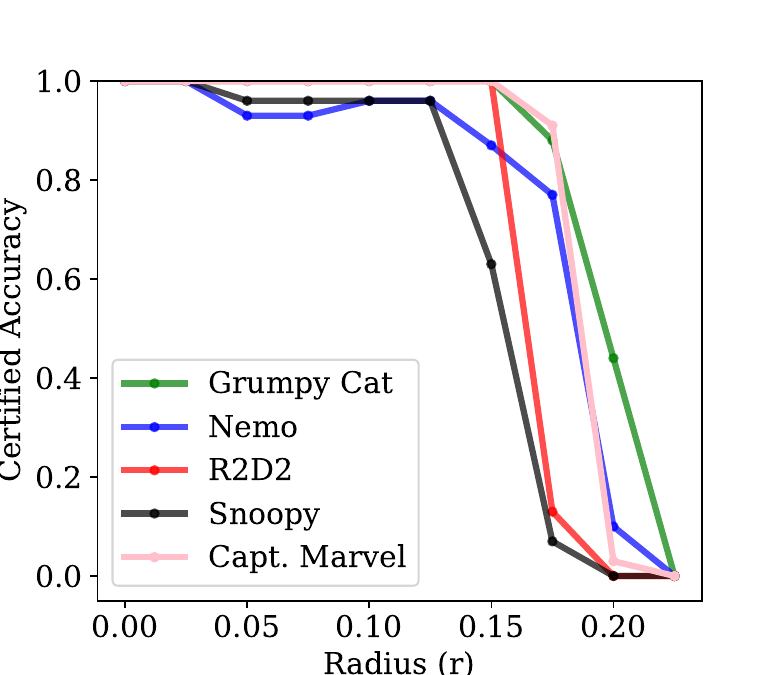}
        \caption{Group-2}
    \end{subfigure}
    \\
    \begin{subfigure}[b]{0.49\columnwidth}
        \centering
        \includegraphics[width=\columnwidth]{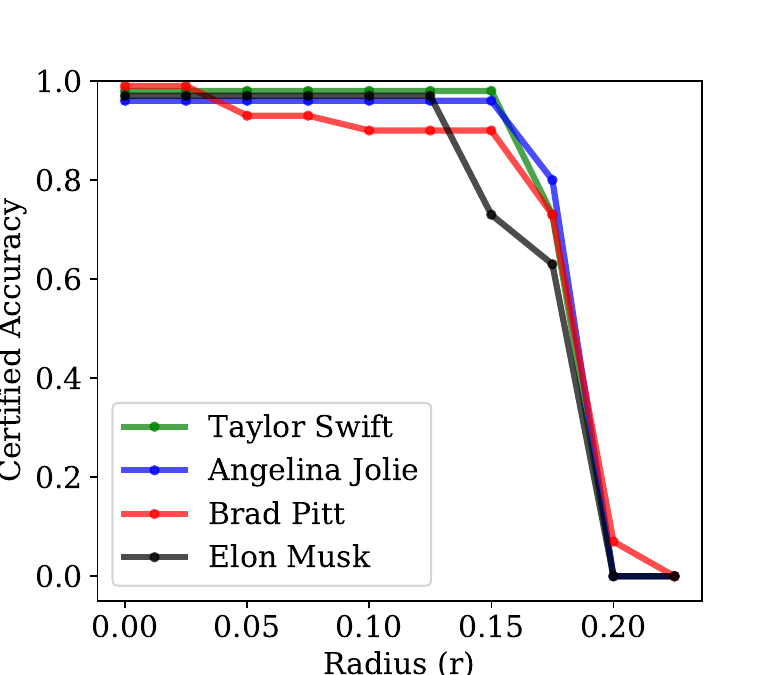}
        \caption{Group-3}
    \end{subfigure}
    \caption{Certified accuracy of \method{} vs. adversarial noise $\delta$, for a strong \adv{} with access to embeddings of generated images.}
    \label{fig:certified}
\end{figure}

We present the results in Figure~\ref{fig:certified} for the three groups of concepts. 
Clean accuracy in Figure~\ref{fig:certified} is the certified accuracy at radius zero. 
\method is robust against $\delta<0.07$, incurring less than a $5\%$ drop in certified accuracy. 
When $\delta<0.15$, the certified accuracy remains higher than $50\%$ for all concepts. 
\method is particularly robust for some concepts in Group-2 (\emph{Grumpy Cat}, \emph{R2D2}, \emph{Captain Marvel}), and Group-3 (\emph{Taylor Swift}, \emph{Angelina Jolie}, and \emph{Elon Musk}). For these concepts, the certified accuracy remains the same for the clean unacceptable images until $\delta>0.15$.
Further, \method is more robust for concepts where the clean accuracy is $1.00$ (CLIP accuracy from Table~\ref{tab:effectiveness}). We find that the robustness is higher for concepts on which \method is more accurate. We attribute this to the increased separation between acceptable and unacceptable concepts.

\subsection*{Practical Implications}\label{sec:implications} 

Having discussed the theoretical bound and empirically validated it on different concepts, we now revisit the practicality of this bound.
We discuss the usefulness of the certification and revisit our assumption about \adv's capability.

\noindent\textbf{Usefulness of Certified Bound.} 
In Figure~\ref{fig:certified}, we find that the certified bound is less than \(0.15\) across all the concepts. We found this to be smaller than the $l_2$-norms of realistic image embeddings, which had a mean of \(17\).
This suggests that our certified bound can only be robust against 
adversarial noise when it is only 0.8\% (=0.15/17) of the embeddings.

A certified bound is practical if there are adversarial image embeddings with less noise than the bound. Then, the bound is effective against these embeddings.
We use \method without fine-tuning with Equation~\ref{eq:ft1} to check the existence of such adversarial image embeddings. We can find embeddings that  \emph{potentially} evade \method (without fine-tuning) when the noise is as small as $0.028$.
Our certified bound is useful against such embeddings\footnote{Note that to find an actual attack against \method, \adv will have to (a) find a prompt that generates this perturbed embedding, and (b) ensure that the resulting image retains the unacceptable content.}.

However, the distance between acceptable and unacceptable images, which is at least \(7\), is much larger than the certified bound. This suggests that our certified bound is loose. 
We leave a tighter certified bound as future work.

\noindent\textbf{\adv's Capability.} To compute the certified bound, we assumed a strong \adv who can directly add adversarial noise to the \emph{embeddings}.
In practice, \adv can only modify the \emph{prompts} sent to the \tti model, and can only obtain the corresponding filtered outputs. Hence, in practice, \adv is much weaker and the robustness of \method is much higher than indicated in Figure~\ref{fig:certified}.

To illustrate this, we consider a concrete attack that \adv could adopt given its inability to directly add adversarial noise to embeddings:
\adv begins with unacceptable images and incorporate adversarial noise using standard evasion techniques (e.g., PGD~\cite{MadryPgd}) to find an adversarial example that evades the \method classifier. 
\adv then finds the corresponding adversarial prompt using one of the attacks (e.g., PEZ+). 
We want to see if \target still generates an adversarial image which evades \method. 
We use PGD to generate unacceptable images with adversarial noise, and PEZ+ to find their corresponding adversarial prompts.
We find that \target fails to generate an adversarial image which evades \method using the adversarial prompt.
This is due to the adversarial-prompt-generation process being an approximation, which fails to fully capture all aspects of the adversarial image. 
Moreover, using the \tti model to generate the image from the adversarial prompt is unlikely to capture the adversarial noise due to the de-noising used in the diffusion model. This claim is further supported by prior literature on the robustness of diffusion models~\cite{Kim_2022_CVPR,zhang2023diffsmooth}. 

\newpage 
We compare the original adversarial images with the images generated from their adversarial prompts. We present one concept from each group in Table~\ref{fig:adversarial_images}. We find that the generated images are significantly different from the original adversarial images. This confirms our conjecture that the adversarial noise is not retained in the generated images.
A more thorough exploration of such an attack is left as future work. Based on the above preliminary exploration, we conjecture that \method is likely to be robust against such attacks by \adv with realistic capabilities.

\begin{table}[htb]
\caption{(Column 1) Adversarial image ($\imgadv$) using PGD~\cite{MadryPgd} against \method, (Column 2) adversarial prompt ($\advprompt$) generated from $\imgadv$ using PEZ~\cite{wen2023hard}, and (Column 3) image generated by SDv1.4 \tti model using $\advprompt$ as input.}
\label{tab:framework}
\centering
\footnotesize
\begin{tabular}{|l|p{2cm}|p{2cm}|p{2cm}|}
\hline 
{\bf Concept} & {\bf Adversarial Image ($\imgadv$)} & {\bf Adversarial Prompt ($\advprompt$)} & {\bf Image Generated from $\advprompt$} \\
\hline     

\textbf{Nudity} 
&
\includegraphics[width=0.20\columnwidth]{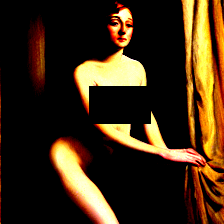}
& \vspace{-1.5cm}``artsy wbo venus moc bday oilandgoddess thru cropped endurindiefilm cropped r underetal <copyright sign>'' & 
\includegraphics[width=0.20\columnwidth]{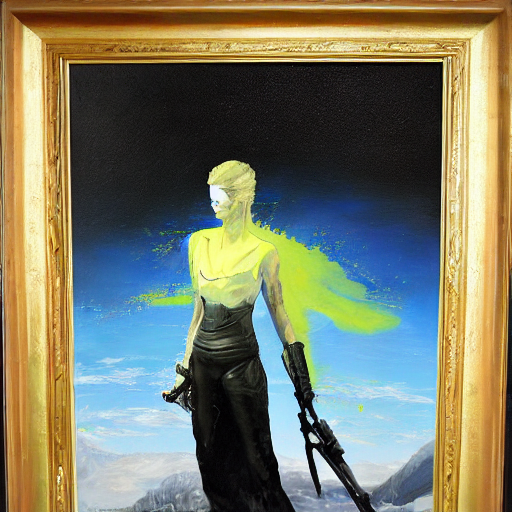}
\\
\hline

\textbf{Nemo}
&
\includegraphics[width=0.20\columnwidth]{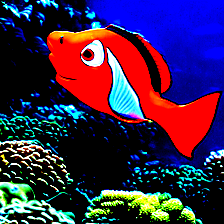}
& \vspace{-1.5cm}``moma fishy pet <heart emoji> constrafrm orange optimistic soaking …… vacancy trippy troubles groovy averages \~!'' & 
\includegraphics[width=0.20\columnwidth]{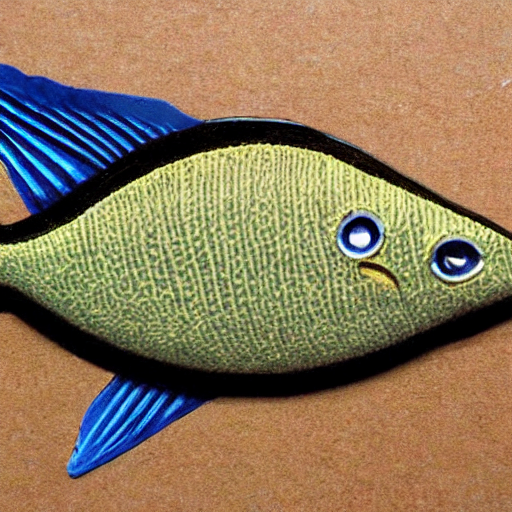}
\\
\hline

\textbf{Elon Musk}
&
\includegraphics[width=0.20\columnwidth]{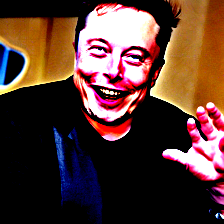}
& \vspace{-1.5cm}``poet moderstare rested wakeupamerica (" blurred vaportide driverless <smiley emoji> broker celebrated mandelclap'' & 
\includegraphics[width=0.20\columnwidth]{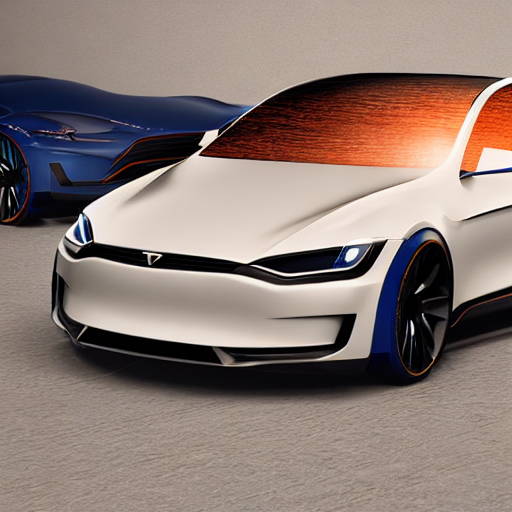}
\\
\hline
\end{tabular}
\label{fig:adversarial_images}
\end{table}

\section{Choice of Alpha}\label{sec:alpha}

For \autoref{eq:ft2} used for \emph{nudity} and \emph{violence} concepts. We tried several values of $\alpha_{(.)}$ (=0.25, 0.5, 1, 1.5, 2) to determine the best values. In \autoref{tab:Alphas}, we show the FNR on unacceptable prompts from $\dvalunacc$, FPR on acceptable prompts from $\dvalacc$, and the sum of both. Lower is better in each case. 
%

\begin{table*}[!htb]
    \centering
    \caption{\underline{Choosing $\alpha_{(.)}$}: Comparing FNR, FPR, and their sum for different $\alpha_{(\cdot)}$ (lower is better). \textit{\textbf{Bold}} indicates the smallest sum.}
    \resizebox{2\columnwidth}{!}{
    \begin{tabular}{|c|ccc|ccc|ccc|ccc|ccc|}
        \hline
        $\alpha_{(\cdot)}$ & \multicolumn{3}{c|}{\textbf{0.25}} & \multicolumn{3}{c|}{\textbf{0.5}} & \multicolumn{3}{c|}{\textbf{1}} & \multicolumn{3}{c|}{\textbf{1.5}} & \multicolumn{3}{c|}{\textbf{2}} \\
        \hline
         & \textbf{FNR} & \textbf{FPR} & \textbf{Sum} & \textbf{FNR} & \textbf{FPR} & \textbf{Sum} & \textbf{FNR} & \textbf{FPR} & \textbf{Sum} & \textbf{FNR} & \textbf{FPR} & \textbf{Sum} & \textbf{FNR} & \textbf{FPR} & \textbf{Sum} \\
        \hline
        \textbf{Nudity} & 0.01 $\pm$ 0.02 & 0.26 $\pm$ 0.09 & 0.27 & 0.03 $\pm$ 0.03 & 0.20 $\pm$ 0.04 & 0.23 & 0.03 $\pm$ 0.01 & 0.06 $\pm$ 0.05 & \textbf{\textit{0.09}} & 0.09 $\pm$ 0.05 & 0.04 $\pm$ 0.01 & 0.13 & 0.08 $\pm$ 0.07 & 0.07 $\pm$ 0.01 & 0.15 \\
        \textbf{Violence} & 0.36 $\pm$ 0.00 & 0.45 $\pm$ 0.07 & 0.81 & 0.22 $\pm$ 0.08 & 0.64 $\pm$ 0.03 & 0.86 & 0.24 $\pm$ 0.09 & 0.50 $\pm$ 0.08 & \textbf{\textit{0.74}} & 0.27 $\pm$ 0.07 & 0.52 $\pm$ 0.07 & 0.79 & 0.34 $\pm$ 0.01 & 0.57 $\pm$ 0.02 & 0.91 \\
        \hline
    \end{tabular}
    }
    \label{tab:Alphas}
\end{table*}
As shown in \autoref{tab:Alphas}, the $\alpha_{(.)}$=1 yields the lowest FPR + FNR, thus it gives the best trade off between effectiveness (\ref{effective}) and utility (\ref{utility}). 
Hence, we use this in our evaluation.

\end{document}